\documentclass[10pt,journal,compsoc]{IEEEtran}
\usepackage{graphicx,times,amsmath,amssymb,multirow,caption,subcaption,color}
\usepackage[LGR,T1]{fontenc}
\usepackage[latin9]{inputenc}
\usepackage{array}
\usepackage{textcomp}
\usepackage{algorithm2e}
\usepackage{stackrel}
\usepackage{url}
\usepackage{mathtools}

\ifCLASSOPTIONcompsoc
  \usepackage[nocompress]{cite}
\else
  \usepackage{cite}
\fi

\newcommand{\rtext}[1]{\textcolor{black}{#1}}

\makeatletter

\begin{document}

\title{Networking the Boids is More Robust Against Adversarial Learning}

\author{Jiangjun Tang, George Leu and Hussein Abbass
 \thanks{Authors are with the School of Engineering \& IT, University of New South Wales,
 Canberra-Australia, (e-mail: \{j.tang,g.leu,h.abbass\}@adfa.edu.au).}
}

\IEEEtitleabstractindextext{
\begin{abstract}
Swarm behavior using Boids-like models has been studied primarily using close-proximity spatial sensory information (e.g. vision range). In this study, we propose a novel approach in which the classic definition of boids\textquoteright  \ neighborhood that relies on sensory perception and Euclidian space locality is replaced with graph-theoretic network-based proximity mimicking communication and social networks. We demonstrate that networking the boids leads to faster swarming and higher quality of the formation. We further investigate the effect of adversarial learning, whereby an observer attempts to reverse engineer the dynamics of the swarm through observing its behavior. The results show that networking the swarm demonstrated a more robust approach against adversarial learning than a local-proximity neighborhood structure. 
\end{abstract}
\begin{IEEEkeywords}
Boids, \and Collective Behavior, \and swarm intelligence, \and swarm behavior
\end{IEEEkeywords}}

\maketitle

\makeatletter
\def\ps@IEEEtitlepagestyle{
  \def\@oddfoot{\mycopyrightnotice}
  \def\@evenfoot{}
}
\def\mycopyrightnotice{
  {\footnotesize
  \begin{minipage}{\textwidth}
  \centering
  Copyright~\copyright~2017 IEEE. J. Tang, G. Leu and H. A. Abbass, "Networking the Boids is More Robust Against Adversarial Learning," in IEEE Transactions on Network Science and Engineering, vol. PP, no. 99, pp. 1-1. doi: 10.1109/TNSE.2017.2745108
  \end{minipage}
  }
}

\IEEEdisplaynontitleabstractindextext
\IEEEpeerreviewmaketitle

\IEEEraisesectionheading{\section{Introduction}\label{intro}}

\IEEEPARstart{C}{ollective} behaviors of swarms have normally been explained through aggregation of individual behaviors that were assumed to be based on local information perceived through biological sensors, especially vision. This has been evident in a variety of literature including animal swarming behavior \cite{Bajec2009}, classic boids of Reynolds \cite{Reynolds1987}, and recent swarm intelligence research \cite{Garnier2007,Brambilla2013}. 

Reynolds, in his classic boids model \cite{Reynolds1987}, explains collective swarm behaviors through aggregation of individual behaviors that use three rules - cohesion, separation and alignment. The neighborhood is spatially defined and includes other individuals situated in the visual range of each individual, where a visual range is defined by a vision angle and a vision distance. 

An assumption in these models is that spatial proximity is the primary basis for defining neighborhood structures. Individual behaviors such as speed and direction of movements are derived from local information about the neighboring boids, which is acquired in the limits of the spatial Euclidian distance within the range of the biological sensors used. As such, the resultant interaction and the subsequent collective behavior depend on the perceived neighborhood and is bounded by the spatial limits of vision or localized sensors.

We conjecture that swarm performance will improve through networking. A particle in the swarm may behave using a set of rules similar to those described by classic boids \cite{Reynolds1987}, but the neighborhood is not established by vision range and distance, instead it is established by a direct connectivity to other particles. This, in turn, leads inevitably to network topology playing an essential role in the emerging behaviors. This begs the question of whether networks with different topological features, e.g. a random network and a scale-free network, will impact swarm dynamics.

In this study, we investigate the effect of defining neighborhood using graph theoretical relations. We first show that exchange of information via network relations between individuals is superior to individual spatial perception such as vision range. Then, we show that inferring individual behavioral rules based on observed activity of the swarm is significantly more difficult in the case of network-based swarming, compared to classic distance-based relations. Thus, a swarm operating based on network relations is more robust to potential attacks from external entities that intend to learn the swarm's behavioral rules.  

The remaining of the paper is organized as follows. In Section~\ref{Section:LitRev}, we provide a background on swarm intelligence, in which we highlight the main aspects that generated the hypotheses presented above. In Section~\ref{Section:Methodology}, we describe the methodology we employed for this investigation, then we present and discuss the results of the experiments in Section~\ref{Section:ExpResults}. In Section~\ref{Section:Conclusions}, we summarize the findings and conclude the paper.

\section{Background on Swarm Behavior} \label{Section:LitRev}

Swarm behavior has strong roots in biology, where the purpose was to understand the underpinnings of the self-organised collective behavior exhibited by various, apparently simple, species. Examples date back as early as 1930\textquoteright  \ s~\cite{Bajec2009}, investigating collective behavior over vast biological domains, from organisms as simple as bacteria~\cite{BenJacob2000} to ant colonies~\cite{Garnier2007}, fish schools~\cite{Reuter2016} and bird flocks~\cite{Bajec2009}, and further to large mamals herds~\cite{King2012} and even humans~\cite{Templeton2015}. 

Bird flocks generated theoretical ramifications in artificial intelligence and related disciplines, and practical applications in various fields of engineering~\cite{Blum2008}. There are two major types of flocking behavior in birds. First, the large birds, usually following seasonal migrating life-cycles, display grouping behaviors that optimise the way the group uses the air currents in order to minimise aerodynamic drag and subsequent fatigue. Popular flocking behaviors observed and investigated in this direction are the so-called line formations, which include patterns like V-shape, J-shape, echelon, straight line, waved line, and others~\cite{Bajec2009}. Second, the small non-migrating birds, adopt grouping behaviors that facilitate aspects like food search, security against predators, territorial ownership, or breeding~\cite{Demsar2014}.

The pioneering work of Reynolds~\cite{Reynolds1987} established the concept of boids for the individuals in a swarm. Reynold\textquoteright  \ s study isolated the well-known set of three simple fundamental forces - cohesion, separation and alignment - which when applied to individuals allows a vast range of complex behaviors at swarm level. However, while stepping away from the animal behavior grounds, Reynolds also intended to keep plausibility; thus, the resultant \textit{animats} still kept the vision-based neighboring mechanism of the biologic counterparts of artificial agents (i.e. boids). 

Subsequent studies proposed various refinements of the forces. Heppner and Grenader~\cite{Heppner1990} considered homing, velocity regulation and interaction as the three governing rules, while later Bajec and colleagues~\cite{Bajec2005} proposed attraction, repulsion and polarisation. Reynolds too continued with the refinement of models of individual behavior, with a series of studies that introduced enhanced definitions of the rules~\cite{Reynolds1999,Reynolds2006}. In addition to rules refinement, various setups for their parameters have been also investigated, mostly from an optimisation point of view. Various authors~\cite{Chen2006,Mammen2012} proposed methods for optimising the parameters of the boids rules, in order to obtain swarms with higher behavioral stability with respect to certain designer purposes. Also, aspects related to emergence and evolution of leadership within swarms have been intensely investigated in order to understand the underpinnings of, and further design and model, coherent collective behaviors~\cite{Karpov2015,Will2016}.

A first example of research derived from boids is the significant interest and success in creating computationally effective animations of collective behavior~\cite{Silva2010}, where boids have been used in motion pictures for special effects~\cite{Phon-Amnuaisuk2016}, as well as in computer games for both motion of crowds and implementation of massively parallel activity of characters~\cite{Mammen2009,Silva2010}. Later, the concept of particle swarm optimisation (PSO) emerged (in-depth reviews of this direction can be found in~\cite{Banks2007,Banks2008}), in which boids in a swarm can be associated to solutions of a problem, and their movements with a search for optimal solutions in the solution space. Thus, guiding swarms\textquoteright  \ motion through the solution space based on various methods/biases can lead to finding the optimal solution ({\it i.e. target}. Refinements of standard PSO have been also proposed in directions that focus on behaviors inspired from certain types of social animals, such as migrating birds optimisation (MGO)~\cite{Duman2012,Pan2014}, or ant colony optimisation (ACO), reviewed in detail in~\cite{Dorigo2005,Dorigo2006}. Another important field emerging from the field of swarm behavior is swarm robotics, reviewed in~\cite{Brambilla2013}, which transfers to real engineering problems the methods and the findings of the research on both theoretical abstract boids and their optimization-related fields.

\rtext{The concept of networking the swarm for improving its dynamics and robustness against adversarial attacks is only in its infancy.} One pioneering work is to reveal the effect of network structure on swarm behavior from an optimization perspective~\cite{liu2014particle,du2015adequate,gao2015selectively}, \rtext{in which the authors found that a heterogeneous network can facilitate information transmission and dissemination.} In this paper, we focus on these two dimensions. First, we propose the idea of networking the boids, where the neighborhood in Boids \rtext{is} defined using a network. Second, we study the robustness of this approach against adversarial attacks, whereby an observer attempts to reverse engineer the factors influencing the dynamics of the swarm through external observations.

\section{Methodology} \label{Section:Methodology}
We perform the investigation by comparing classic boids, which consider neighbors using visual perception, with network-based boids, which consider neighbors using network connectivity. The investigation is conducted in two stages. 

In the first stage, classic boids and network-based boids are simulated successively in the same environment in order to evaluate the resultant collective swarm behavior under various communication mechanisms. The purpose of this direction is to reveal potential advantages and/or disadvantages in the quality of the collective behavior achieved by network-based boids in comparison with classic vision-based boids. The focus here is on the ability of the network-based boids to generate stable swarming behavior, and a number of metrics for swarm behavior quality will be used, as explained later in this section.

In the second stage, classic boids and network-based boids are simulated in the same environment. The collective behavior becomes subjected to observations from an external entity that attempts to learn in real-time the individual boids rules. Thus, a (desirably real-time) learning algorithm to be used by an external AI observer has to be employed. Once the learning algorithm is in place, the focus is on the ability of the same learner to extract the parameters of the individual boids rules, when the swarms operate under different neighbor perception mechanisms.

\subsection{Definition of Boids and Environment}
We define the population of boids as a set $B$ that operates in a bounded \rtext{2-D} space ($S$) defined by a given width ($s_W$) and a given length ($s_L$). Individual behavior of boids in this space is governed by (1) the three classic boids rules \cite{Reynolds1987} - cohesion, separation and alignment - and (2) the neighborhood. We use for investigation classic boids, where neighborhood is defined by vision range and the proposed network-based boids, where neighborhood is defined by network connectivity between them. Thus, the two categories of boids, vision-based and network-based, have a set of common features, which are related to motion rules and position update, and a set of own (distinct) features, which are related to the way the neighborhood is defined.

\subsection{Common Features}
The features common to both types of boids are as follows:
\begin{itemize}
    \item Position ($p$), $p\in S$, is a 2-D coordinate. The initial position of each boid is randomly assigned.
    \item Velocity ($v$) is a 2-D vector representing boid\textquoteright  \ s movement (heading and speed) in a time unit. The initial velocity of each boid is randomly assigned.
    \item Cohesion Velocity ($cohesionV$) of a boid is the velocity calculated based on the center of mass of all boids in its neighborhood.
    \item Alignment Velocity ($alignmentV$) of a boid is the velocity calculated based on the average velocity of all its neighbors.
    \item Separation Velocity ($separationV$) of an boid is the velocity that forces the boid to keep a minimum distance from its neighbors, in order to avoid collision.
    \item Safe Distance ($d_s$) is the minimum Euclidean distance between two boids, and drives the Separation Velocity.
    \item Velocity weights\rtext{~\cite{Reynolds1987}}:
    \begin{itemize}
        \item Cohesion weight ($w_c$): a scaler for the cohesion velocity.
        \item Alignment weight ($w_a$): a scaler for the alignment velocity.
        \item Separation weight ($w_s$): a scaler for the separation velocity.
    \end{itemize}
\end{itemize}

\subsubsection{Cohesion}
The cohesion velocity vector describes the tendency of boids to move towards their neighbors\textquoteright  \  location. The Cohesion Velocity ($cV$) of a boid is the velocity calculated based on the center of mass (average position) of all boids in its neighborhood. For a boid $B_i$, we denote its neighbors as a set $N$. Then, $cohesionV_i$ of boid $B_i$ at time $t$ can be derived from the position of its neighbors as in Equation~\ref{Equation:Cohesion}.
\begin{equation}
    cV_i = \frac{\sum_{j=0}^{|N|} p_j}{|N|} - p_i \;\; \mbox{where} \; i\neq j
    \label{Equation:Cohesion}
\end{equation}
where, $|N|$ is the cardinality of $N$.
 
\subsubsection{Alignment}
The tendency of boids to align with the direction of movement of their neighbors is described by the alignment velocity vectors. Alignment Velocity ($aV$) of a boid is the velocity calculated based on the average heading of all its neighbors. The alignment vector of a boid $B_i$ at time $t$ is derived from the velocities of all its neighbors $N$, as in Equation~\ref{Equation:Alignment}.
\begin{equation}
    aV_i = \frac{\sum_{j=0}^{|N|} v_j}{|N|} - v_i \;\; \mbox{where} \; i\neq j
    \label{Equation:Alignment}
\end{equation}

\subsubsection{Separation}
Separation expresses the tendency of boids to steer away from their neighbors in order to avoid crowding them or colliding with them, and is described by the Separation Velocity ($sV$) vector. Separation velocity of a boid $B_i$ at time $t$ is calculated using its neighbors $N_d$ that are closer than a minimum Euclidean distance $d_s$ called Safe Distance. Equation~\ref{Equation:Separation} steers boids from each-other, in order to avoid collisions.
\begin{equation}
    sV_i = -\sum_{j=0}^{|N_d|} (p_j - p_i) \;\; \mbox{where} \; i\neq j
    \label{Equation:Separation}
\end{equation}

\subsubsection{Velocity Update}
Velocity ($v$) is a vector representing the movement of a boid (heading and speed) in a time unit. Velocity of a boid $B_i$ at time $t$ is updated according to Equation~\ref{Equation:Velocity}.
\begin{equation}
    v_i(t) = v_i(t-1) + w_c \times cV_i(t) + w_a \times aV_i(t) + w_s \times sV_i(t)
    \label{Equation:Velocity}
\end{equation}
where $cV$, $aV$, and $sV$ are normalized vectors.
 
\subsubsection{Position Update} \label{Sec:PosUpdate}
Position $p_i$ of a boid $B_i$ is a 2-D coordinate in the operating space ($p_i \in S$). Based on the velocity calculated in Equation~\ref{Equation:Velocity}, the position at time $t$ of each boid $B_i$ can be updated as in Equation~\ref{Equation:Position}. If the new position of a boid is outside the boundary of space $S$, then reflection rule is applied.
\begin{equation}
    p_i(t) = p_i(t-1) + v_i(t)
    \label{Equation:Position}
\end{equation}

\subsection{Distinct Features - Neighborhood Definition}

\begin{figure}[h]
    \centering
    \begin{subfigure}[b]{0.45\textwidth}
        \includegraphics[width=0.8\textwidth,height=0.5\textwidth]{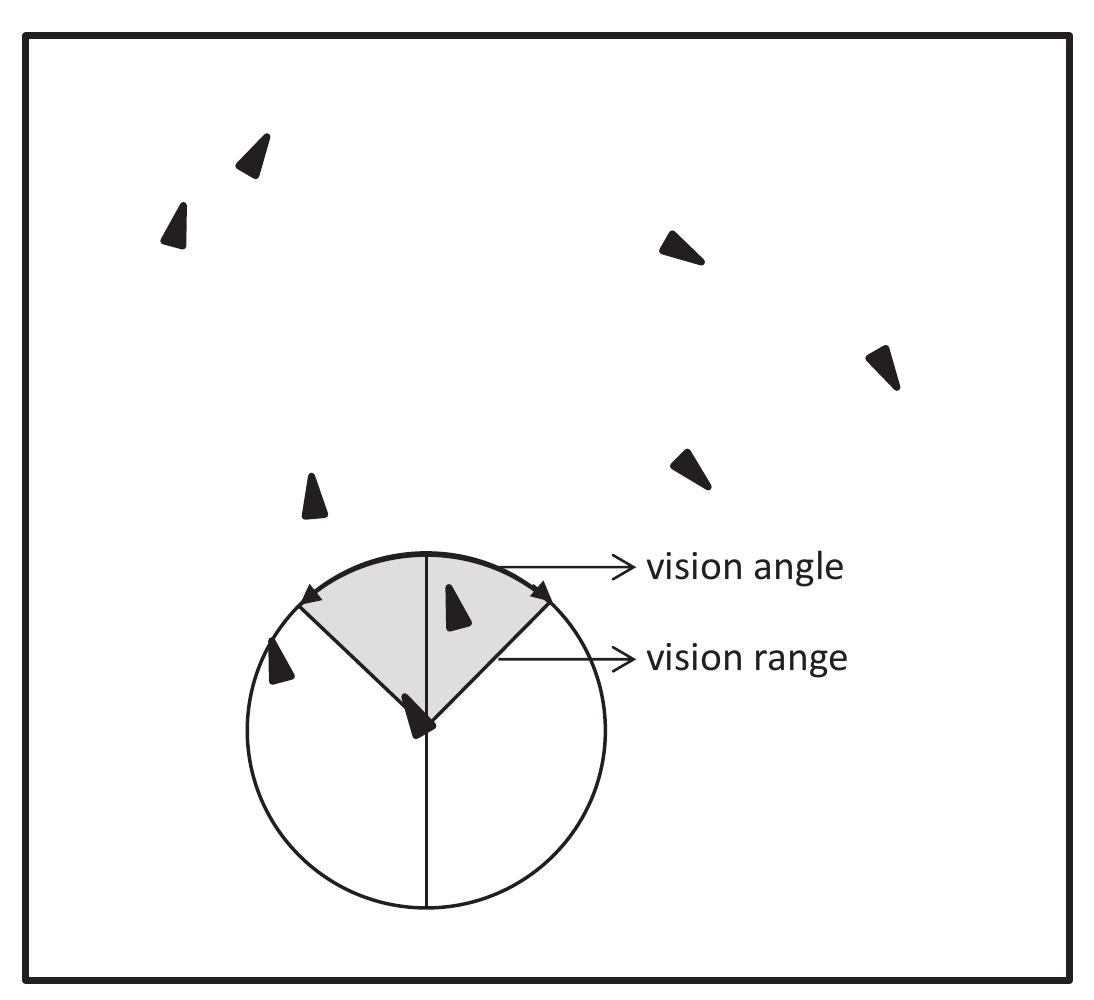}
        \caption{Vision-based neighborhood}
        \label{Fig:NeighVis}
    \end{subfigure}
    \begin{subfigure}[b]{0.45\textwidth}
        \includegraphics[width=0.8\textwidth,height=0.5\textwidth]{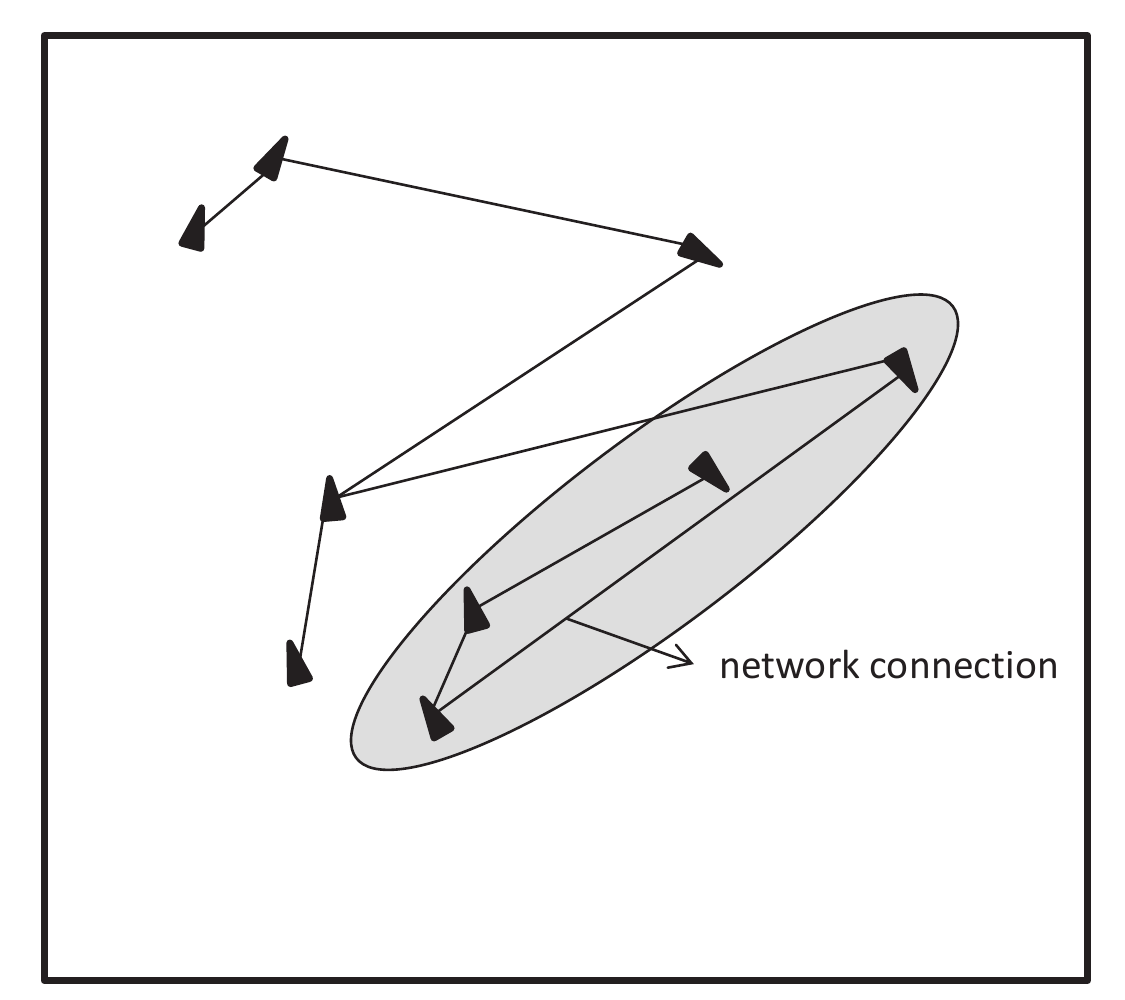}
        \caption{Network-based neighborhood}
        \label{Fig:NeighNet}
    \end{subfigure}
    \caption{Example of classic and network-based neighbouring mechanism}
    \label{Fig:Neighbours}
\end{figure}

\subsubsection{Vision-based Neighborhood}
For the classic boids, we use vision range as the basis for defining boids\textquoteright  \  neighborhoods. Thus, the boids present in current boid\textquoteright  \ s neighborhood are given by the vision range ($vision_r$) and vision angle ($vision_a$). An example of classic vision-based boids is illustrated in Figure~\ref{Fig:Neighbours}.

\subsubsection{Network-based Neighborhood} \label{Sec:Network}

In network-based neighborhood, a boid establishes its neighbours based on a one-hop connectivity with other boids. 
Cohesion and alignment rules are updated based on network connectivity. However, the safe distances between boids are still maintained by Euclidean distance in order to avoid boids collisions in the given space. \rtext{Thus, even though neighborhood is different, the principle is the same, i.e. the safety distance is still considered for the purpose of proper implementation of separation force. In addition, collision avoidance is implemented for all network boids based on this distance, given that in the networked swarm disconnected boids can not be part of the respective neighborhood.} In Figure~\ref{Fig:Neighbours}, we illustrate through a simple example the network-based neighborhood of a boid in a swarm.

\begin{figure*}[h]
    \center
    \begin{subfigure}[b]{0.25\textwidth}
        \includegraphics[width=\textwidth]{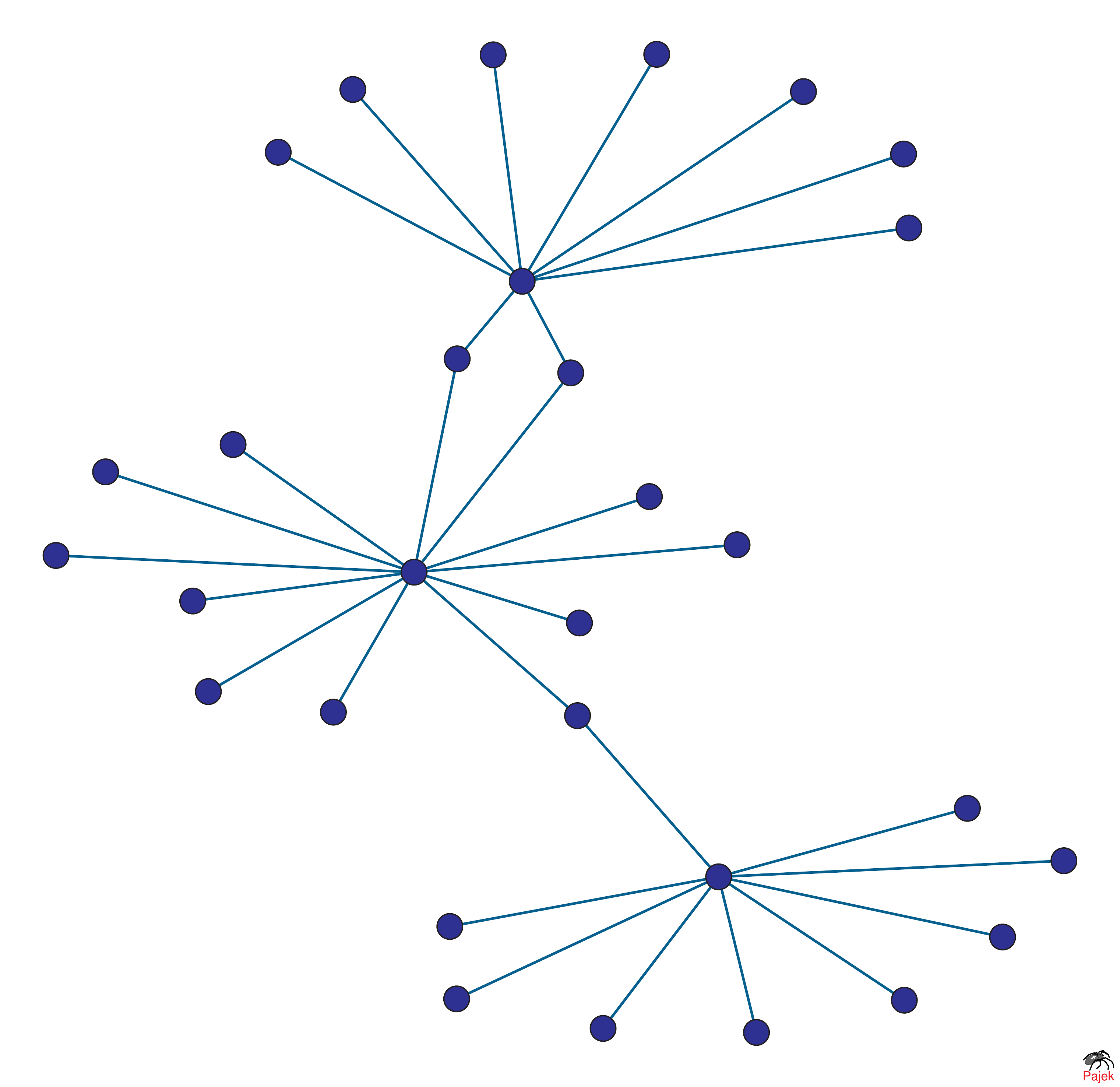}
        \caption{Scale-free network}
        \label{Fig:SFNet}
    \end{subfigure}
    \begin{subfigure}[b]{0.25\textwidth}
        \includegraphics[width=\textwidth]{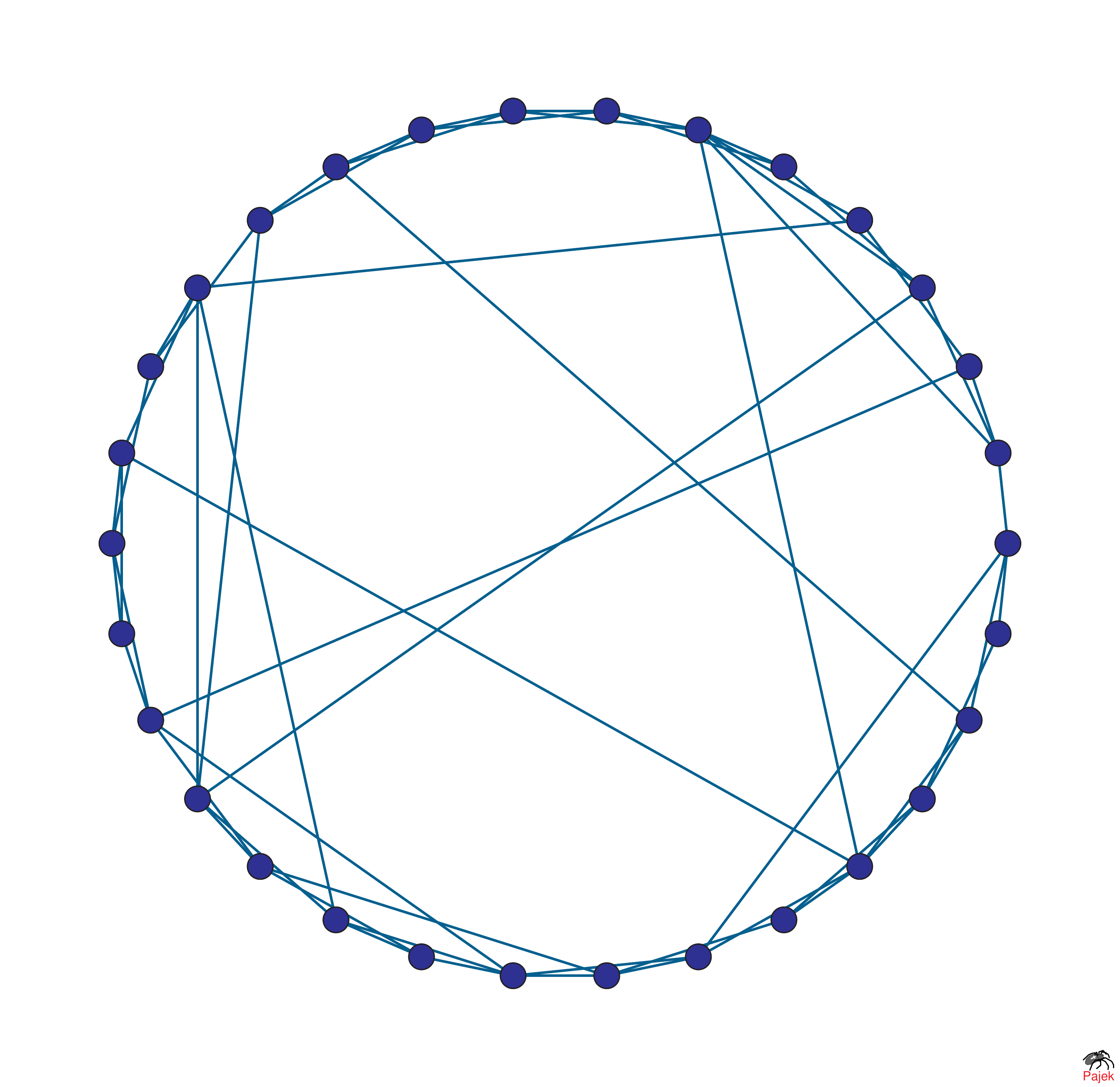}
        \caption{Small-world network}
        \label{Fig:SMNet}
    \end{subfigure}
    \begin{subfigure}[b]{0.25\textwidth}
        \includegraphics[width=\textwidth]{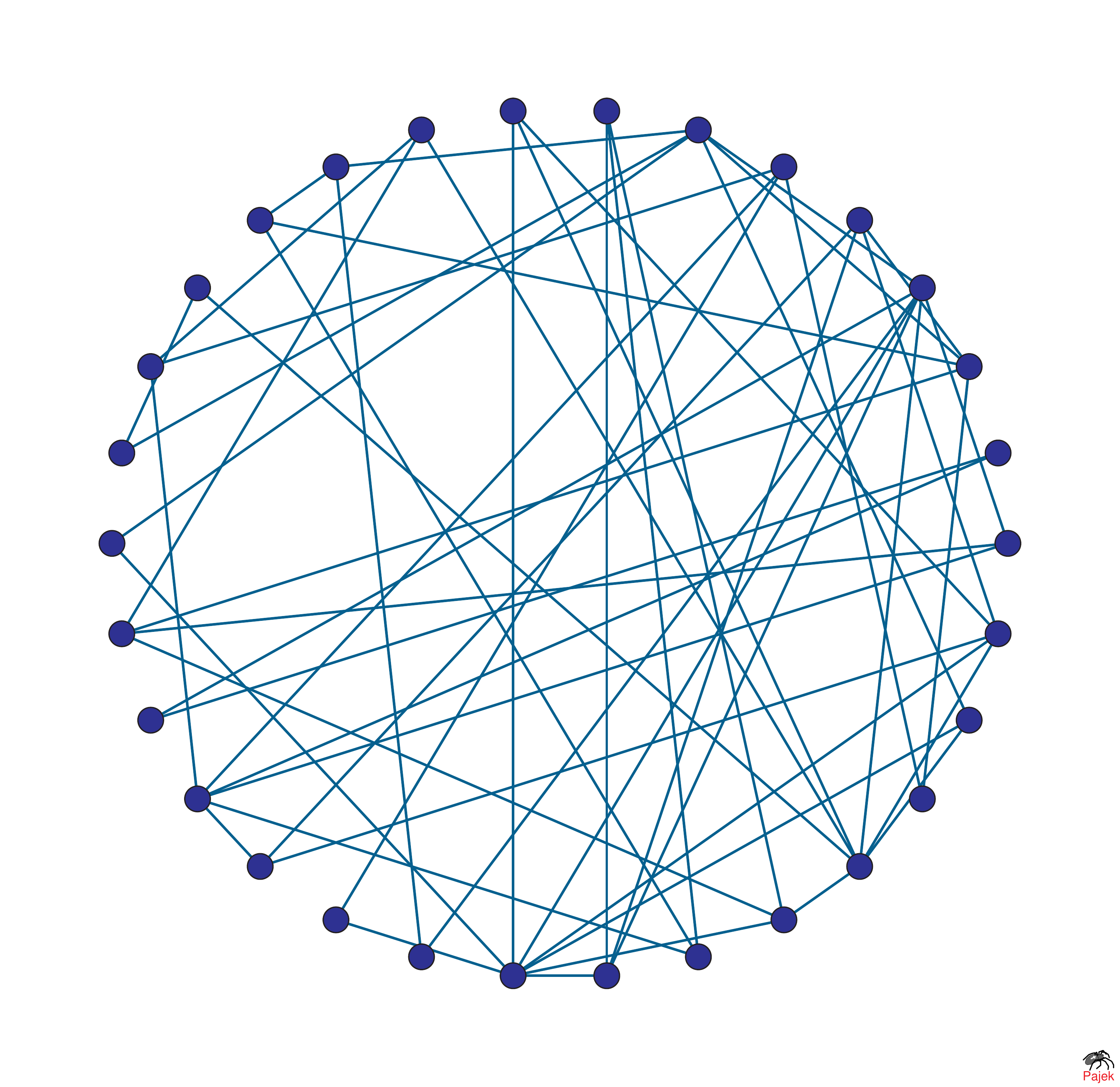}
        \caption{Erd\H{o}s\textendash R{\'e}nyi network}
        \label{Fig:ERNet}
    \end{subfigure}
    \caption{Typical network topologies visualized using Pajek.}
    \label{Fig:Networks}
\end{figure*}

Since neighbors are established based on connectivity, various network topologies, such as scale-free~\cite{Barabasi509}, random~\cite{Erdos1959}, ad-hoc~\cite{perkins2008ad}, and small-world~\cite{watts1998collective}, influence the dynamics. Of these, scale-free, small-world and Erd\H{o}s\textendash R{\'e}nyi are most prominent in the literature. We investigate these topologies, which are illustrated in Figure~\ref{Fig:Networks}.

\subsection{Measuring the Quality of the Swarming Behavior}
We adopt two classic measures used for evaluating swarming: \textit{Order} and \textit{grouping}. 

\textit{Order} represents the average of the normalized boids\textquoteright  \  velocities~\cite{vicsek1995novel}, and can be mathematically formalized as in Equation~\ref{Equation:Order}. According to this equation, all boids are perfectly aligned when the average of their normalized velocities equals 1, and all boids are heading to totally different directions when this equals 0. 
\begin{equation}
    order = \frac{1}{N}\vert \sum_{i=1}^N{v_i}\vert
    \label{Equation:Order}
\end{equation}

\textit{Grouping} evaluates the cohesion of boids~\cite{vicsek1995novel} based on positioning information. Given an attraction range $R_a$, boids that are within half of $R_a$ can be labeled and included in a group. The number of groups revealed by the measurement indicates whether flocking occurs or not. Fewer groups indicate that boids emerge into flocks, while high number of groups show that boids are scattered throughout the experiment space. \rtext{The attraction range value we use for comparing the classic and networked boids is the one used in a very recent paper of Harvey et al. ~\cite{harvey2015application} for measuring the grouping of classic boids.}

\subsection{Adversarial Learning of Observed Swarm Behavior}

The second objective of this paper is to understand the complexity of the learning problem if an adversary attempts to learn and predict in real-time the behaviors of boids for both vision-based and network-based situations, using pure observational measures on the boids. First, the parameters of individual behaviors are learned from past behaviors of the swarm by minimizing the error between the observed behaviors and the outputs from the learning algorithm. Then, the future is predicted by using the learned parameters to project collective behavior onto time.

We assume that the learning algorithm has no prior knowledge of the topological structure underpinning the swarming behavior. 
However, we assume that the learning algorithm knows the boids rules (cohesion, alignment and separation) and update mechanism, but does not know:
\begin{itemize}
    \item the parameters of the boid rules: cohesion weight $w_c$, alignment weight $w_a$, separation weight $w_s$, and safe distance ($d_s$);
    \item the actual vision range ($vision_r$) and vision angle ($vision_a$).
\end{itemize}

Consequently, the learning algorithm uses the basic Boids model as the governing rules for the dynamics and attempts to learn the set of parameters of the Boids model. \rtext{Thus, the aim of the learning algorithm is to approximate the actual values of the parameters, \{$w_c$,$w_a$,$w_s$,$d_s$,$vision_r$,$vision_a$\}, in order to be able to predict the future position of the Boids.}


\subsubsection{The Learning Process} \label{Sec:learningProcess}

Let us assume that the learning algorithm uses observed boids behaviors from time ($t - \Delta t$) to $t$.  With respect to boids simulation, the behaviors observed in the past $\Delta t$ are actually observed from a number of boids updates in $\Delta t$, so they depend on the updating rates of boids. For example, if $\Delta t=60s$ and boids are updated every $50ms$, there are $1200$ updates during $\Delta t$. Therefore, the number of updates actually affect the time constraint on the learning algorithm to observe the Boids and run, that is, the learning algorithm runs within a time window of $\Delta t$.

The outcomes from the learning algorithm are then used to predict future boids behavior \rtext{at ($t + \Delta t'$)}. Then, the learning period of the learning algorithm should be also less than the prediction period, so that $\Delta t < \Delta t'$.

\begin{figure*}[h]
    \includegraphics[width=\textwidth,height=0.55\textwidth]{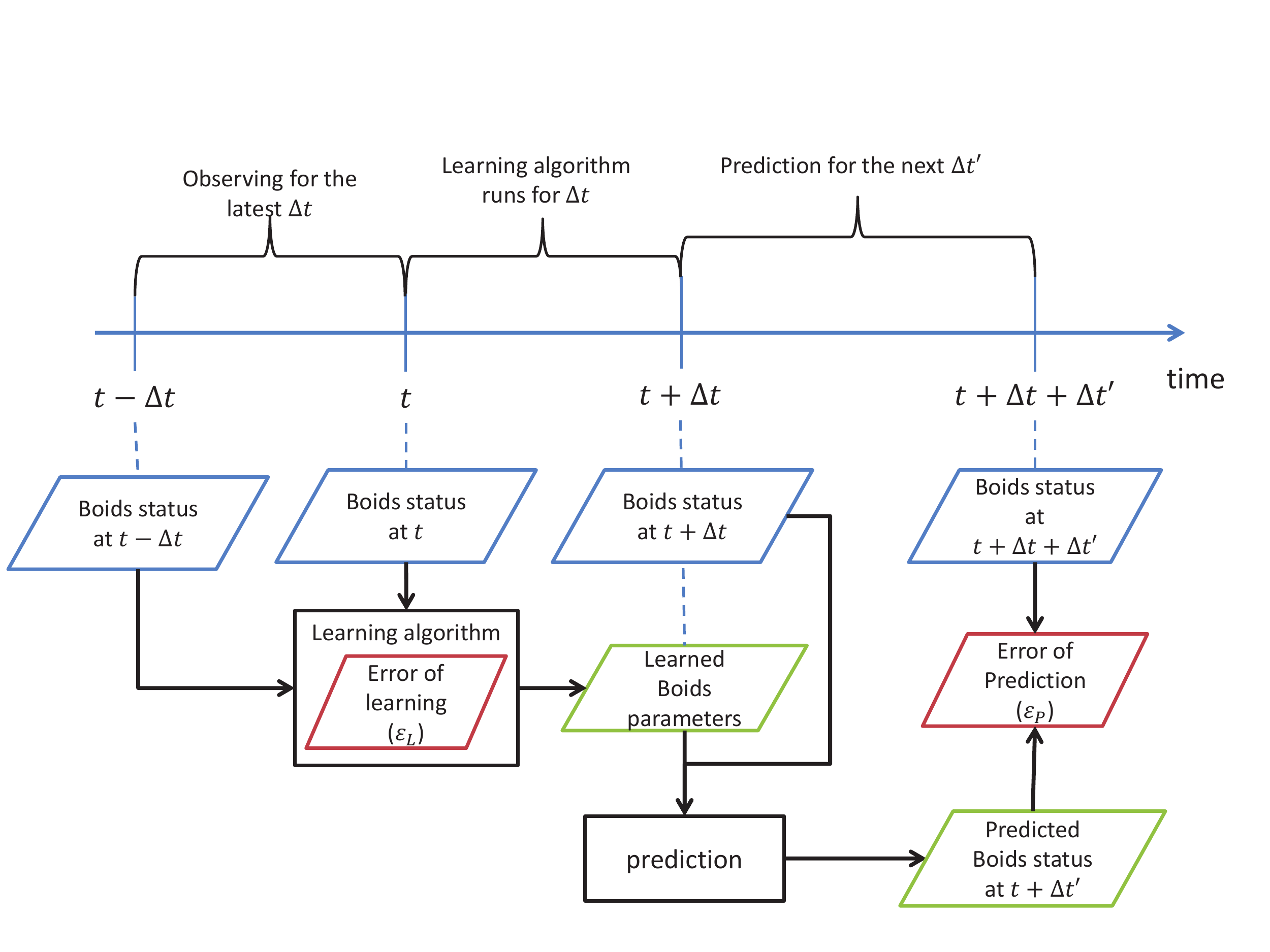}
    \caption{Framework for real-time (on-line) learning}
    \label{Fig:FastLearning}       
\end{figure*}

In general, learning period can be different from the look-back period ($\Delta t$) and prediction period. If learning period is equal or lower than the cumulated look-back period and prediction periods, then true real-time learning and prediction are possible. This case is illustrated in Figure~\ref{Fig:FastLearning}. The trade-off for on-line learning is that learning time must be truncated to the maximum duration that allows real-time learning and prediction, hence the learning algorithm may not reach the end of the learning process and, thus, may produce imperfect results. In the context of this paper, where we investigate the ability of an external observer to infer individual behavioral rules of various boid types in real-time, these imperfections are acceptable for two reasons: (1) they allow real-time learning and prediction and (2) they apply to both vision-based and network-based boids, thus the comparison of learning ability for the two cases is consistent.

\begin{figure*}[h]
    \includegraphics[width=\textwidth,height=0.55\textwidth]{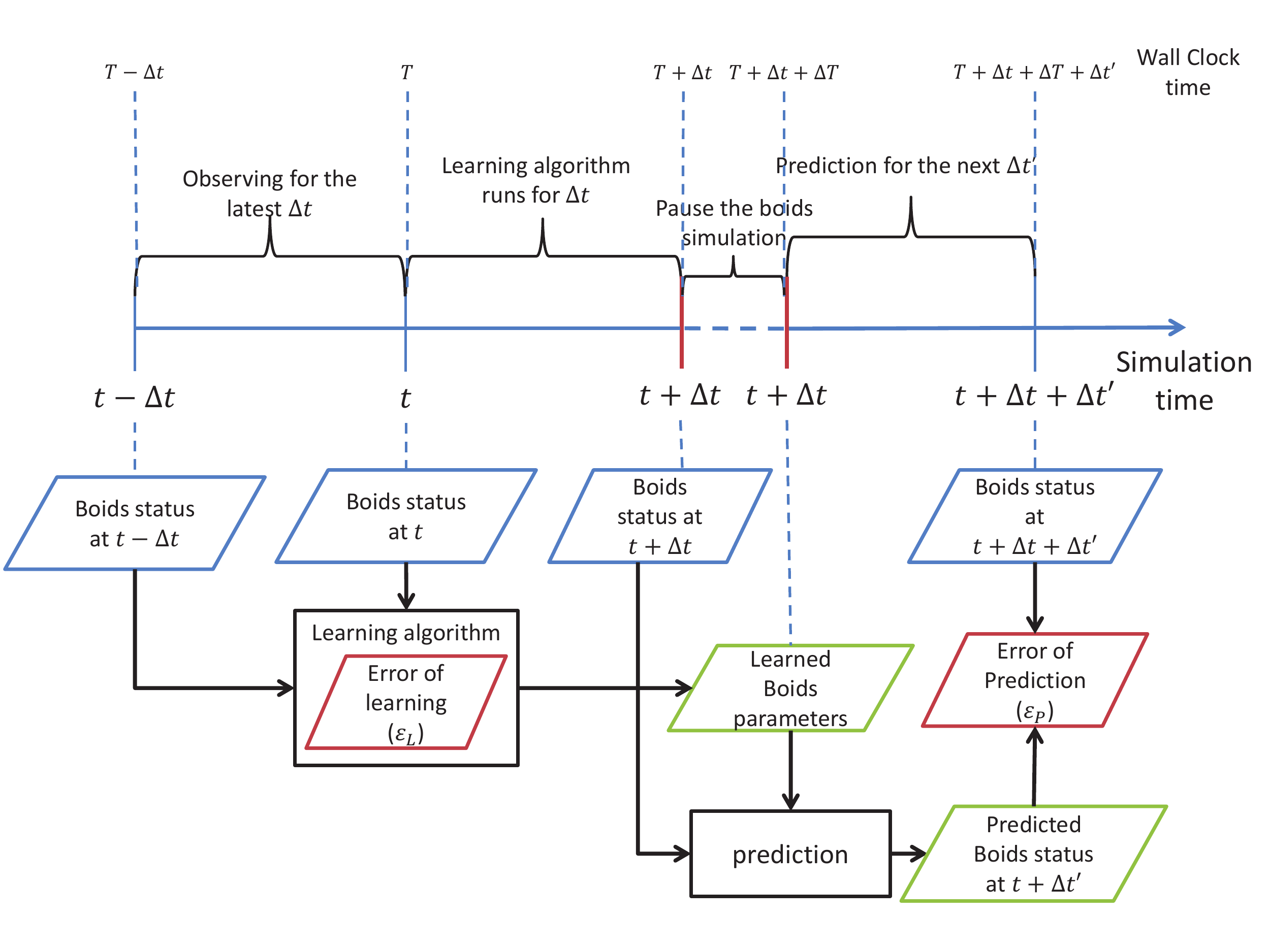}
    \caption{Framework for off-line learning}
    \label{Fig:SlowLearning}       
\end{figure*}

However, in order to provide a solid validation of the learning process, and complete the investigation of the learning ability for the two types of boids, we also consider the case of off-line learning, which allows the learning algorithm to run as long as it needs in order to reach the end of the learning process. This case is obtained by pausing the boids simulation after the learning samples are achieved, performing the entire learning process as needed, and then resuming the boids simulation after the learning is finished. The off-line learning case is illustrated in Figure~\ref{Fig:SlowLearning}.

\subsubsection{The Learning Algorithm}
In this paper, we use Differential Evolution to approximate Boids' parameters based on collected observations. Differential Evolution (DE)~\cite{storn1997differential} is an efficient evolutionary algorithm that searches for global optimal solutions over continuous spaces, and has been used successfully for on-line learning in problems where speed of learning is essential~\cite{Magoulas2004,Das2011}, proving to be superior to classic GAs~\cite{Price2005}.

We use the standard DE proposed by~\cite{storn1997differential}, in which we represent the chromosome using a vector that contains boids\textquoteright  \ parameters as genes, as follows: cohesion weight ($w_c$), alignment weight ($w_a$), separation weight ($w_s$), separation distance ($d_s$), vision range ($vision_r$), and vision angle ($vision_a$). Thus, the $i$'th individual in each generation $G$ can be formalized as in Equation~\ref{Equation:Chromosome}:

\begin{equation}
    x_{i,G}=[w_{c,i}, w_{a,i}, w_{s,i}, d_{s,i}, vision_{r,i}, vision_{a,i}] \;\; i=1,2,...N
    \label{Equation:Chromosome}
\end{equation}
where, $N$ is the total number of individuals in the population. 

The initial population is generated by randomly initializing individuals\textquoteright  \  chromosomes. Then, for each target vector $x_{i,G}$ in each generation, three different individuals, $x_{r1}$, $x_{r2}$, and $x_{r1}$ get randomly selected from the population and are treated as vectors. Based on Equation~\ref{Equation:DEMutation}, a donor vector $v_{i,G+1}$ is produced by the sum of one vector and the weighted difference of other two vectors. As suggested in~\cite{storn1997differential}, $F$ is a scalar factor that controls the amplification of the differential variation ($x_{r2,G}-x_{r3,G}$). The value of $F$ is usually in the interval $[0,2]$.
\begin{equation}
    v_{i,G+1}=x_{r1,G} + F\times(x_{r2,G}-x_{r3,G}).
    \label{Equation:DEMutation}
\end{equation}

Further, a trial vector $u_{i,G+1}$ is generated by combining the elements from both the donor vector ($v_{i,G+1}$) and the target vector ($x_{i,G}$) based on Equation~\ref{Equation:DECrossover}:
\begin{equation}
    u_{ji,G+1} = \left\{
    \begin{array}{l l}
        v_{ij,G+1} & \quad \mbox{if} \; \mathrm{rand}([0,1]) \leq \mathrm{CR} \\
        x_{ij,G} & \quad \mbox{if} \; \mathrm{rand}([0,1]) > \mathrm{CR} 
    \end{array} \right. \quad j=1,2,...,6
    \label{Equation:DECrossover}
\end{equation}
where, $j$ is the index of the parameter inside a vector, and $CR$ is the crossover rate which is within $[0,1]$. The $j$\textquoteright  \ th element in the trial vector is replaced by the $j$\textquoteright  \ th element in the donor vector if a generated uniform random number is less than or equal to the crossover rate. Otherwise, the $j$\textquoteright  \ th element of the target vector is used. 

Once the trial vector is evaluated against the target vector, the one with better fitness becomes a member of the next generation $G+1$. The fitness function used here is the distance between boids locations resulting from DE and boids locations obtained from observation, which is, the learning error $\epsilon_L$ as explained in Equation~\ref{Equation:LeraningError}:
\rtext{
\begin{equation}
    \epsilon_L = \sum_{i=1}^{|B|}{\parallel p_i(t) - p'_i(t)\parallel}
    \label{Equation:LeraningError}
\end{equation}
}
where $p_i(t)$ represents boids locations observed at time step $t$, $p'_i(t)$ represents boids locations produced by the learning algorithm at time step $t$, $i$ is the index of a boid, and $|B|$ is the total number of boids.

To generate $p'$, the learning algorithm takes two samples from the observation of boids: $p(t_0)$ and $p(t_1)$ for all boids. $p(t_0)$ is used by the learning algorithm as the initial boids locations directly. Then, the learning algorithm applies the three boids rules with the parameters $X_{i,G}$ to estimate boids locations at $t_1$, starting from $p(t_0)$. In this way, the estimated $p'(t_1)$ can be generated and the difference between $p(t_1)$ and $p'(t_1$) for each boid can be calculated. The vector with smaller $\epsilon_L$ is the better one. Consequently, the DE learner minimizes the learning errors.

As explained earlier in~\ref{Sec:learningProcess}, we treat the case where learning happens off-line in order to validate the learning concept. However, the desired operation mode for the learning algorithm is real-time, where learning and prediction take place on the fly, while the swarm evolves. In this case, the DE learner described above can only generate some acceptable solutions which can produce collective behaviors approximating the actual observed behaviors. Hence, differences ($\epsilon_L$) between the learning outcome and observations may exist. This type of errors can further generate errors in prediction ($\epsilon_P$), which are calculated in the same manner the learning error is, as in Equation~\ref{Equation:LearningPrediction}:
\rtext{
\begin{equation}
    \epsilon_P = \sum_{i=1}^{|B|}{\parallel p_i(t) - p''_i(t)\parallel}
    \label{Equation:LearningPrediction}
\end{equation}
}
where $p''(t)$ represents the future boids locations predicted by the best parameter estimations resultant from the learning algorithm.

\section{Experimental Results and Discussion} \label{Section:ExpResults}

Experiments are designed to evaluate the quality of swarm behavior and the ability of an external learner to infer boids behavior under classic and network-based neighboring mechanisms. Below, we describe the experimental setup for the two investigations.

\subsection{Setup of Boids Parameters}
In order to proceed with the comparison between classic boids and the proposed network-based boids, we first need to establish the actual parameters for boids and their environment. For ensuring traceability, consistency and validity of results, we searched the literature for classic boids simulation setups with proven results. A very recent study of Harvey et al.~\cite{harvey2015application} presented a simulation of classic vision-based boids which was based on previous studies of Reynolds~\cite{Reynolds1987} and Parker~\cite{Parker2010}. The simulation showed remarkable results in terms of swarming behavior and was well documented. Thus, we adopt the same parameter settings for the classic boids in our study, in order to have a solid baseline for the comparison with network-based boids. Table~\ref{Tab:ClassicParameters} summarizes the simulation parameters and the parameters of the classic boids.

\begin{table}[h]
    \center
    \begin{tabular}{|l|l|l|l}\hline
        Parameter & Variable name & Value  \\ \hline
        \multirow{2}{*}{space size}&$spaceW$ & 1000 units \\ \cline{2-3} 
        & $spaceH$ & 1000 units \\ \hline
        number of boids & $n$ & 100 \\ \hline
        \multirow{3}{*}{velocity weights} & $w_c$ & 0.01\\ \cline{2-3}
        & $w_a$  & 0.125\\ \cline{2-3}
        & $w_s$ & 1\\ \hline
        vision range & $vision_r$ & 50 units \\ \hline
        vision angle &$vision_a$ & $2\pi$ \\ \hline
        separation distance & $d_s$ & 10 units \\ \hline
    \end{tabular}
    \caption{Classic boids: summary of parameter settings}
    \label{Tab:ClassicParameters}
\end{table}

In addition to the parameters presented in the table, we fix the speed of boids to a constant value of 1~unit/iteration throughout the whole simulation, and we randomly assign boids\textquoteright  \  initial positions and orientations.

Further, we establish the parameters for network-based boids, corresponding to the three types of topologies to be investigated. Table~\ref{Tab:NetworkParameters} summarizes these parameters.

\begin{table}[h]
    \center
    \begin{tabular}{|l|l|l|l}\hline
        Topology type & Model parameter & Value  \\ \hline
        \multirow{2}{*}{scale-free network}& initial number of nodes & 6 \\ \cline{2-3} 
        & rewire probability & 0.05 \\ \hline
        small-world network & connection probability & 0.1 \\ \hline
        Erd\H{o}s\textendash R{\'e}nyi network & number of edges & 300 \\ \hline
    \end{tabular}
    \caption{Network-based boids: Summary of parameter settings}
    \label{Tab:NetworkParameters}
\end{table}

For the learning algorithm, the six genes in the chromosome are randomly initialized within a given range, as in Table~\ref{Table:ChromosomeValueRange}.
\begin{table}[h]
    \center
    \begin{tabular}{|l|c|c|l|c|c|}\hline
        Parameter & min & max & Parameter & min & max\\ \hline
        $w_c$ & 0 & 1 & $d_s$ & 0 & 1 \\ \hline
        $w_a$ & 0 & 1 & $vision_r$ & 10 & 150\\ \hline
        $w_s$ & 0 & 1 & $vision_a$ & $\frac{\pi}{2}$ & $2\pi$\\ \hline
    \end{tabular}
    \caption{Value ranges of the genes in DE chromosome}
    \label{Table:ChromosomeValueRange}
\end{table}

\subsection{Evaluating Swarm Behavior} \label{Sec:EvalSwarmBehav}
All simulations run for 10000 time-steps to allow the boids to stabilize in a consistent swarm behavior.  Boids are initialized at random locations at the beginning of the simulation. We conducted 10 runs for each communication method. In each run, the initial locations of boids are randomly initialized. 

\begin{figure}[h]
    \center
\includegraphics[width=4cm,height=4cm]{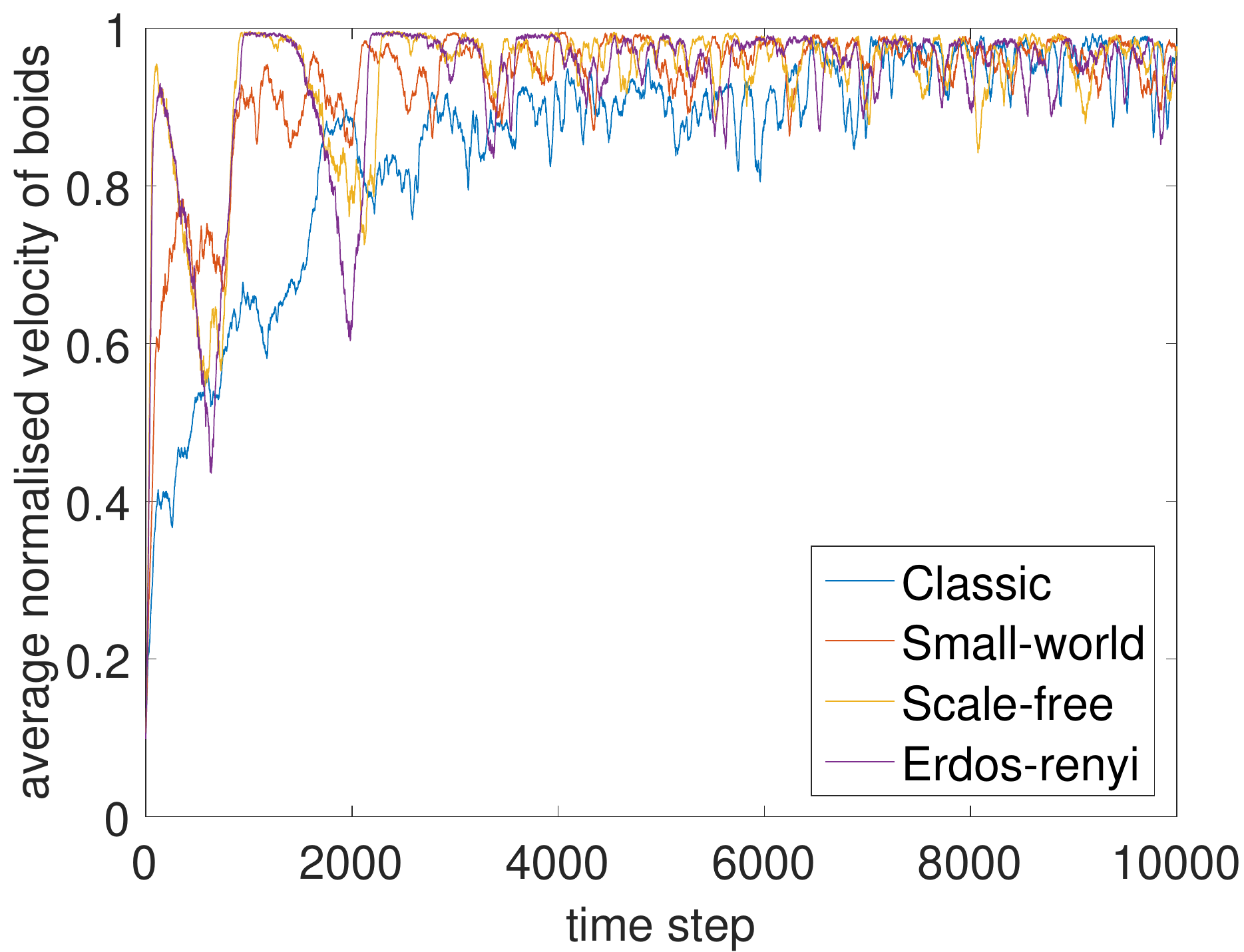}
\includegraphics[width=4cm,height=4cm]{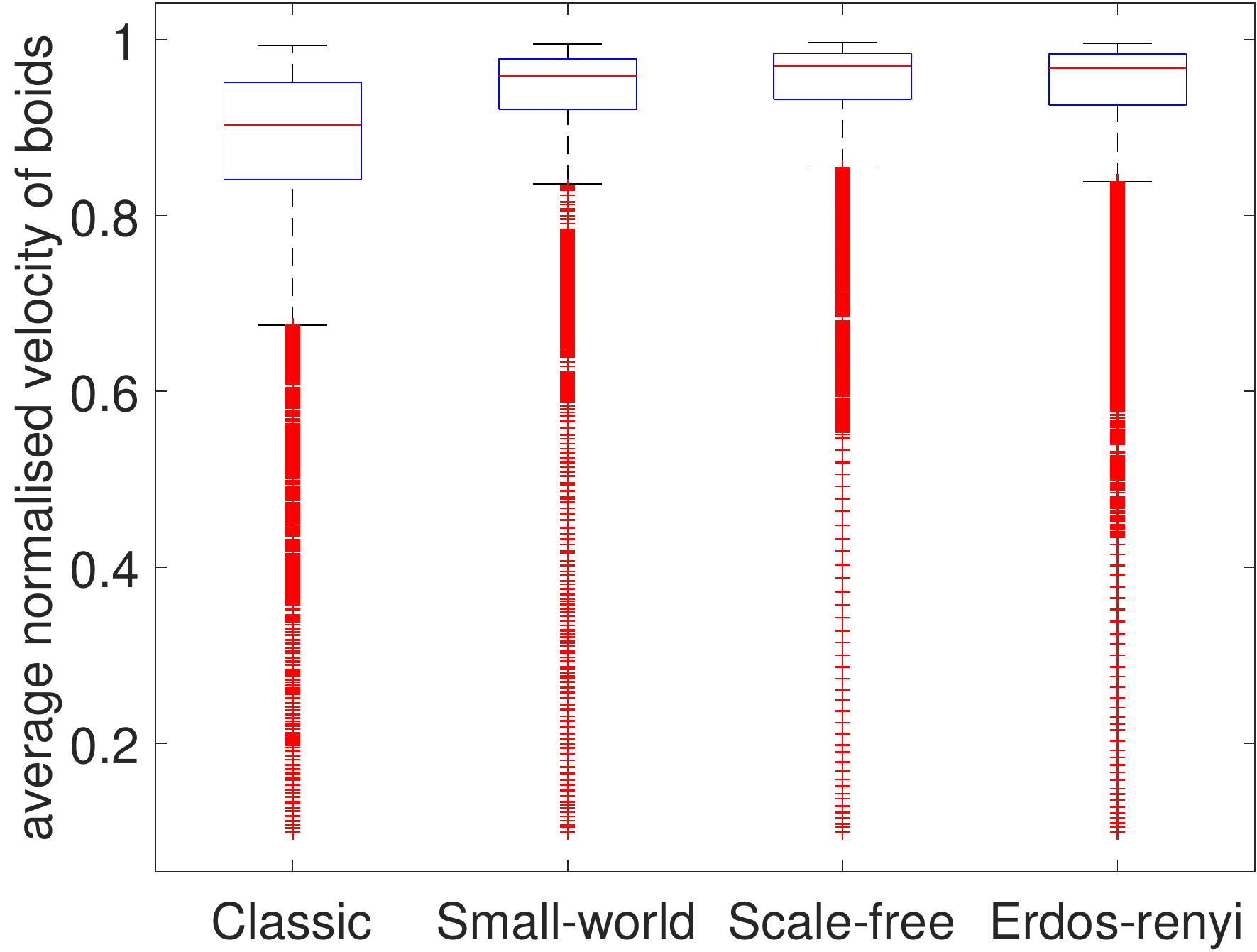}
    \caption{``order'' metric for all boid types from 10 runs.  \rtext{Figure on left shows the ``order" metric over time. Figure on right summarized the metric over the whole simulation period.}}
    \label{Fig:Order}
\end{figure}

Figure~\ref{Fig:Order} shows the \textit{order} metric for boids with different communication methods. It can be observed in Figure~\ref{Fig:Order} that all types of network-base boids converge from the initial states towards high level values of the order metric more rapidly than the classic boids. It can be seen that swarm behavior, from the point of view of order metric, becomes consistent after around 5000 time-steps. There are some noticeable periodic upside-down spikes shown in the figure, which are caused by the reflection rules described earlier in the paper, in Section~\ref{Sec:PosUpdate}. When ordered boids hit the boundary of the given space, they start readjusting their headings, fact that causes a sudden drop in the resultant order value. However, \rtext{results show that network-based boids are able to recover from the order drop, and the evolution of the order measure shows faster convergence of network-based boids, despite the large spikes in the early stages.}

\rtext{In Figure~\ref{Fig:Order}, we illustrate a statistical summary of the order metric corresponding to the whole simulation period (from $t=0$ to $t=10000$). This figure shows again that all three types of network-based boids have better order values than classic boids. The mean values of network-based boids are all close to ``1'', while classic boids have a mean value visibly lower. In addition, the bottom 5 percentile of classic boids order is also lower than all types of network-based boids. Based on the two sample t-test results, the mean order value of classic boids is statistical different from the mean order values of all three network based boids at the significant level of $5\%$.}

\begin{figure}[h]
    \center
\includegraphics[width=4cm,height=4cm]{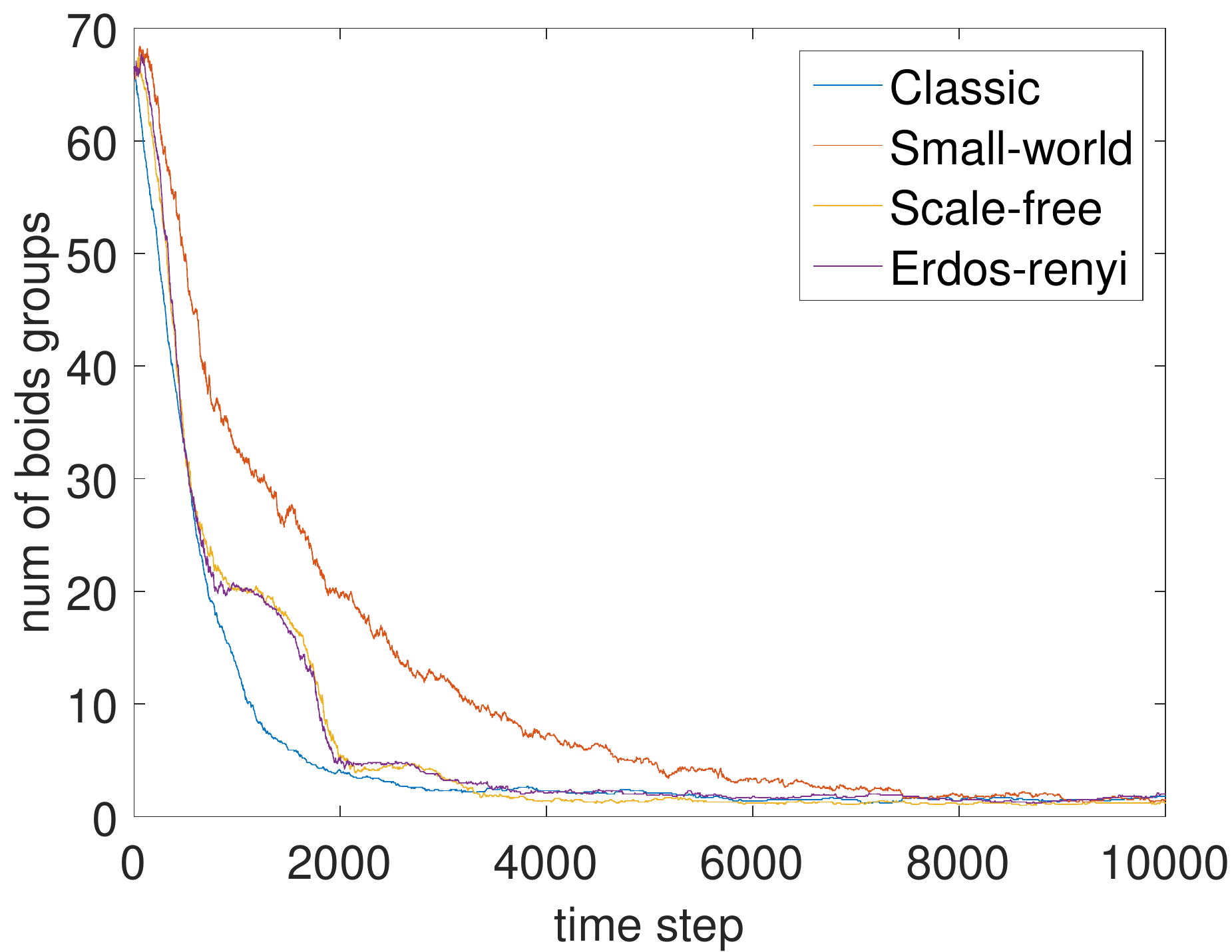}
\includegraphics[width=4cm,height=4cm]{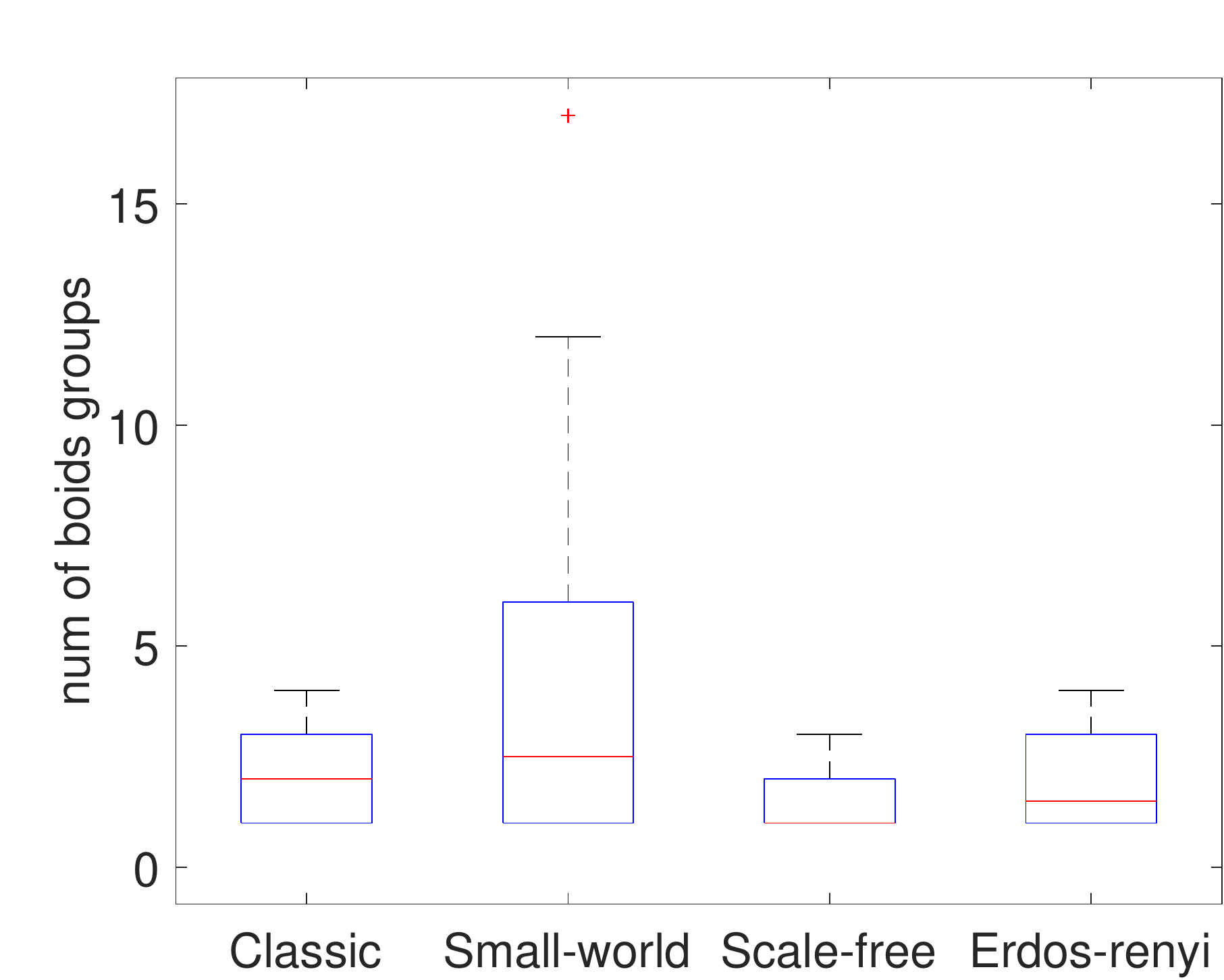}
    \caption{``grouping'' metric for all boid types from 10 runs.  \rtext{Figure on left shows the ``grouping" metric over time. Figure on right summarized the metric over the whole simulation period.}}
    \label{Fig:Group}
\end{figure}

The values of grouping metric are illustrated in Figure~\ref{Fig:Group}. It can be seen in Figure~\ref{Fig:Group} that all boid types, both classic and network-based, achieve high quality swarm behavior from the grouping point of view from around 8000 time-steps. However, the grouping behavior can be considered consistent from around 4000-5000 iterations for all boids except \rtext{Small-world} type, which converges slightly slower. For consistency with the results of order metric, we illustrate in Figure~\ref{Fig:Group} the grouping metric summary at 5000 iterations, even though \rtext{Small-world} type is expected to have a worse summary compared to the other three. It can be seen that in terms of mean value, network-based boids perform visibly better than classic boids, except \rtext{Small-world} type, which is actually almost at the same mean with classic boids, and only deviation slightly higher. 

Overall, the internally-focused investigation shows that network-based boids can perform overall better than classic boids in terms of convergence to a stable swarm behavior. We conjecture that these results can be explained by the very nature of the neighborhood. For network-based boids, neighborhoods of each agents remain constant over the simulation, since network connections to other boids are not changing over time. In addition, if the network has only one cluster, or at least a giant component, then a diffusion phenomenon exists through the network, which facilitates faster convergence. If one boid $B_i$ is connected to a number of neighbors, and these neighbors are further connected to their neighbors, then all boids, including the $n-th$ order neighbors, carry an influence of the boid $B_i$. Aggregating the influences of all boids on each-other, both direct and through neighbors of various orders, and considering that these connections are constant in time, then we can suggest that the bond between network-based boids is tighter than that of the classic vision-based boids. We note that vision based boids do not have fixed neighbors. For a vision-based boid $B_i$, the other boids enter and exit its visual range continuously, hence the neighborhood composition is variable, and subsequently the bonds to other boids are lower and the diffusion of influence throughout the group does not manifest.

\subsection{Evaluate Learning by External Observer}

In order to evaluate the capability of an external learner to infer the individual rule parameters of boids from the observed collective behavior, we perform the learning in both off-line and on-line modes. First, the off-line mode will allow us to evaluate the learning algorithm, which can operate without time constraints, and hence reveal the maximal learning capabilities achievable. Then, the on-line mode will ensure plausibility with potential real-world applications, where an external observer must act in real-time so that its learning outcome is meaningful. In this mode, the learning algorithm must be constrained to a fixed and short learning window, hence learning may be imperfect. However, this mode will demonstrate the realistic robustness of the swarm against external observers and the afforded opportunity by the observer to predict the dynamics.

For both learning modes the procedure consists of acquiring sets of samples from the observation of the whole swarm and then feeding these into the learning algorithm to learn individual boids parameters. In the DE learning algorithm, the population size is set to 100, and the number of generations is set to 300. Also, in order to ensure statistic validity of the DE, which is a heuristic method, we run each experimental instance for 30 times (i.e. with 30 seeds corresponding to the random initialization of individuals\textquoteright  \  genes in the first generation).

\subsubsection{Off-line Learning} \label{Sec:OfflineLearning}
We investigate the off-line learning capabilities in two cases. In the first case, observation takes place at the beginning of swarm simulation, when swarm behavior is not yet established. In the second case, observation takes place at time-step 5000, when swarm behavior already becomes consistent.

For the first case, observation samples are taken at the very beginning of boids simulation, using four different sampling intervals (i.e. look-back periods): 2, 4, 8 and 16 time steps. Running experiments using observation windows of different lengths allows us to investigate the impact of different look-back periods on the performance of our learning algorithm.

\begin{figure*}[h]
    \center
    \begin{subfigure}[b]{0.24\textwidth}
        \includegraphics[width=\textwidth,height=3.2cm]{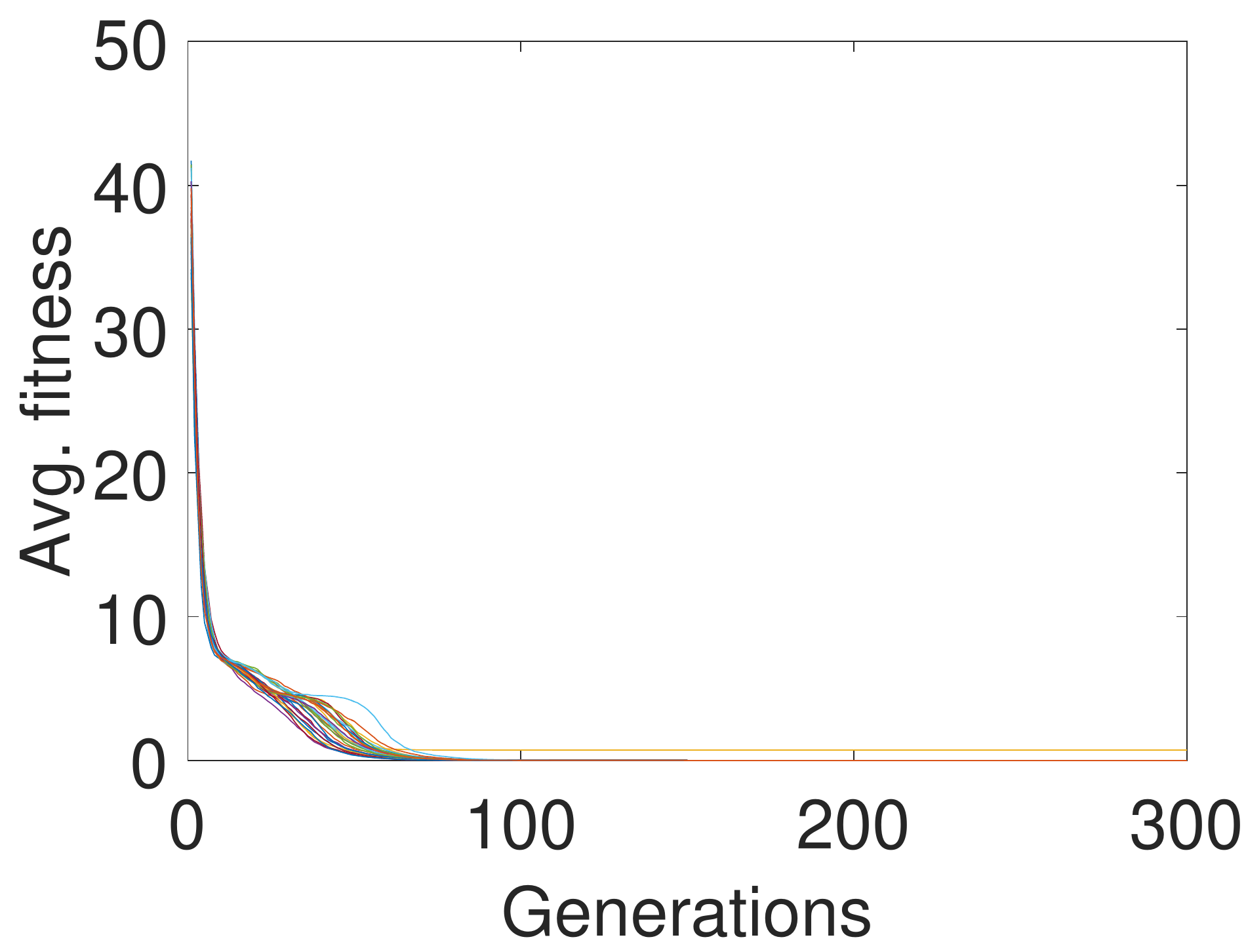}
        \caption{Classic Boids}
        \label{Fig:Classic2step}
    \end{subfigure} 
    \begin{subfigure}[b]{0.24\textwidth}
        \includegraphics[width=\textwidth,height=3.2cm]{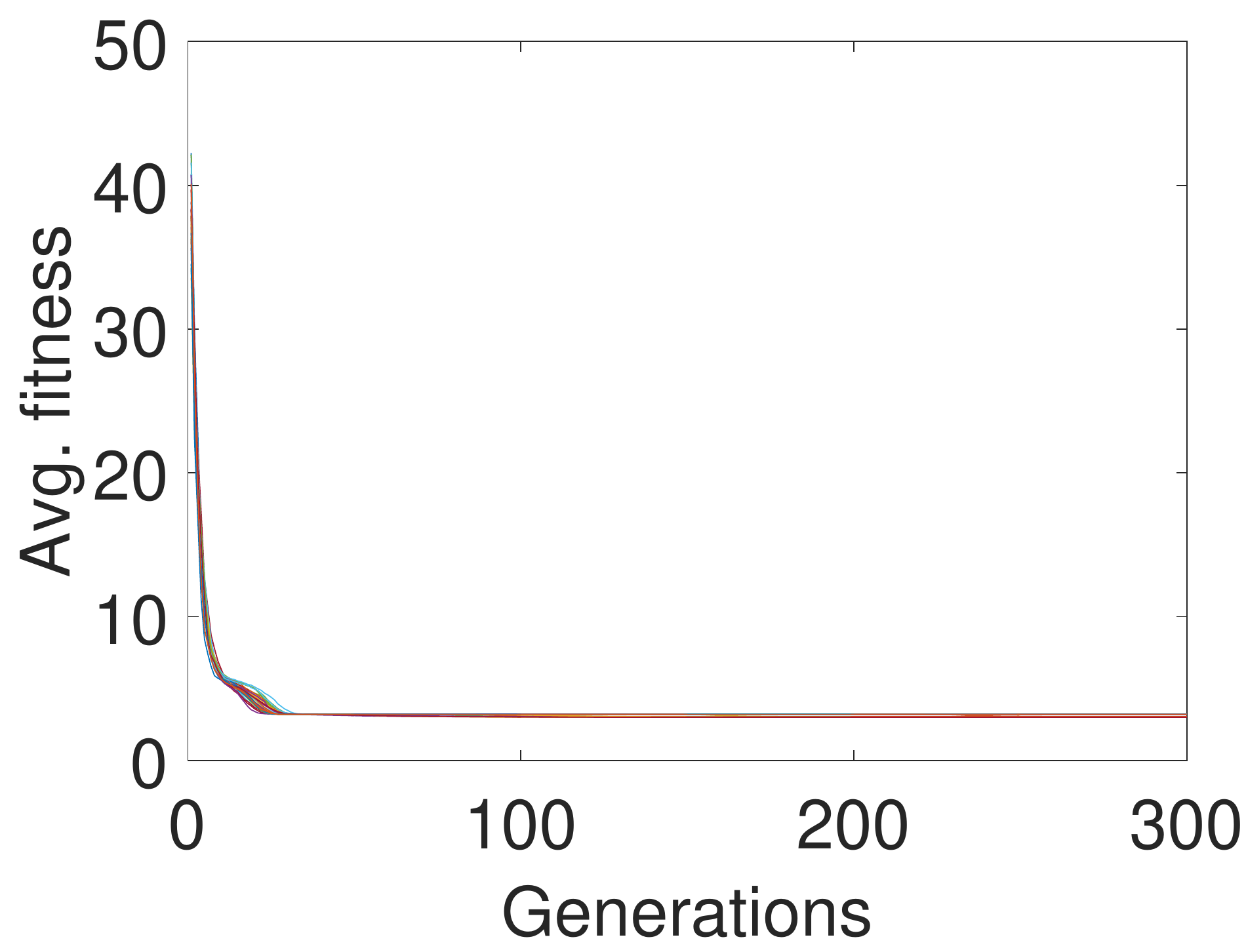}
        \caption{Erd\H{o}s\textendash R{\'e}nyi network}
        \label{Fig:ERNet2step}
    \end{subfigure} 
    \begin{subfigure}[b]{0.24\textwidth}
        \includegraphics[width=\textwidth,height=3.2cm]{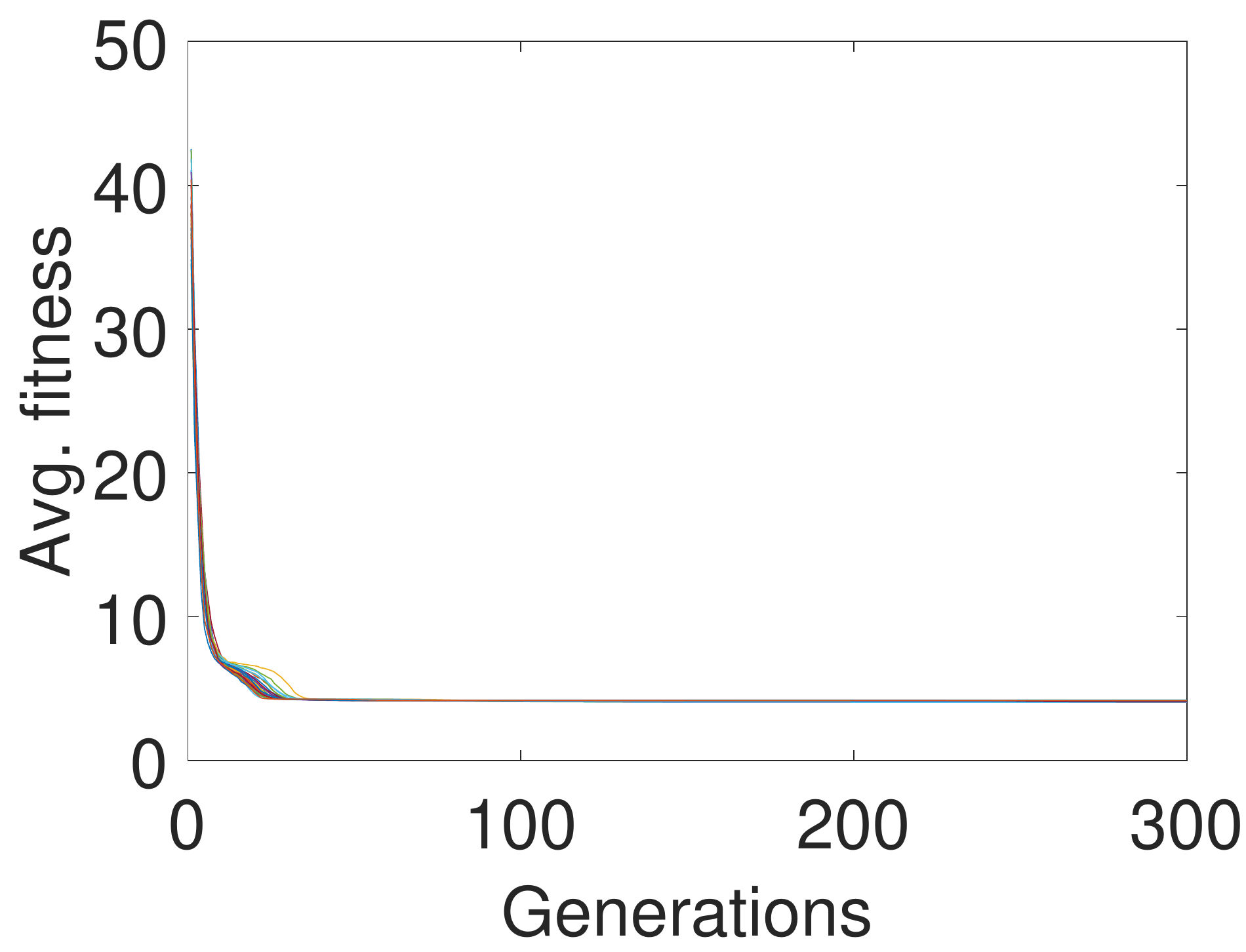}
        \caption{Scale-free network}
        \label{Fig:ScaleFree2step}
    \end{subfigure} 
    \begin{subfigure}[b]{0.24\textwidth}
        \includegraphics[width=\textwidth,height=3.2cm]{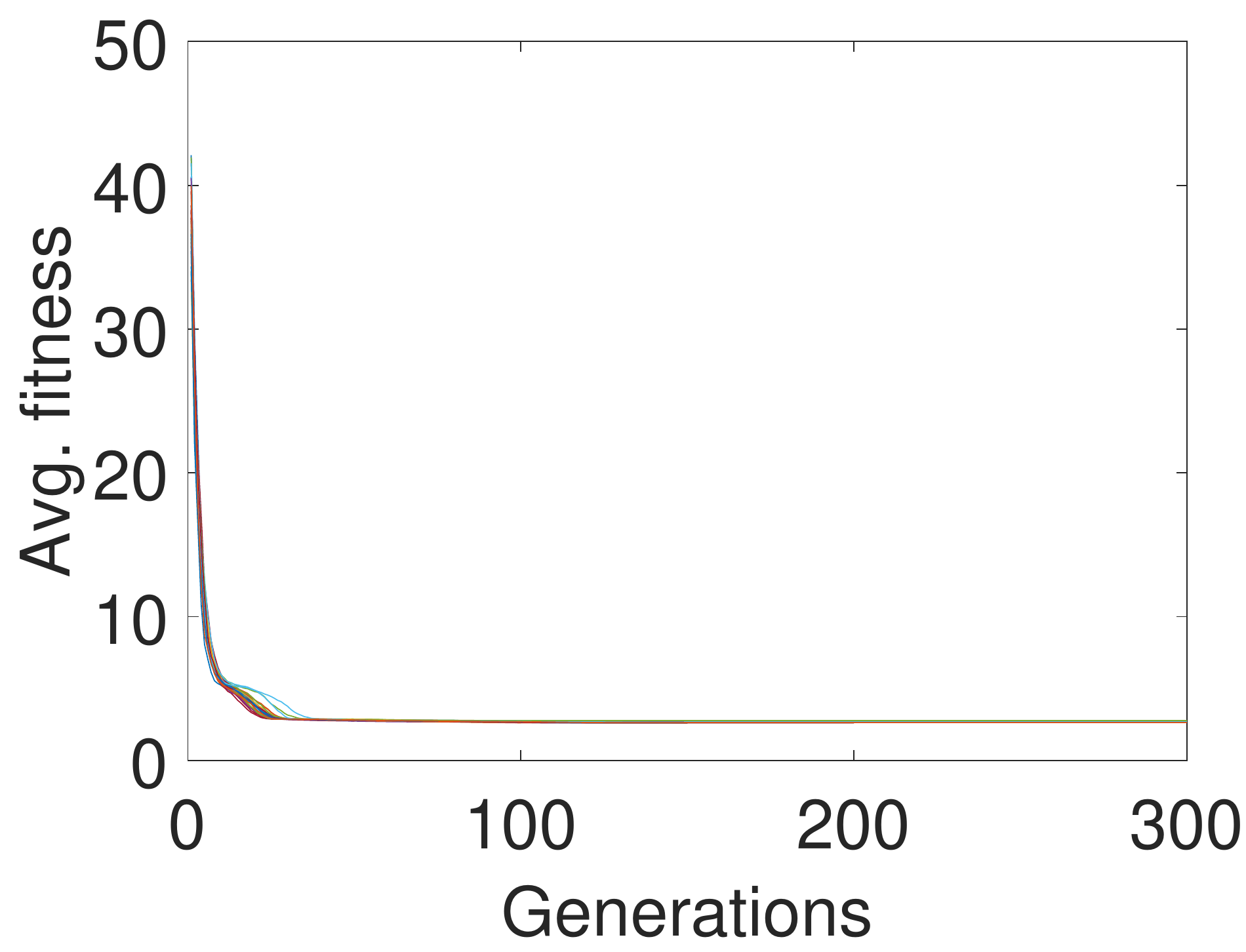}
        \caption{Small-world network}
        \label{Fig:SmallWorld2step}
    \end{subfigure}
    \caption{Average fitness evolution of 30 runs \rtext{(presented by colored curves)} for all boid types when the observation period is 2 iterations (from $t=0$ to $t=1$).}
    \label{Fig:DEEvo2Step}
\end{figure*}

\begin{figure*}[h]
    \center
    \begin{subfigure}[b]{0.24\textwidth}
        \includegraphics[width=\textwidth,height=3.2cm]{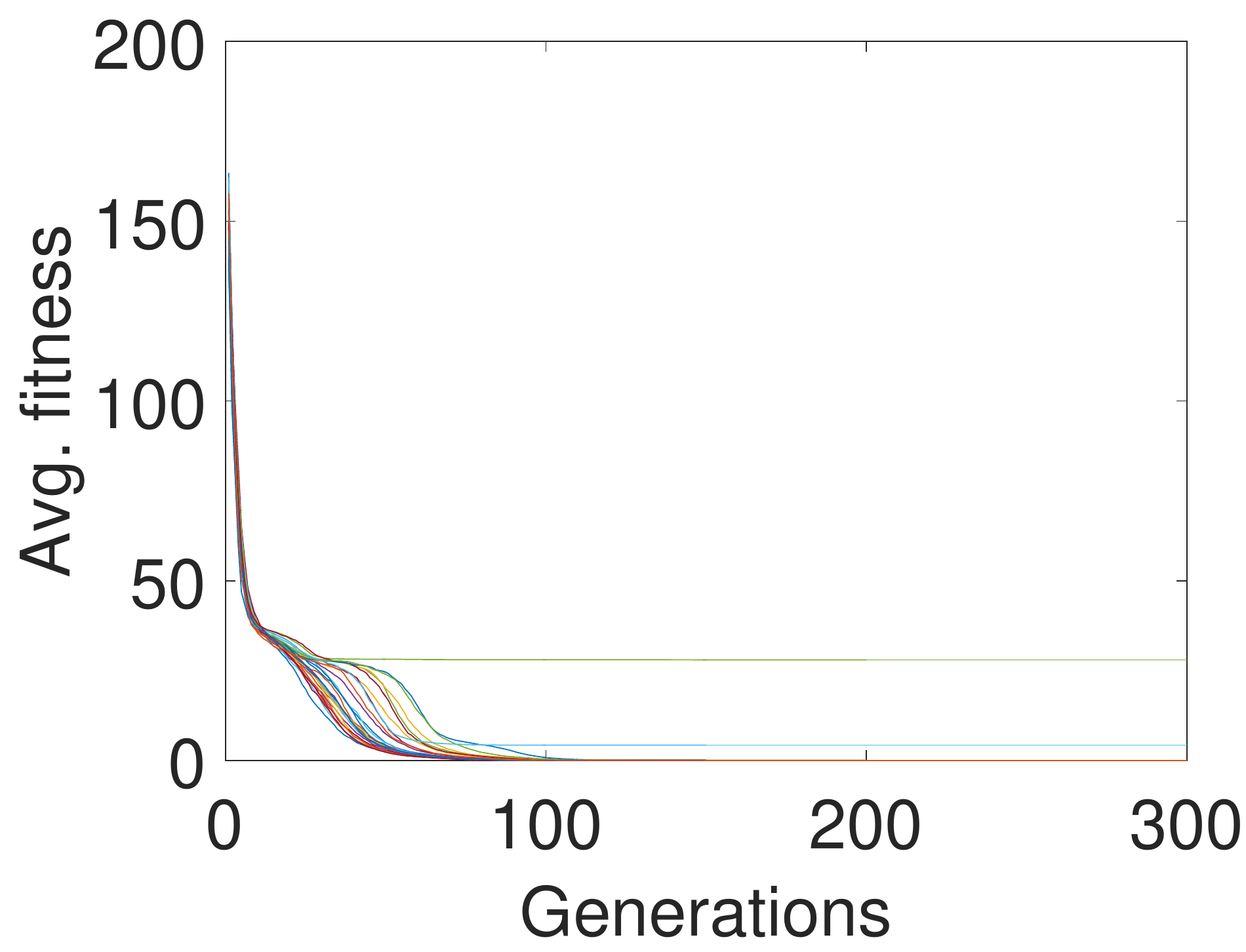}
        \caption{Classic Boids}
        \label{Fig:Classic4step}
    \end{subfigure} 
    \begin{subfigure}[b]{0.24\textwidth}
        \includegraphics[width=\textwidth,height=3.2cm]{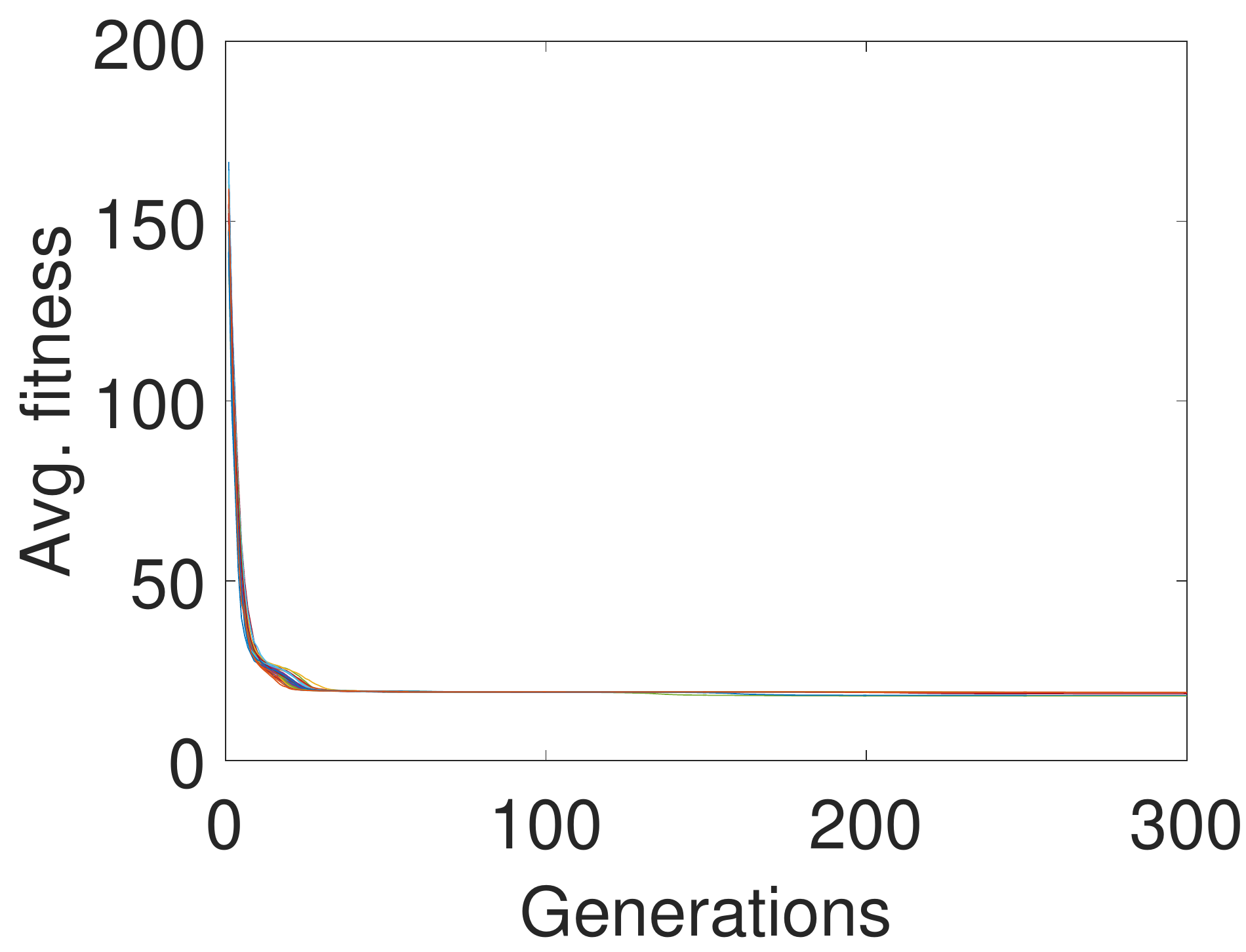}
        \caption{Erd\H{o}s\textendash R{\'e}nyi network}
        \label{Fig:ERNet4step}
    \end{subfigure} 
    \begin{subfigure}[b]{0.24\textwidth}
        \includegraphics[width=\textwidth,height=3.2cm]{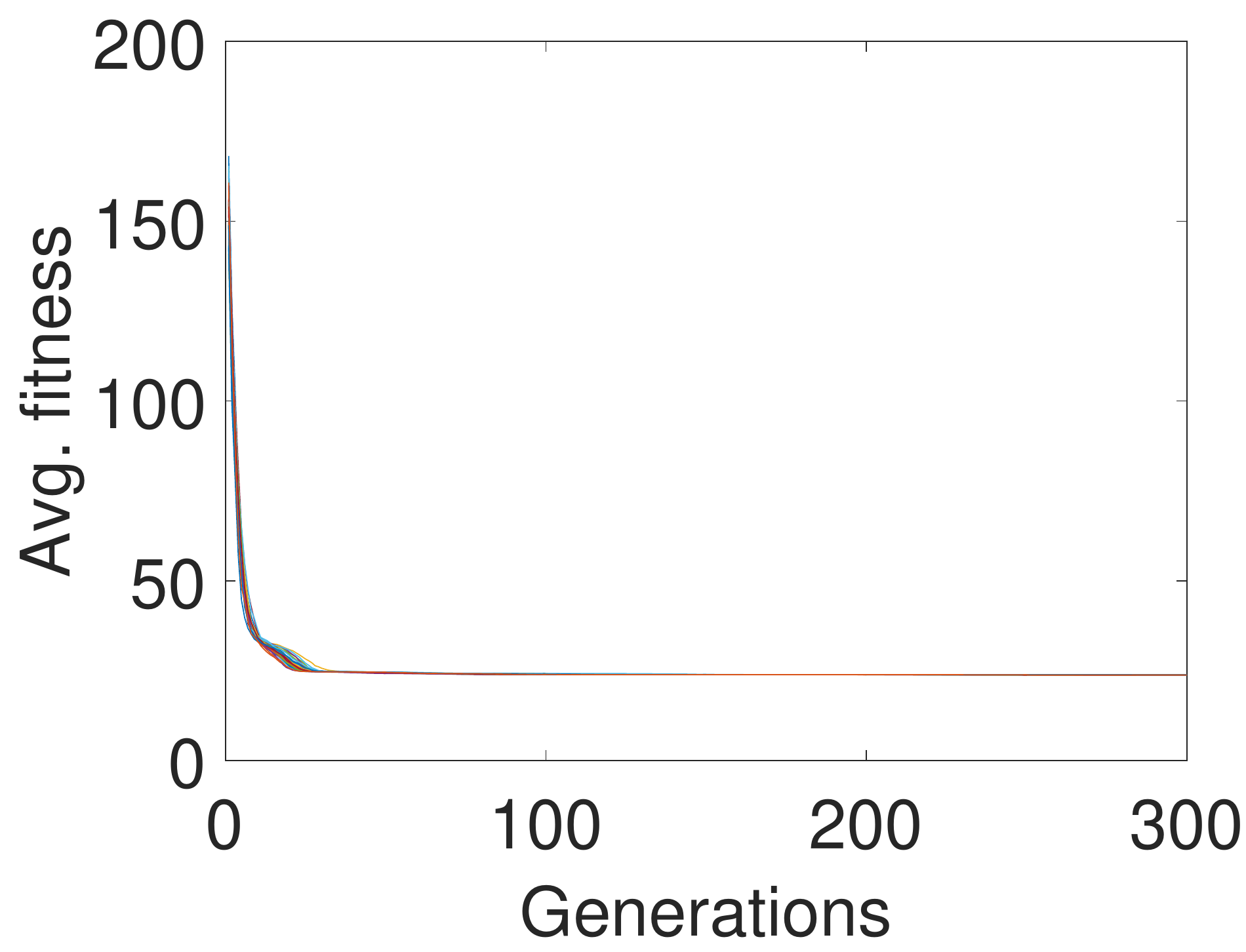}
        \caption{Scale-free network}
        \label{Fig:ScaleFree4step}
    \end{subfigure} 
    \begin{subfigure}[b]{0.24\textwidth}
        \includegraphics[width=\textwidth,height=3.2cm]{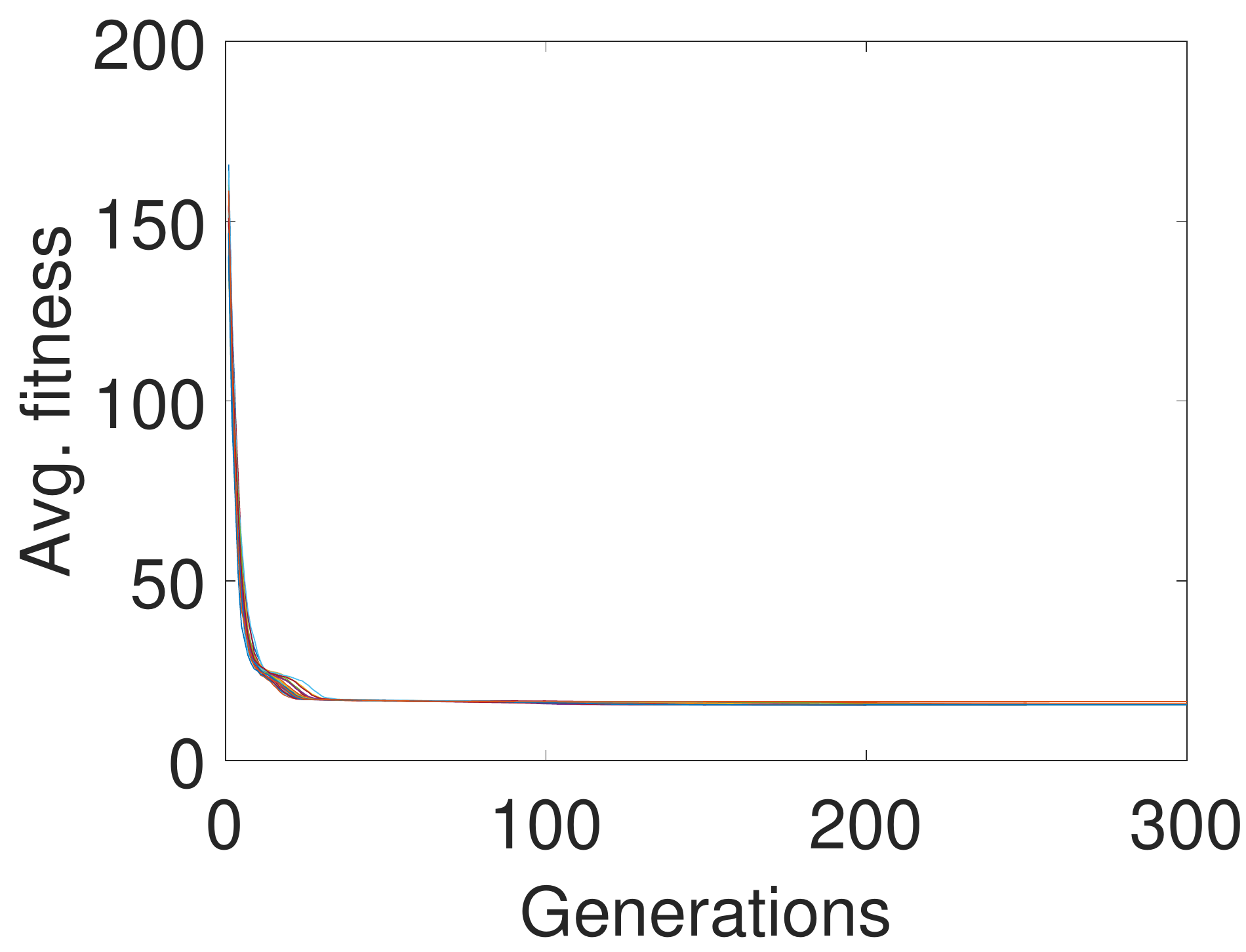}
        \caption{Small-world network}
        \label{Fig:SmallWorld4step}
    \end{subfigure}
    \caption{Average fitness evolution of 30 runs \rtext{(presented by colored curves)} for all boid types when the observation period is 4 iterations \rtext{(from $t=0$ to $t=3$).}}
    \label{Fig:DEEvo4Step}
\end{figure*}

\begin{figure*}[h]
    \center
    \begin{subfigure}[b]{0.24\textwidth}
        \includegraphics[width=\textwidth,height=3.2cm]{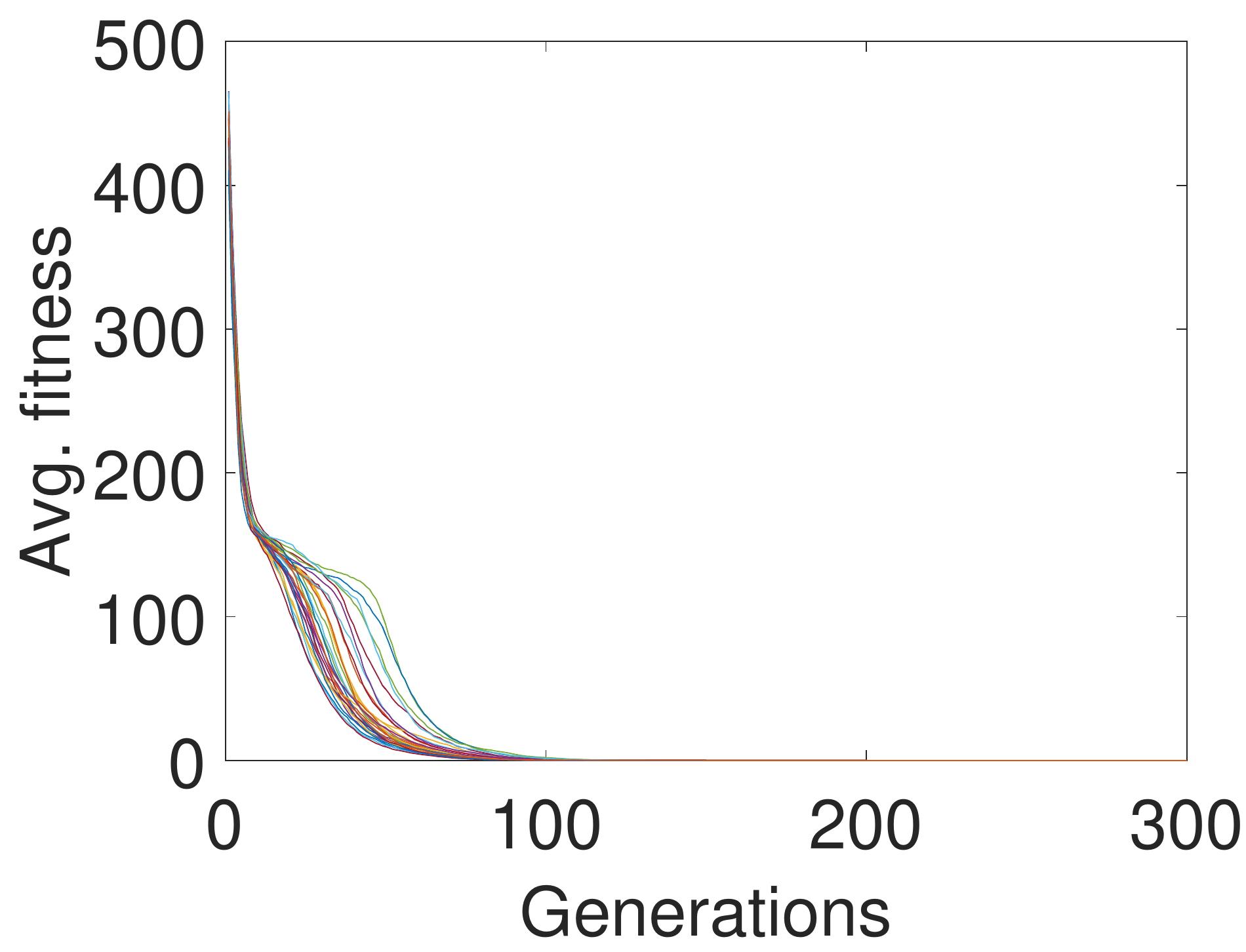}
        \caption{Classic Boids}
        \label{Fig:Classic8step}
    \end{subfigure} 
    \begin{subfigure}[b]{0.24\textwidth}
        \includegraphics[width=\textwidth,height=3.2cm]{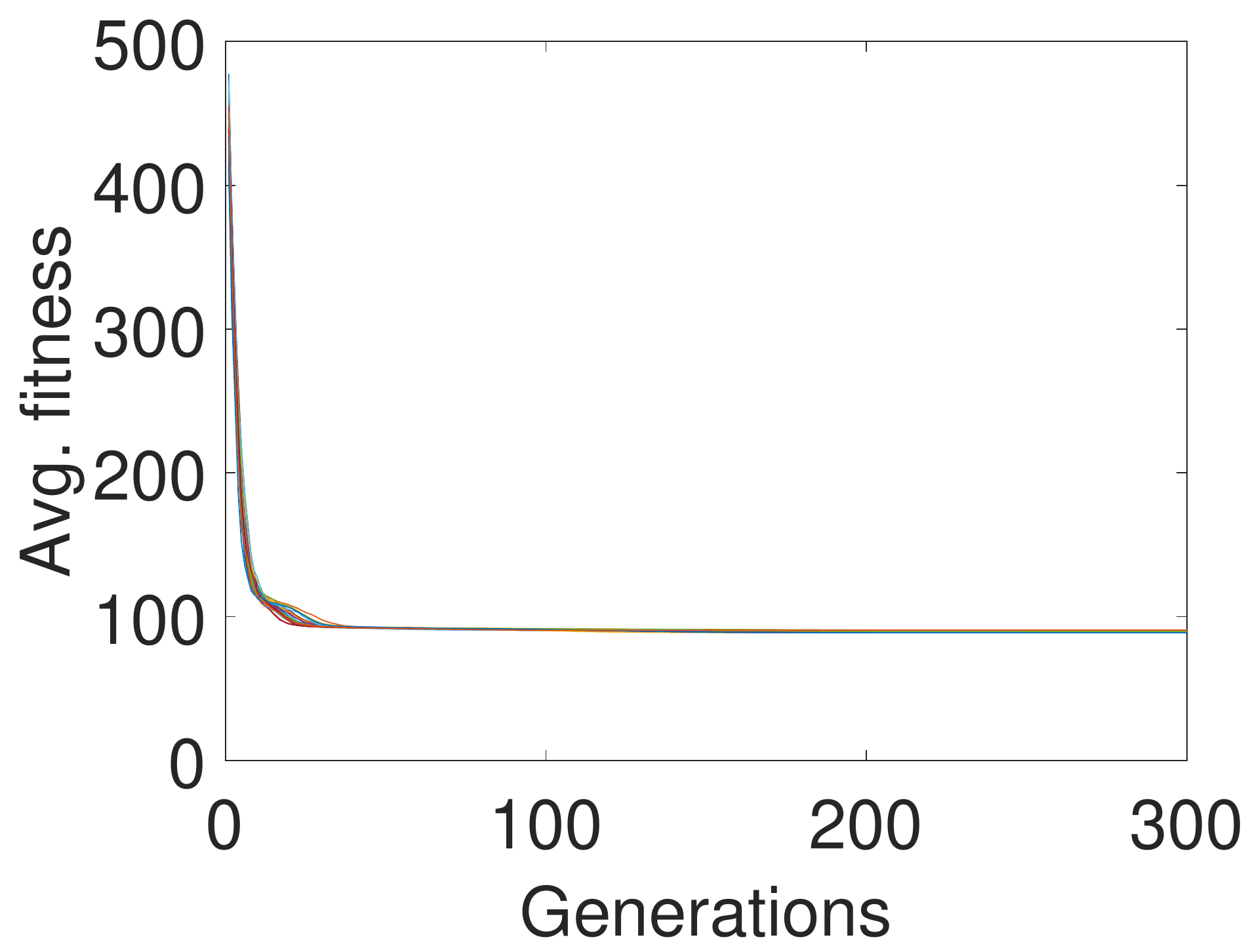}
        \caption{Erd\H{o}s\textendash R{\'e}nyi network}
        \label{Fig:ERNet8step}
    \end{subfigure} 
    \begin{subfigure}[b]{0.24\textwidth}
        \includegraphics[width=\textwidth,height=3.2cm]{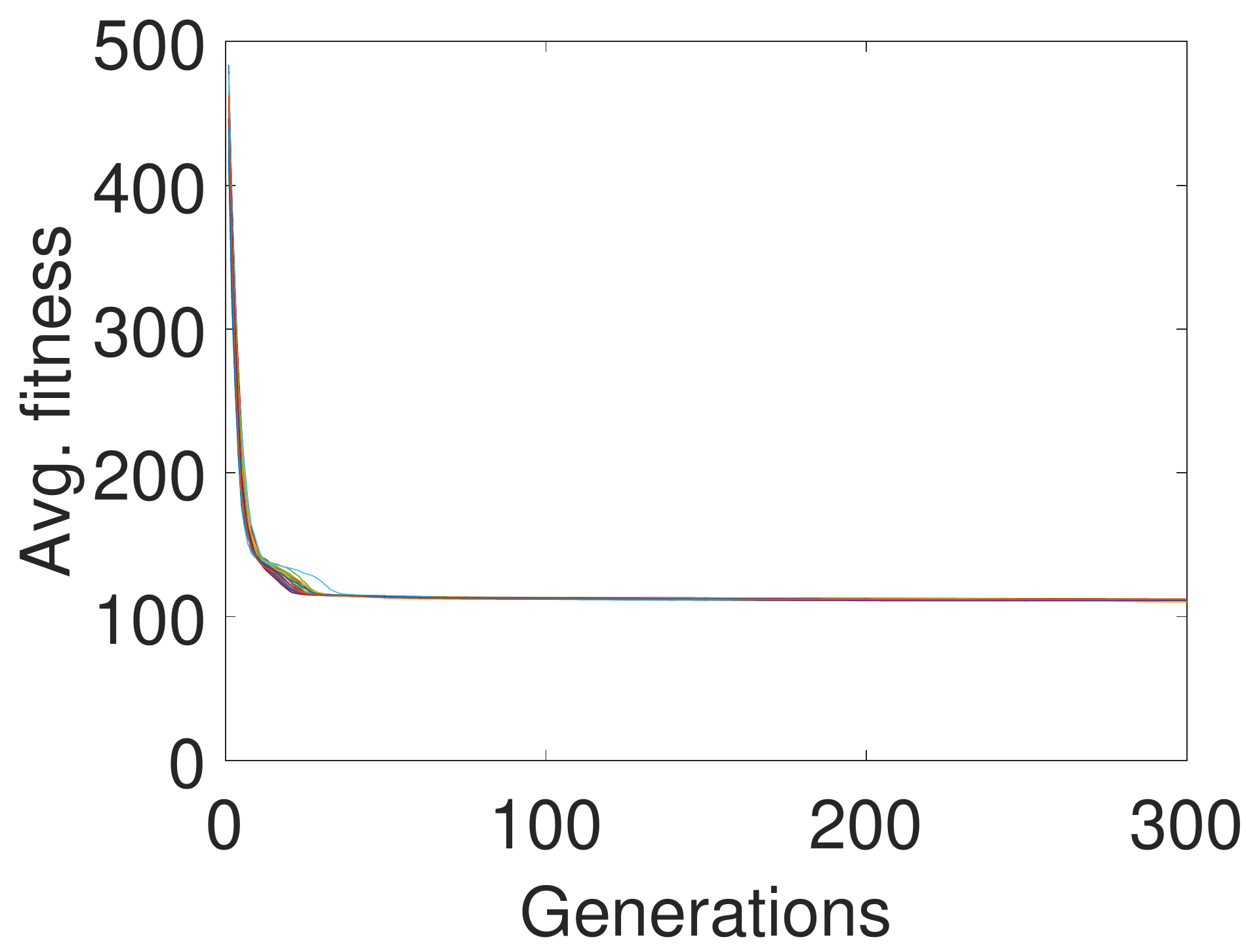}
        \caption{Scale-free network}
        \label{Fig:ScaleFree8step}
    \end{subfigure} 
    \begin{subfigure}[b]{0.24\textwidth}
        \includegraphics[width=\textwidth,height=3.2cm]{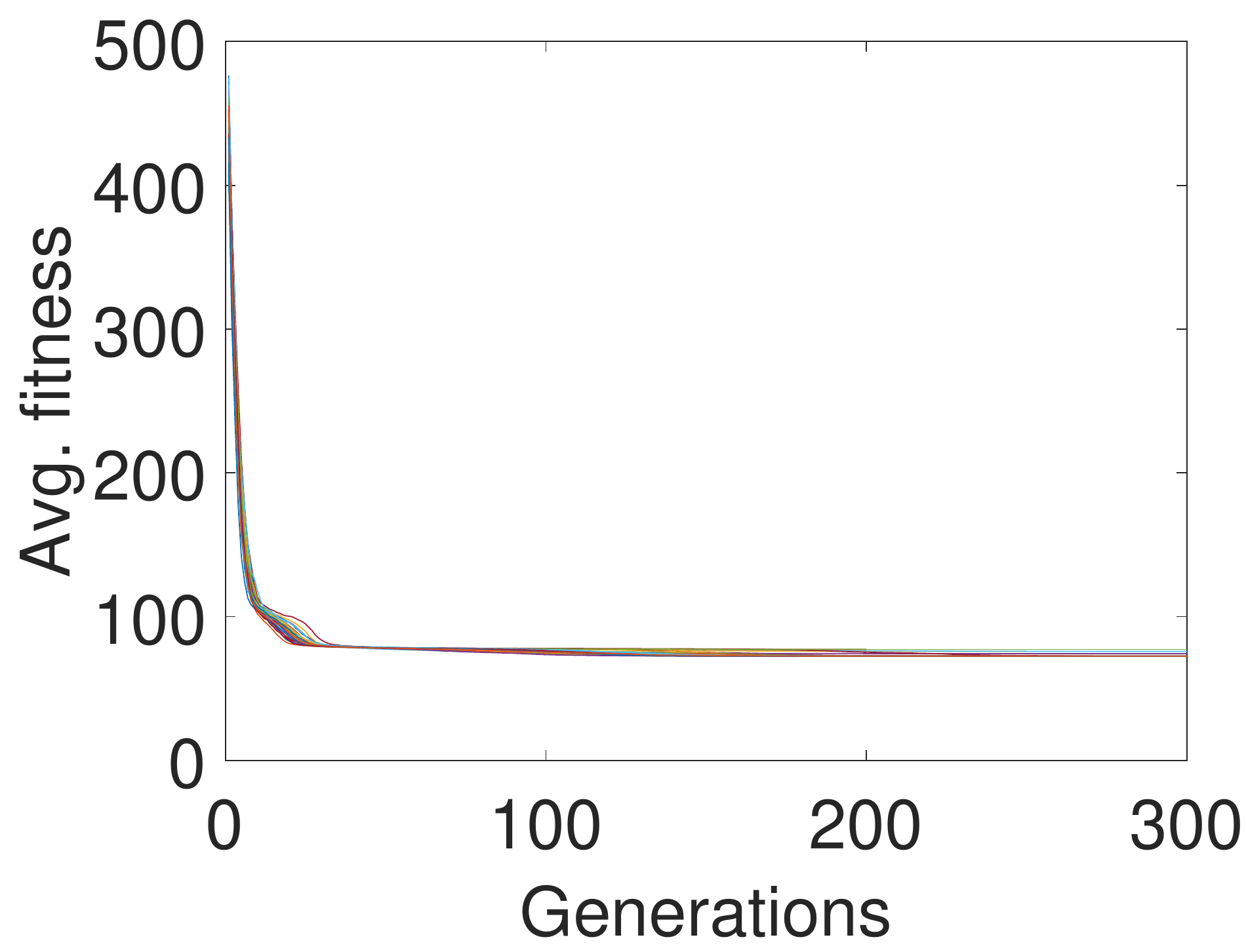}
        \caption{Small-world network}
        \label{Fig:SmallWorld8step}
    \end{subfigure}
    \caption{Average fitness evolution of 30 runs \rtext{(presented by colored curves)} for all boid types when the observation period is 8 iterations (from $t=0$ to $t=7$).}
    \label{Fig:DEEvo8Step}
\end{figure*}

\begin{figure*}[h]
    \center
    \begin{subfigure}[b]{0.24\textwidth}
        \includegraphics[width=\textwidth,height=3.2cm]{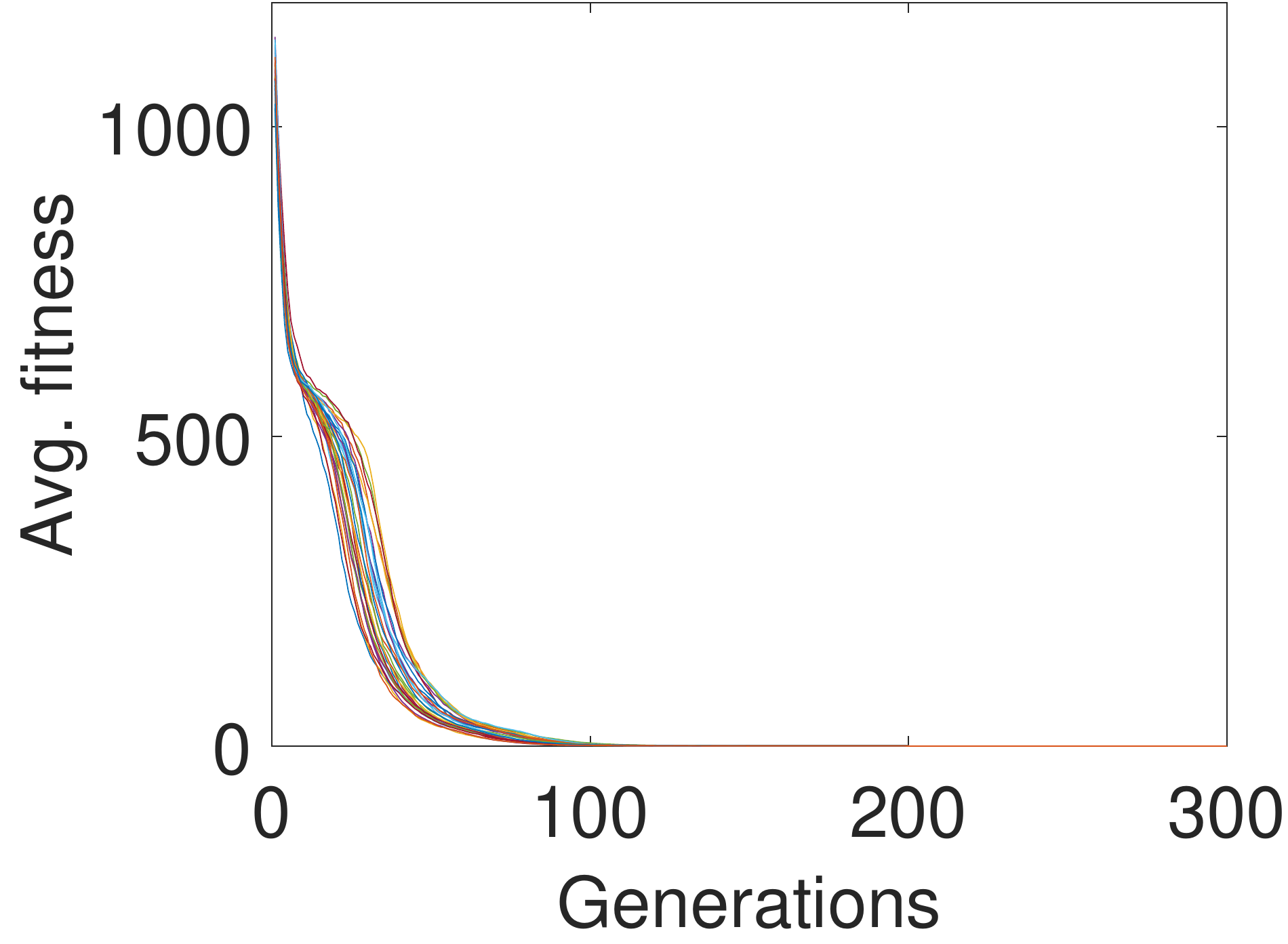}
        \caption{Classic Boids}
        \label{Fig:Classic16step}
    \end{subfigure} 
    \begin{subfigure}[b]{0.24\textwidth}
        \includegraphics[width=\textwidth,height=3.2cm]{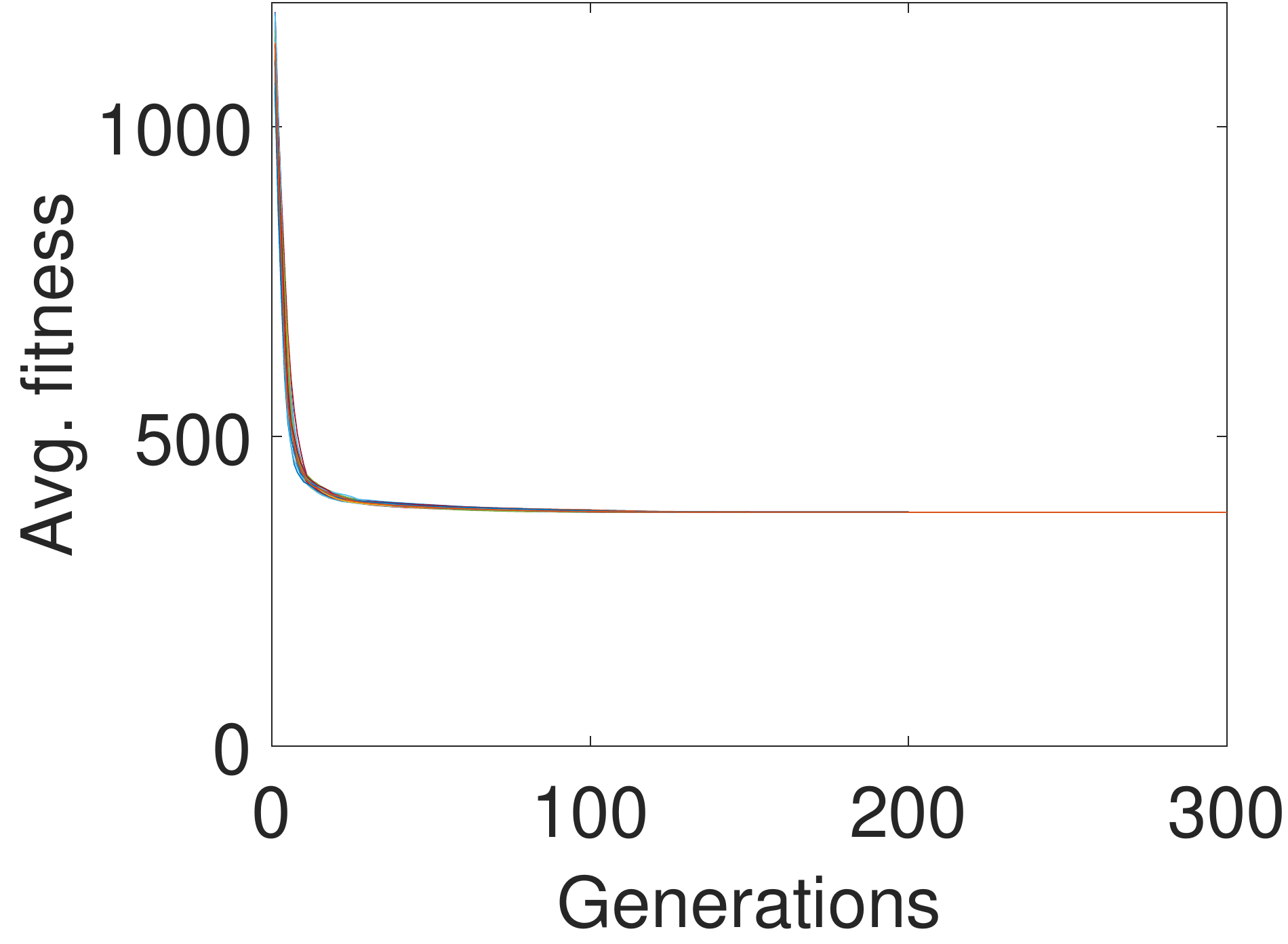}
        \caption{Erd\H{o}s\textendash R{\'e}nyi network}
        \label{Fig:ERNet16step}
    \end{subfigure} 
    \begin{subfigure}[b]{0.24\textwidth}
        \includegraphics[width=\textwidth,height=3.2cm]{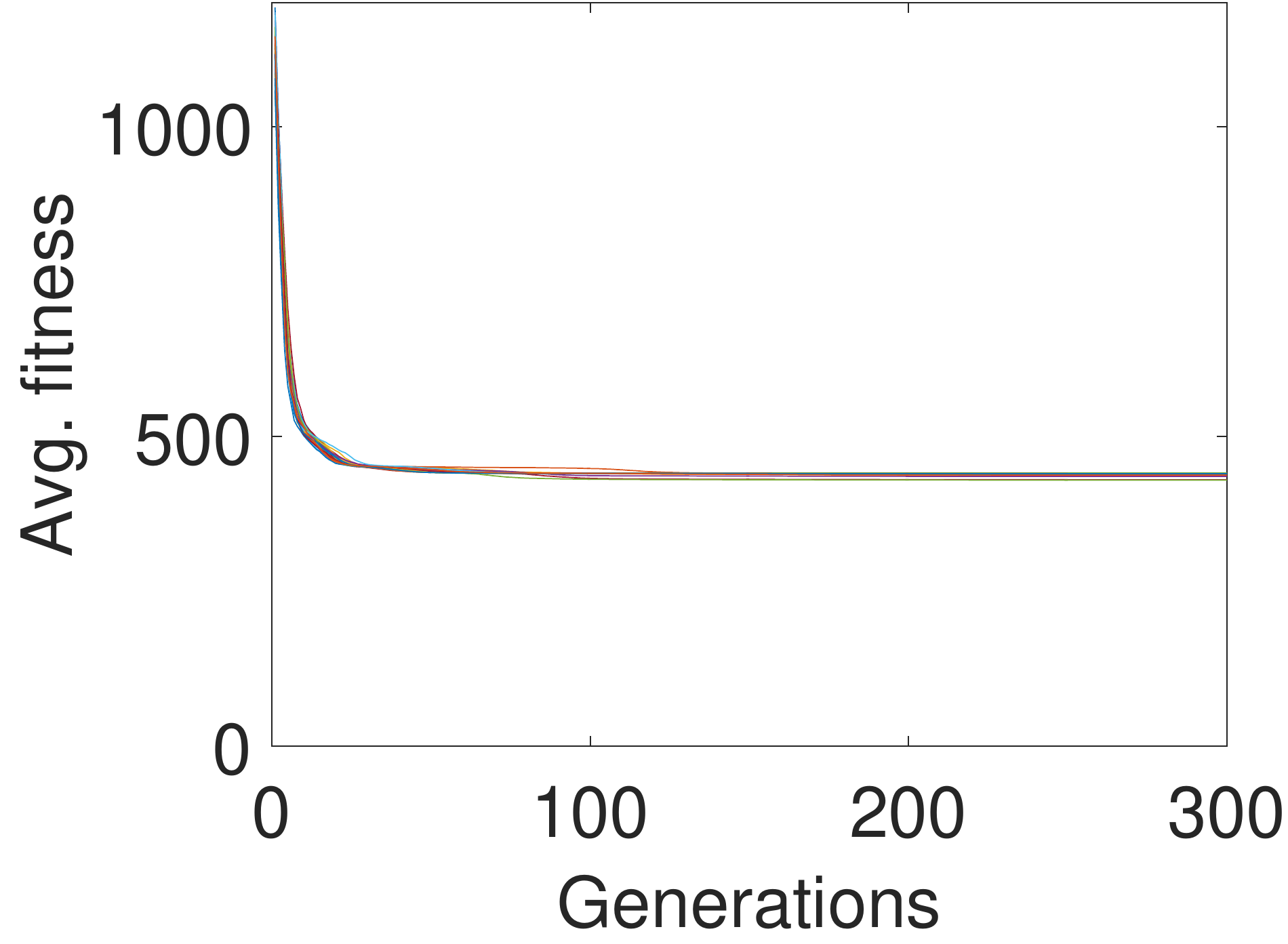}
        \caption{Scale-free network}
        \label{Fig:ScaleFree16step}
    \end{subfigure} 
    \begin{subfigure}[b]{0.24\textwidth}
        \includegraphics[width=\textwidth,height=3.2cm]{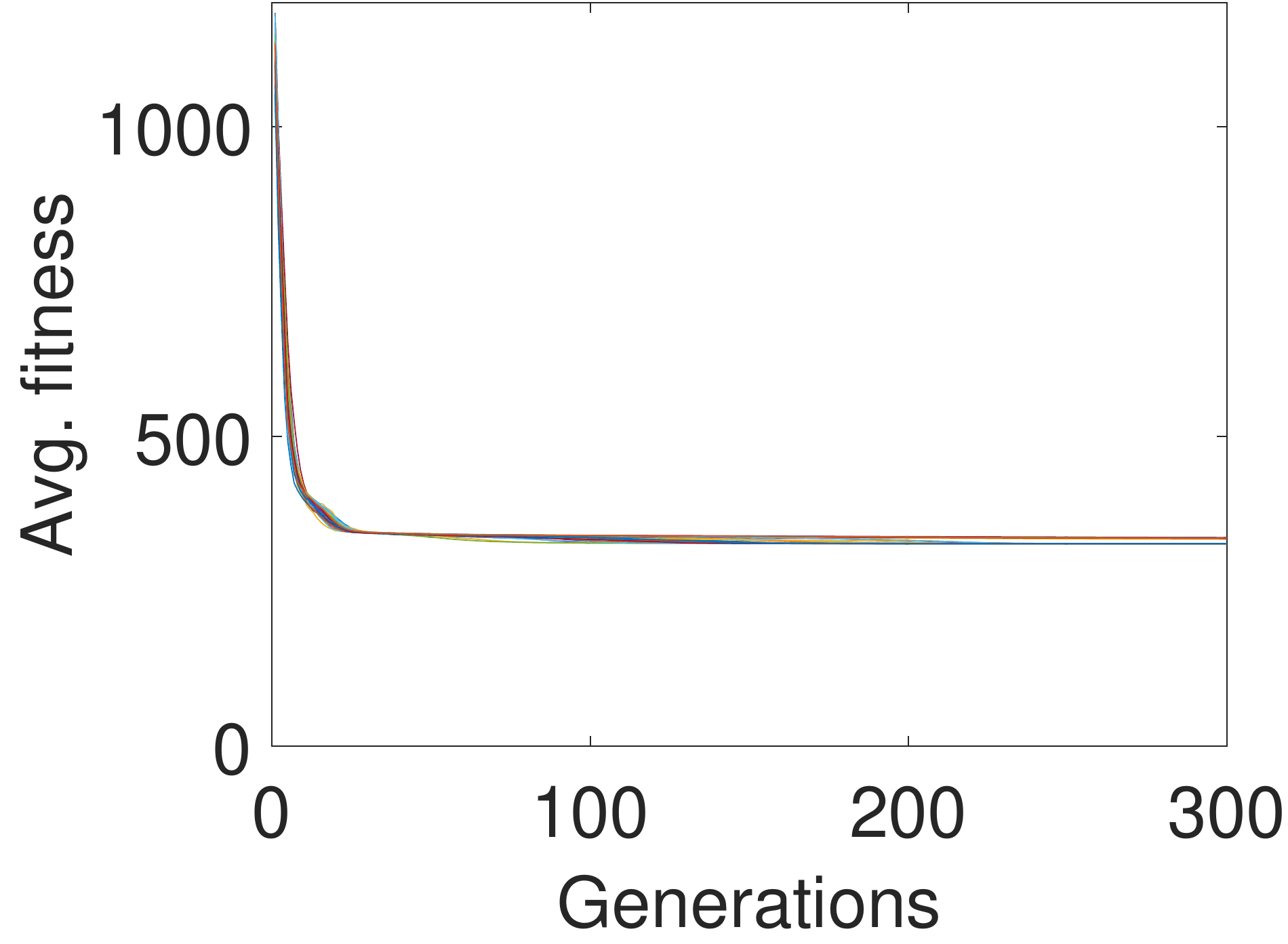}
        \caption{Small-world network}
        \label{Fig:SmallWorld16step}
    \end{subfigure}
    \caption{Average fitness evolution of 30 runs \rtext{(presented by colored curves)} for all boid types when the observation period is 16 iterations (from $t=0$ to $t=15$).}
    \label{Fig:DEEvo16Step}
\end{figure*}

Figures~\ref{Fig:DEEvo2Step},~\ref{Fig:DEEvo4Step},~\ref{Fig:DEEvo8Step}~and~\ref{Fig:DEEvo16Step} show the learning error ($\epsilon_L$) of the learning algorithm for all boid types when observation samples are taken from the first 2, 4, 8 and  16 time-steps, respectively. It can be seen that for all four look-back periods considered, the algorithm is able to learn well the parameters of the classic vision-based boids, i.e. error is reduced almost to $0$ regardless of the length of observation. This shows on the one hand that the learning algorithm is consistent in terms of concept, implementation, and convergence, and on the other hand that classic boids have reduced robustness to external learning of their behavioral rule parameters, even when the observer only takes the minimum possible sample set, i.e. 2 time-steps.

In contrast, for the network-based boids, the algorithm is unable to reduce the learning error to the level of the classic boids one. Not only that the error for network-based boids remains at significantly higher level, but also it seems that the learner becomes more confused as the length of the observation increases. That is, for all network-based boid types if error level decreases to roughly 20-25 for observation containing 2 iterations, this error increases significantly for 4, 8 and 16 iterations, reaching only around 400 in the latest case. 

To complement the visual indications given by the figures, in  Tables~\ref{Tab:BestOffline2steps},~\ref{Tab:BestOffline4steps},~\ref{Tab:BestOffline8steps},~and~~\ref{Tab:BestOffline16steps} we provide the values of the learned parameters corresponding to the best estimations for all boids types, for observation windows of 2, 4, 8, and 16 iterations, respectively. An interesting, but somewhat expected aspect revealed by the data in the tables is that the errors are influenced greatly by the learning of all parameters except the safe distance $d_s$. This is, we suggest, due to the fact the safe distance is a crisp parameter solely used for collision avoidance, and is independent of neighborhood type or boids motion rules/forces; hence it can be easily inferred.

\begin{table*}[h]
    \center
    \begin{tabular}{|l|r|r|r|r|r|r|r|r|} \hline
        Communication & $\epsilon_L$ & $w_c$ & $w_a$  & $w_s$   & $d_s$ & $vistion_r$ &  $vision_a$\\ \hline
        Classic & 0.000000 & 0.010000 & 0.125000 & 1.000000 & 10.008017 & 100.055706 & 360.000000 \\\hline 
        Small-world & 2.619003 & 0.016291 & 0.042408 & 1.000000 & 10.389850 & 43.378060 & 217.368837 \\\hline 
        Scale-free & 4.053324 & 0.000000 & 0.029161 & 1.000000 & 11.060216 & 117.657485 & 166.620710 \\\hline 
        Erdos-renyi & 3.001435 & 0.023727 & 0.033282 & 1.000000 & 10.692984 & 131.562692 & 99.927292 \\\hline 
    \end{tabular}
    \caption{The best estimations of the boids parameters generated by off-line learning when the observation period is 2 iterations (from $t=0$ to $t=1$).}
    \label{Tab:BestOffline2steps}
\end{table*}

\begin{table*}[h]
    \center
    \begin{tabular}{|l|r|r|r|r|r|r|r|} \hline
        Communication & $\epsilon_L$ & $w_c$ & $w_a$  & $w_s$   & $d_s$ & $vistion_r$ &  $vision_a$\\ \hline
        Classic & 0.000000 & 0.010000 & 0.125000 & 1.000000 & 10.000155 & 100.009700 & 360.000000 \\\hline 
        Small-world & 15.416319 & 0.018548 & 0.039641 & 0.914656 & 12.756913 & 43.837231 & 212.046604 \\\hline 
        Scale-free & 23.763785 & 0.025768 & 0.262515 & 1.000000 & 11.434917 & 15.937590 & 188.890207 \\\hline 
        Erdos-renyi & 17.921862 & 0.024994 & 0.033572 & 0.879487 & 10.888589 & 130.649365 & 100.111378 \\\hline 
    \end{tabular}
    \caption{The best estimations of the boids parameters generated by off-line learning when the observation period is 4 iterations (from $t=0$ to $t=3$).}
    \label{Tab:BestOffline4steps}
\end{table*}

\begin{table*}[h]
    \center
    \begin{tabular}{|l|r|r|r|r|r|r|r|} \hline
        Communication & $\epsilon_L$ & $w_c$ & $w_a$  & $w_s$   & $d_s$ & $vistion_r$ &  $vision_a$\\ \hline
        Classic & 0.000000 & 0.010000 & 0.125000 & 1.000000 & 9.994598 & 100.001520 & 360.000000 \\\hline 
        Small-world & 72.437172 & 0.019789 & 0.040718 & 0.786630 & 11.933876 & 44.659179 & 216.722268 \\\hline 
        Scale-free & 109.808123 & 0.014750 & 0.000000 & 0.834566 & 12.649336 & 132.718430 & 323.038468 \\\hline 
        Erdos-renyi & 88.760579 & 0.169959 & 0.569699 & 1.000000 & 11.147494 & 12.265409 & 286.316729 \\\hline  
    \end{tabular}
    \caption{The best estimations of the boids parameters generated by off-line learning when the observation period is 8 iterations (from $t=0$ to $t=7$).}
    \label{Tab:BestOffline8steps}
\end{table*}

\begin{table*}[h]
    \center
    \begin{tabular}{|l|r|r|r|r|r|r|r|r|} \hline
        Communication & $\epsilon_L$ & $w_c$ & $w_a$  & $w_s$   & $d_s$ & $vistion_r$ &  $vision_a$\\ \hline
        Classic & 0.000000 & 0.010000 & 0.125000 & 1.000000 & 9.998525 & 99.999680 & 360.000000 \\\hline 
        Small-world & 326.614503 & 0.022598 & 0.032899 & 0.559174 & 12.034676 & 33.348825 & 166.825602 \\\hline 
        Scale-free & 429.290060 & 0.000110 & 0.049553 & 0.732560 & 13.034045 & 70.677480 & 106.999366 \\\hline 
        Erdos-renyi & 377.022850 & 0.123470 & 0.999998 & 1.000000 & 9.900569 & 10.000000 & 290.785531 \\\hline 
    \end{tabular}
    \caption{The best estimations of the boids parameters generated by off-line learning when the observation period is 16 iterations (from $t=0$ to $t=15$).}
    \label{Tab:BestOffline16steps}
\end{table*}

\begin{figure*}[h]
    \center
    \begin{subfigure}[b]{0.32\textwidth}
        \includegraphics[width=\textwidth,height=3.2cm]{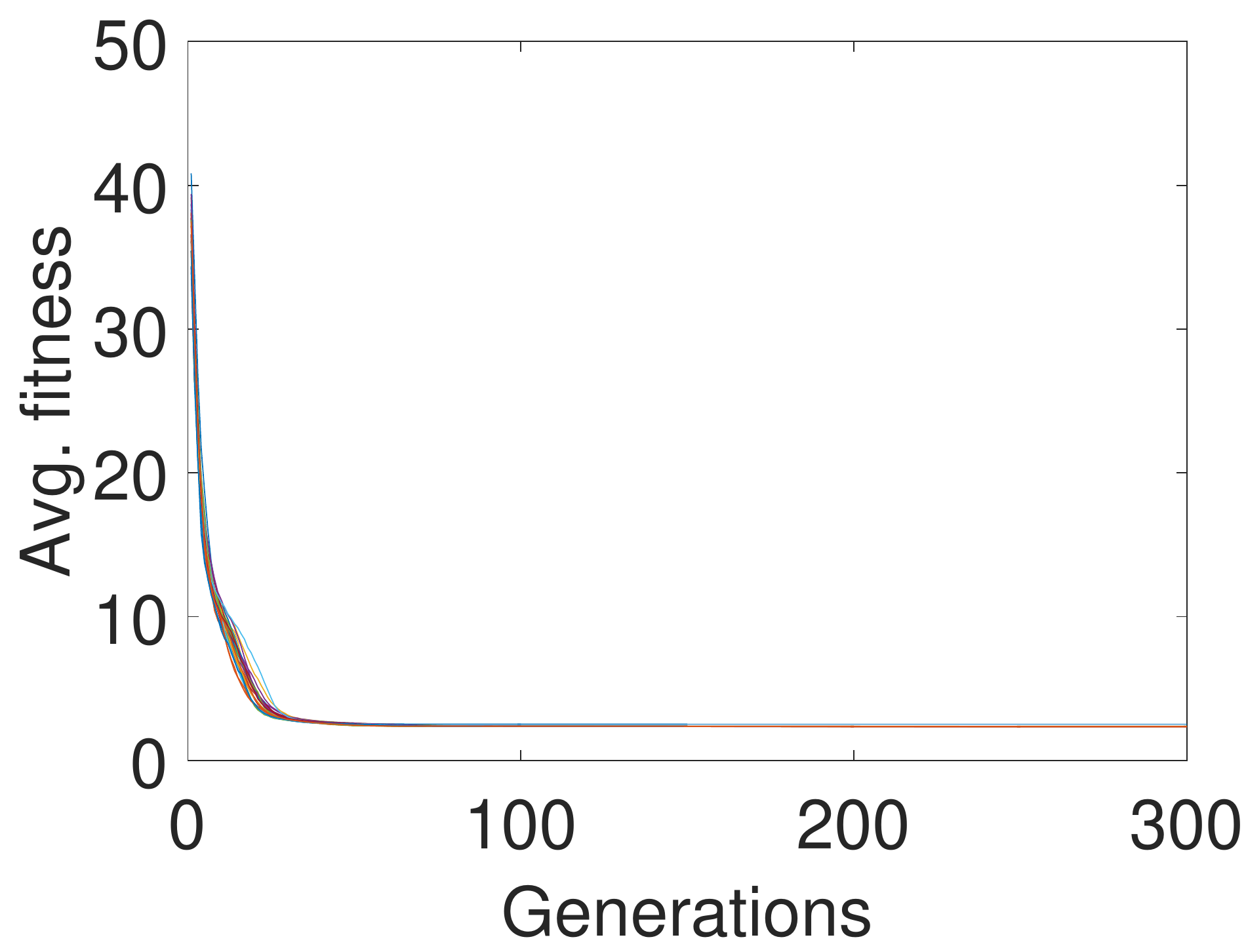}
        \caption{ER-net: 2 Observations}
        \label{Fig:ERNet2step5k}
    \end{subfigure} 
    \begin{subfigure}[b]{0.32\textwidth}
        \includegraphics[width=\textwidth,height=3.2cm]{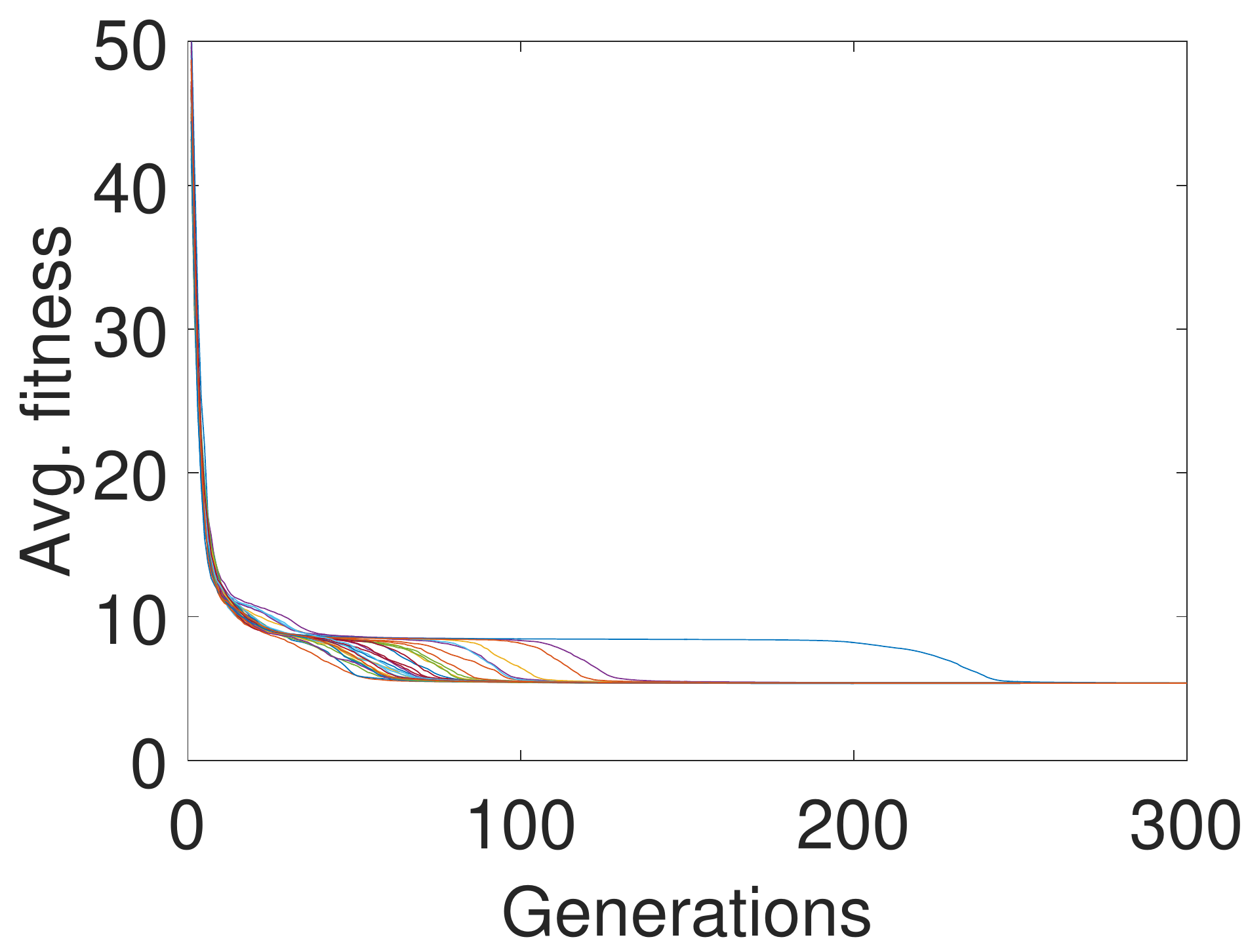}
        \caption{SF-net: 2 Observations}
        \label{Fig:ScaleFree2step5k}
    \end{subfigure} 
    \begin{subfigure}[b]{0.32\textwidth}
        \includegraphics[width=\textwidth,height=3.2cm]{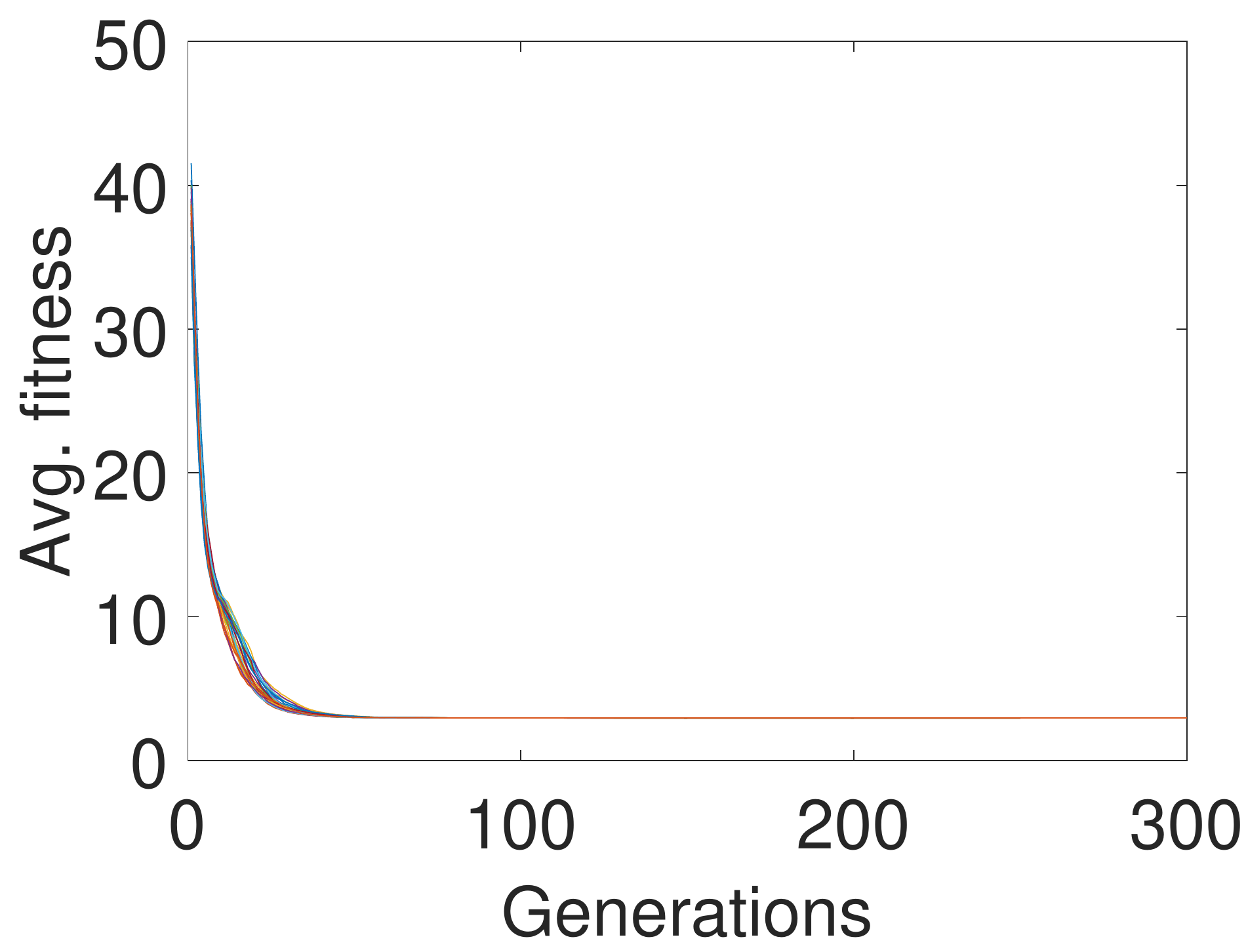}
        \caption{SW-net: 2 Observations}
        \label{Fig:SmallWorld2step5K}
    \end{subfigure} \\
    \begin{subfigure}[b]{0.32\textwidth}
        \includegraphics[width=\textwidth,height=3.2cm]{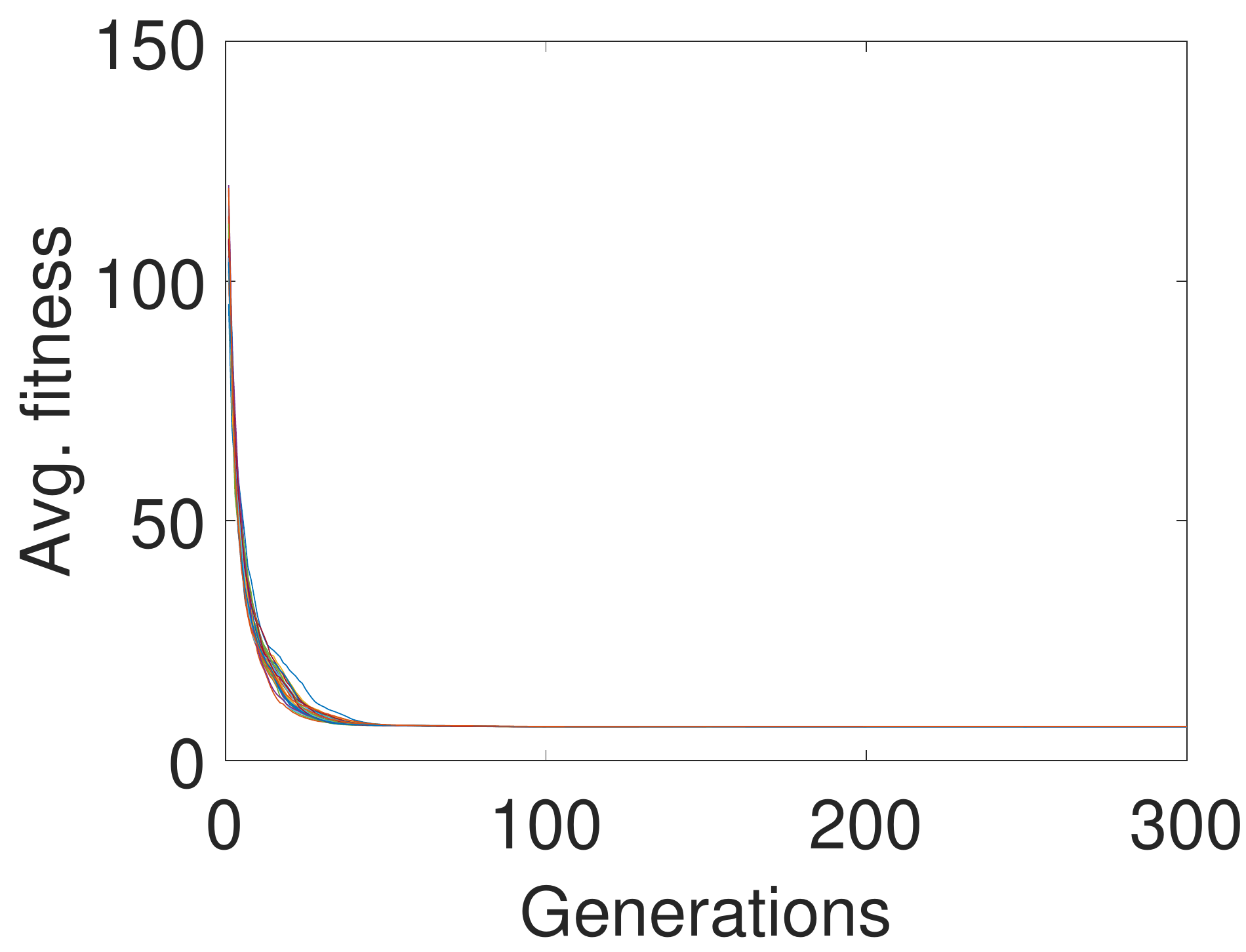}
        \caption{ER-net: 4 Observations}
        \label{Fig:ERNet4step5k}
    \end{subfigure} 
    \begin{subfigure}[b]{0.32\textwidth}
        \includegraphics[width=\textwidth,height=3.2cm]{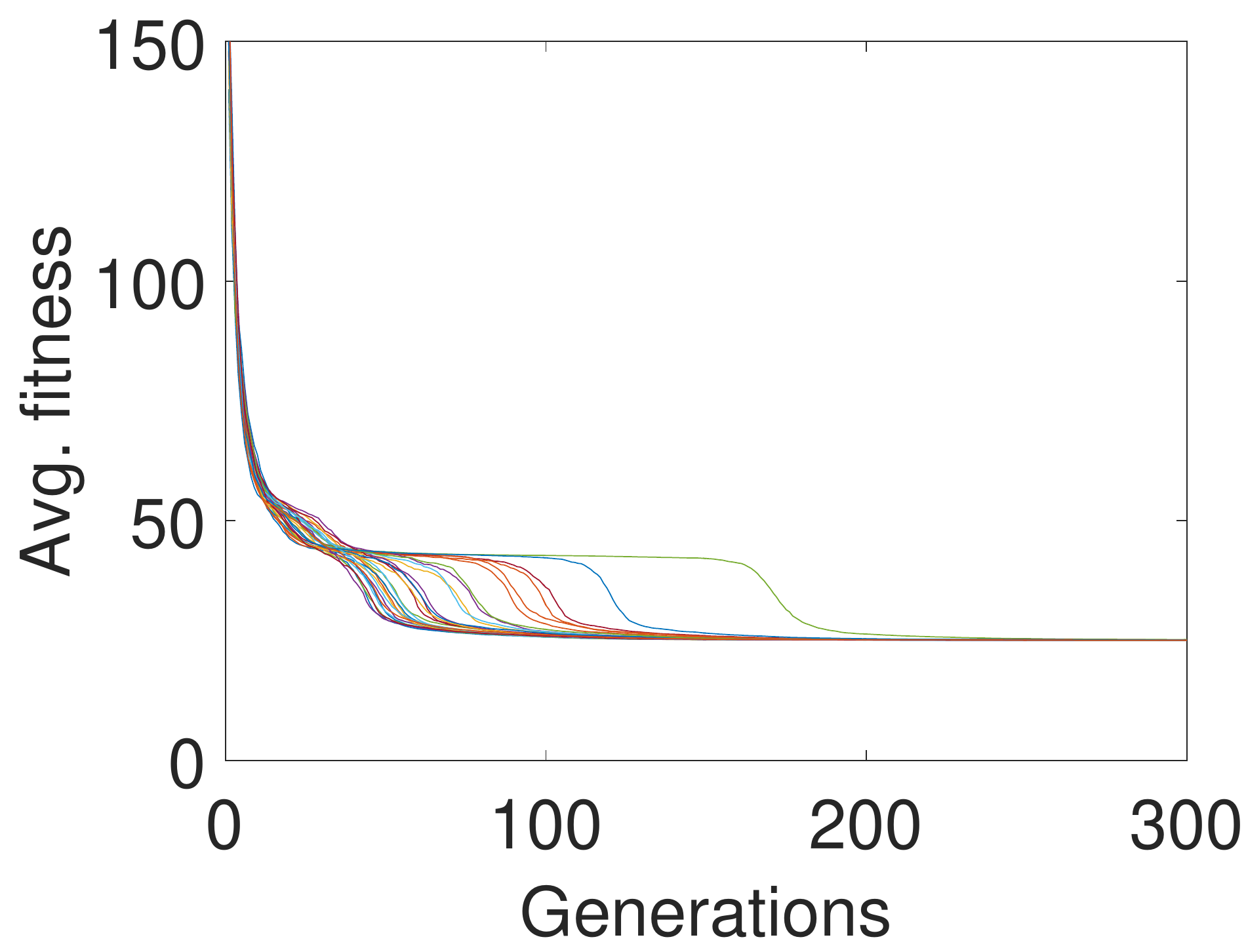}
        \caption{SF-net:: 4 Observations}
        \label{Fig:ScaleFree4step5k}
    \end{subfigure} 
    \begin{subfigure}[b]{0.32\textwidth}
        \includegraphics[width=\textwidth,height=3.2cm]{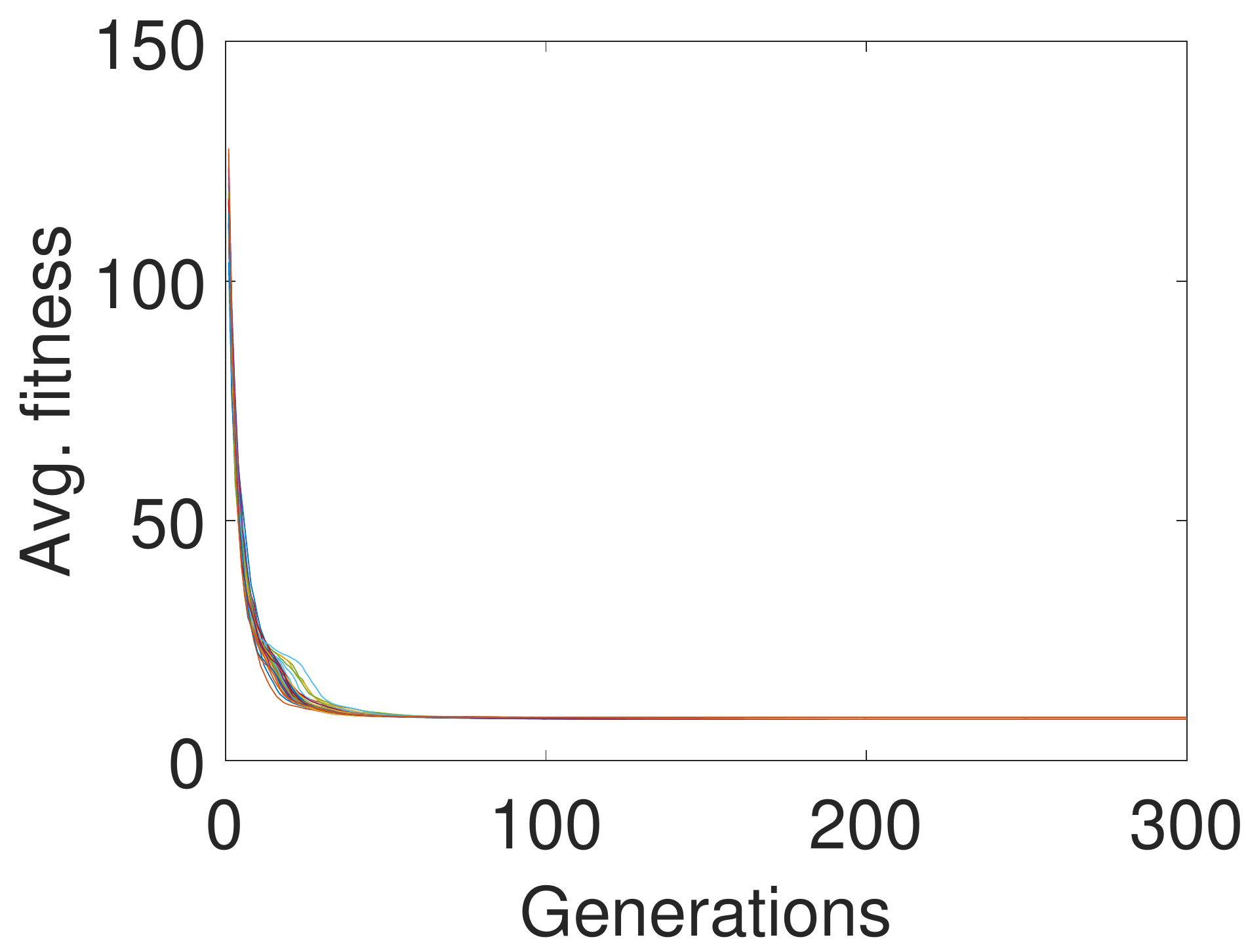}
        \caption{SW-net:: 4 Observations}
        \label{Fig:SmallWorld4step5K}
    \end{subfigure} \\
    \begin{subfigure}[b]{0.32\textwidth}
        \includegraphics[width=\textwidth,height=3.2cm]{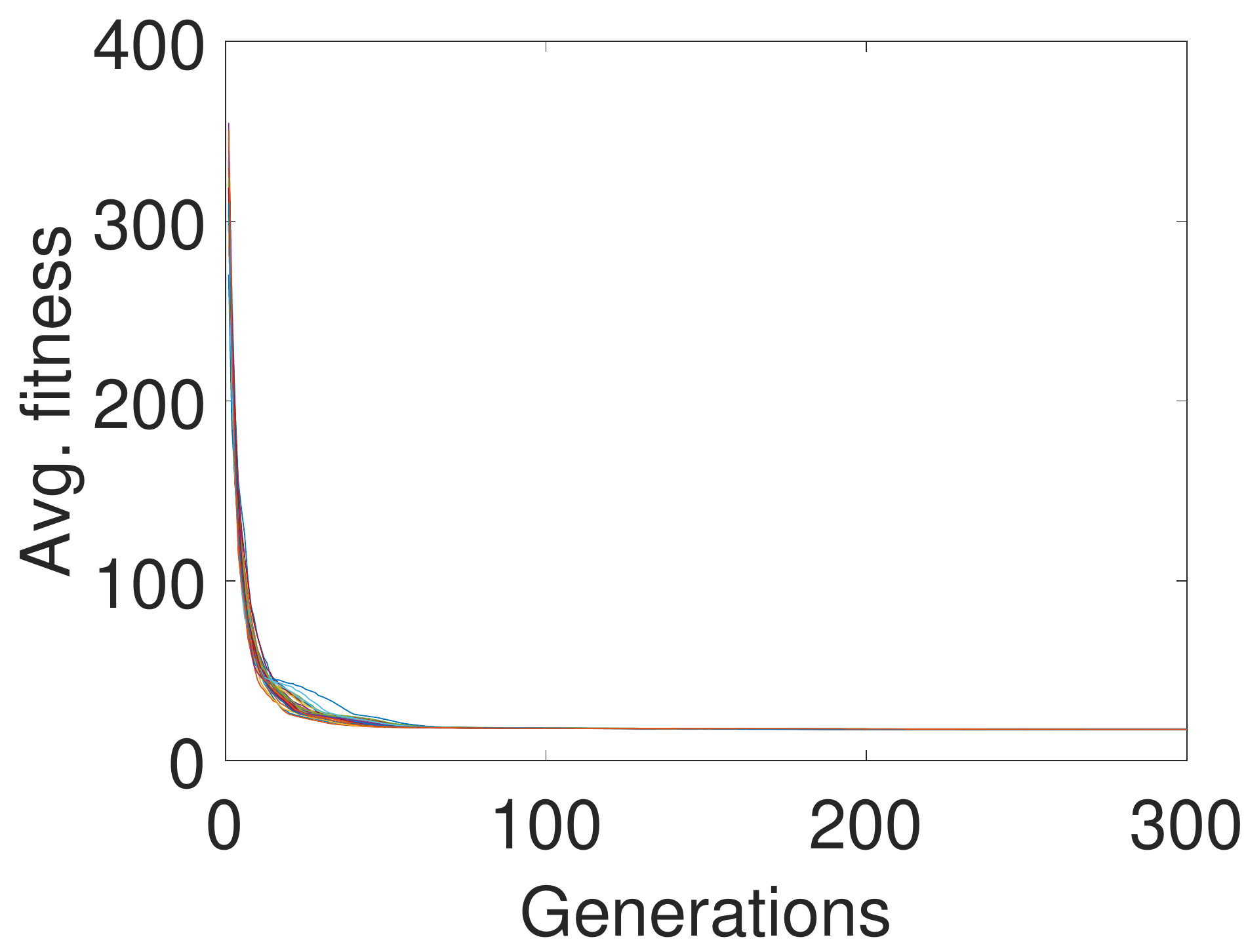}
        \caption{ER-net: 8 Observations}
        \label{Fig:ERNet8step5k}
    \end{subfigure} 
    \begin{subfigure}[b]{0.32\textwidth}
        \includegraphics[width=\textwidth,height=3.2cm]{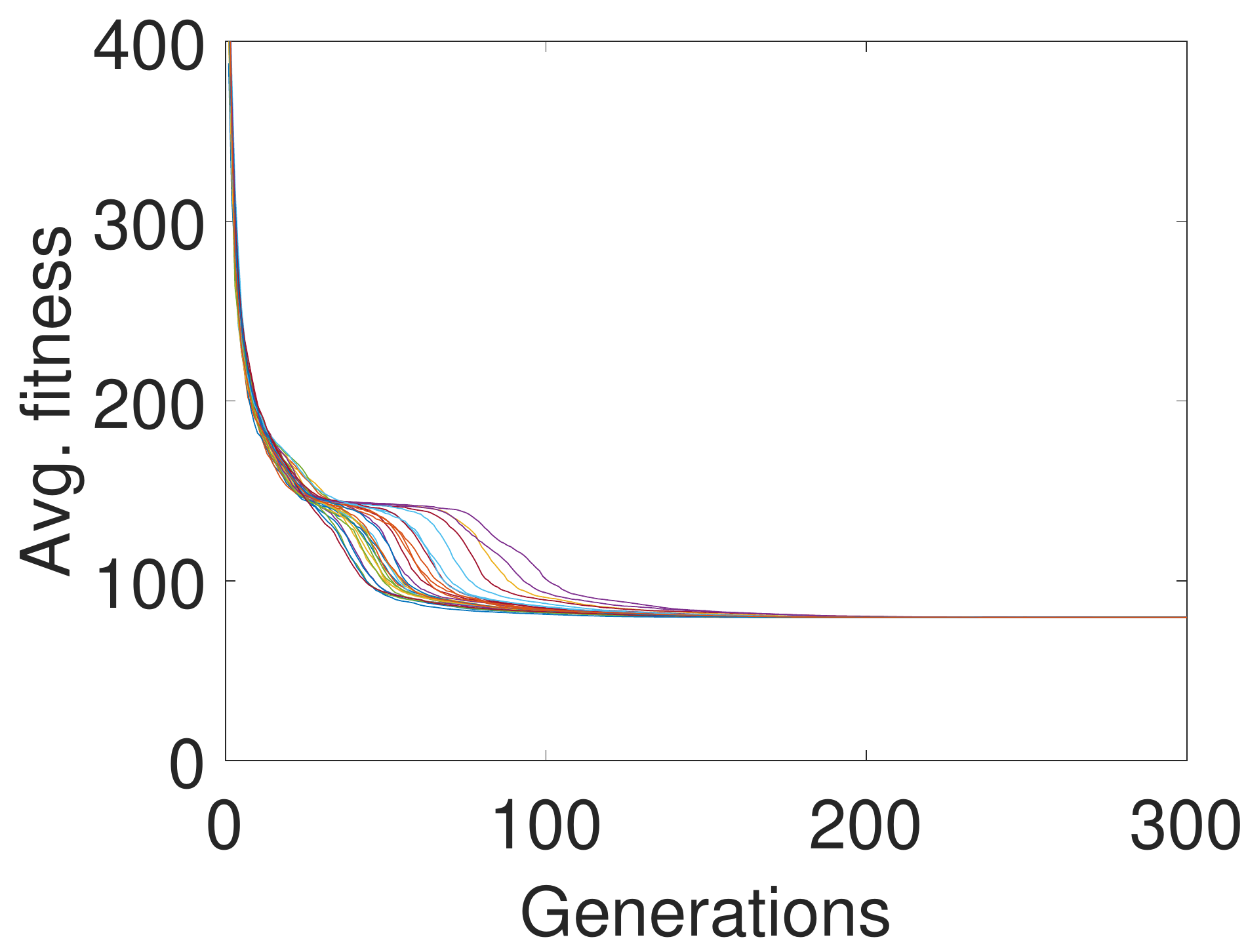}
        \caption{SF-net: 8 Observations}
        \label{Fig:ScaleFree8step5k}
    \end{subfigure} 
    \begin{subfigure}[b]{0.32\textwidth}
        \includegraphics[width=\textwidth,height=3.2cm]{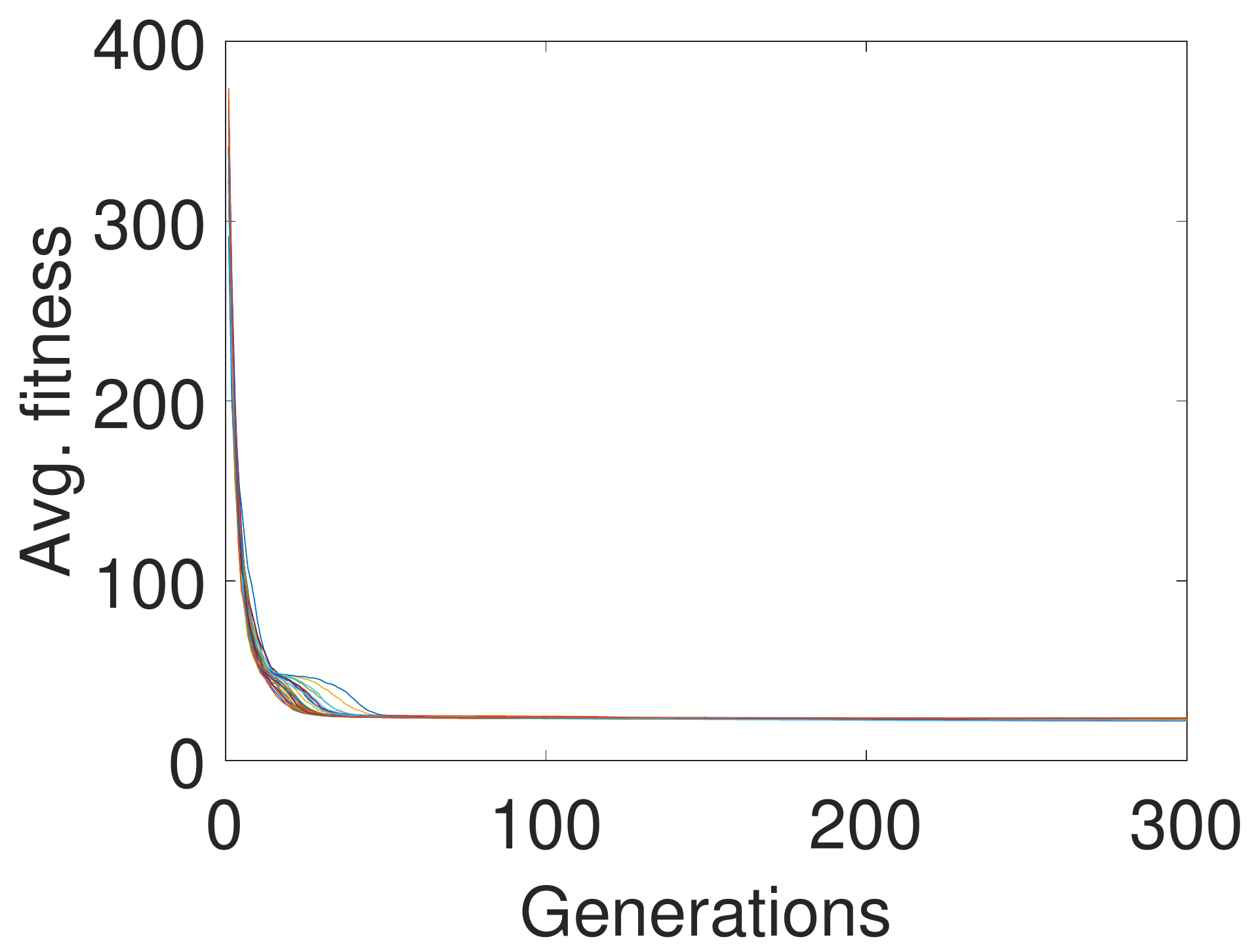}
        \caption{SW-net: 8 Observations}
        \label{Fig:SmallWorld8step5K}
    \end{subfigure} \\
    \begin{subfigure}[b]{0.32\textwidth}
        \includegraphics[width=\textwidth,height=3.2cm]{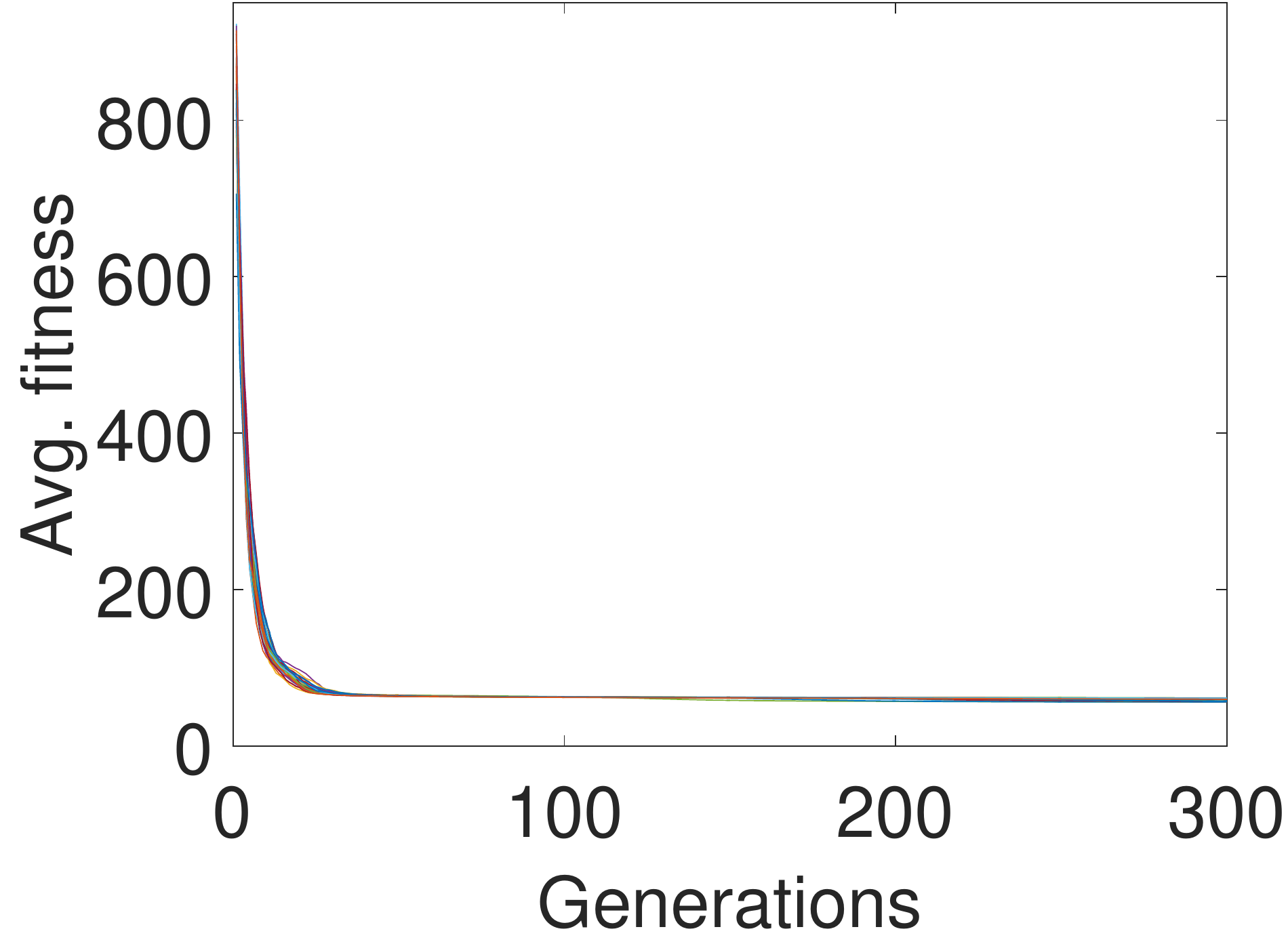}
        \caption{ER-net: 16 Observations}
        \label{Fig:ERNet16step5k}
    \end{subfigure} 
    \begin{subfigure}[b]{0.32\textwidth}
        \includegraphics[width=\textwidth,height=3.2cm]{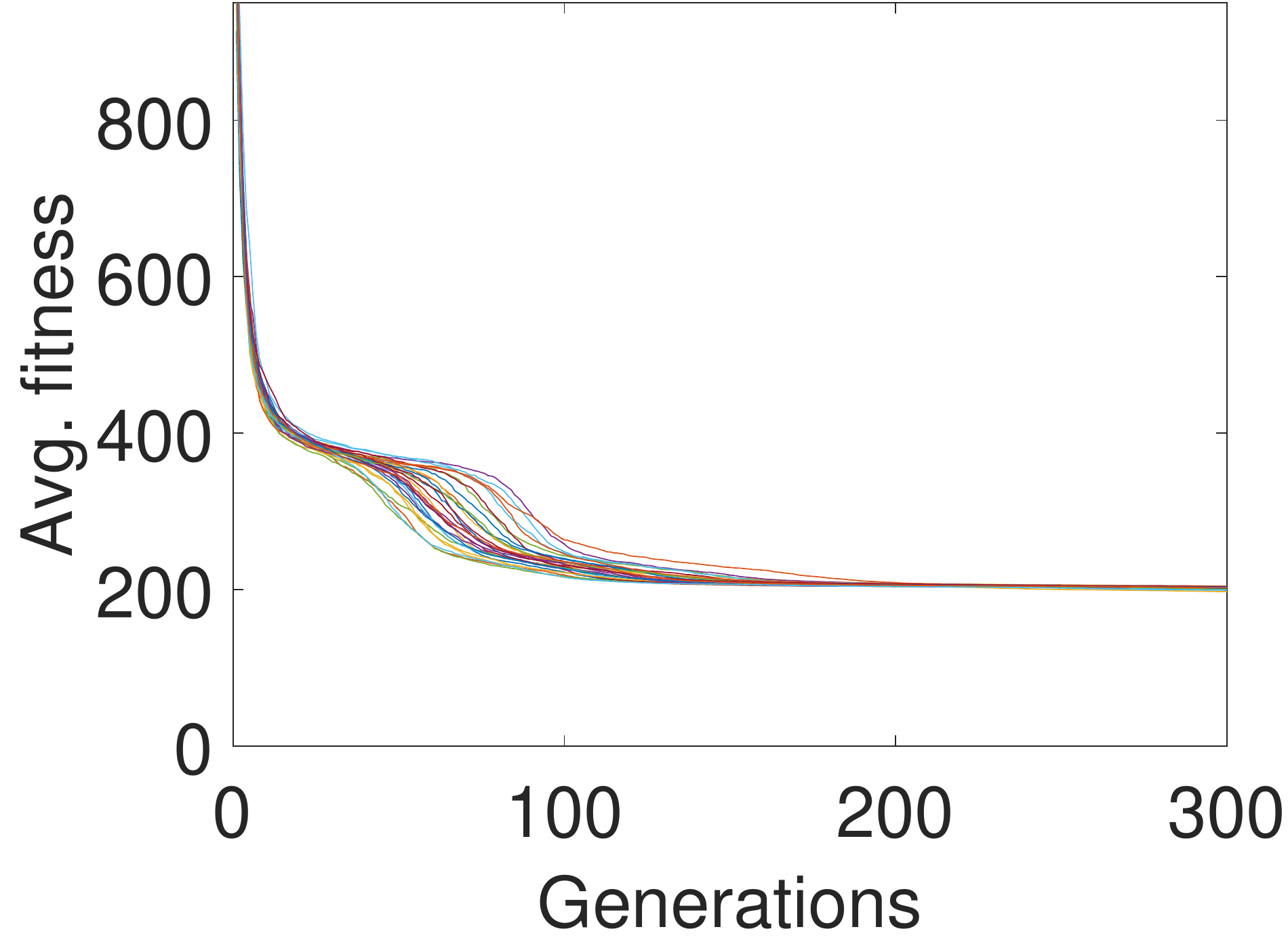}
        \caption{SF-net: 16 Observations}
        \label{Fig:ScaleFree16step5k}
    \end{subfigure} 
    \begin{subfigure}[b]{0.32\textwidth}
        \includegraphics[width=\textwidth,height=3.2cm]{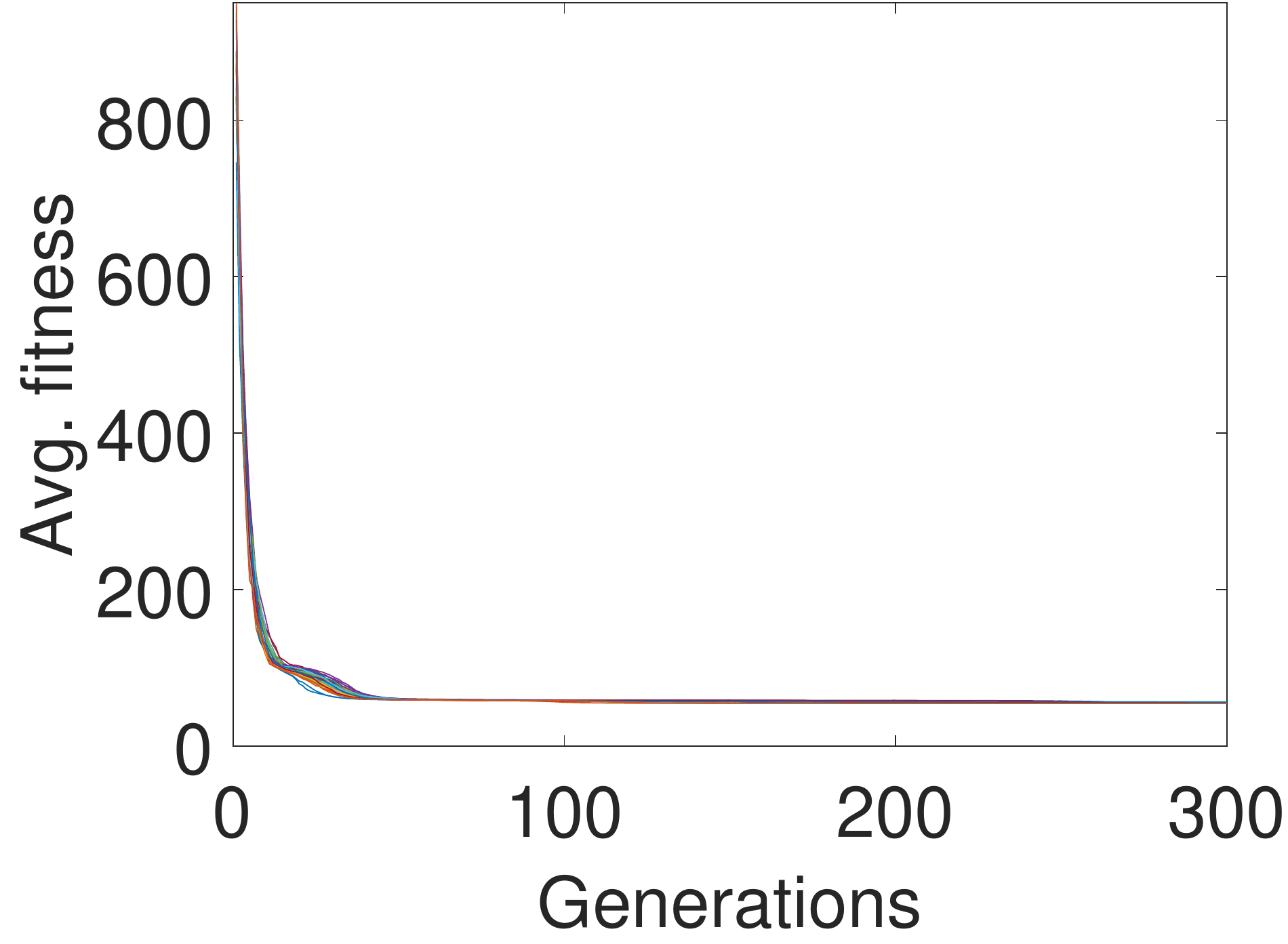}
        \caption{SW-net: 16 Observations}
        \label{Fig:SmallWorld16step5K}
    \end{subfigure} \\
    \caption{Average fitness evolution of 30 runs for all boid types, observation period of 2, 4, 8, and 16 iterations (from $t=5000$).}
    \label{Fig:DEEvoStep5K}
\end{figure*}

For the second case, observation samples are taken at $t=5000$, where swarm behavior is already established, as shown earlier in the paper in Section~\ref{Sec:EvalSwarmBehav}. For this case the expectation is that the learning will be easier, since boids are already ordered and grouped consistently in a good quality swarm motion; thus, learning errors should reach lower levels then in the first case. Since classic vision-based boids already achieved errors close to zero in the first case, when learning at the beginning of the simulation, there is no room left for improvement in the second case. For this reason, we only consider the network-based boids for this second case. Results of learning the network-based boids behaviors are illustrated in Figure~\ref{Fig:DEEvoStep5K}. As expected, the DE learner is able to reach lower errors then the first case, for all three network-based boids and for all observation periods. However, while an improvement in learning performance can be seen, we also note that learning has the same trend like in the first case, where the ability of the learning algorithm to reduce error level decreased as the observation period increased. This shows that even if the swarm behavior is established and boids are well ordered and grouped, the external observer still gets confused when observing multiple iterations of the swarm behavior. From a different point of view, when swarm behavior is well established, the learner can learn from network-based boids better than in the first case, where boids behaviors were still \textquoteleft unstable\textquoteright  \ with respect to the observer, but this is still significantly below what it can learn from the classic vision-based boids.


In summary, the proposed DE-based learning algorithm is able to learn well the classic boids behavior in any situation, but is not able to learn well enough the behavior of network-based boids, even when boids are well ordered and grouped, at 5000 iterations. 

\subsubsection{On-line Learning Mode}
The next stage is to investigate the real-time learning and prediction, where the learning algorithm operates in an on-line mode, in which it repeats for the whole duration of the boids simulation the following cycle: sample-learn-predict (as explained in Figure~\ref{Fig:FastLearning}).

\begin{figure}[h]
    \center
    \includegraphics[width=7cm,height=6cm]{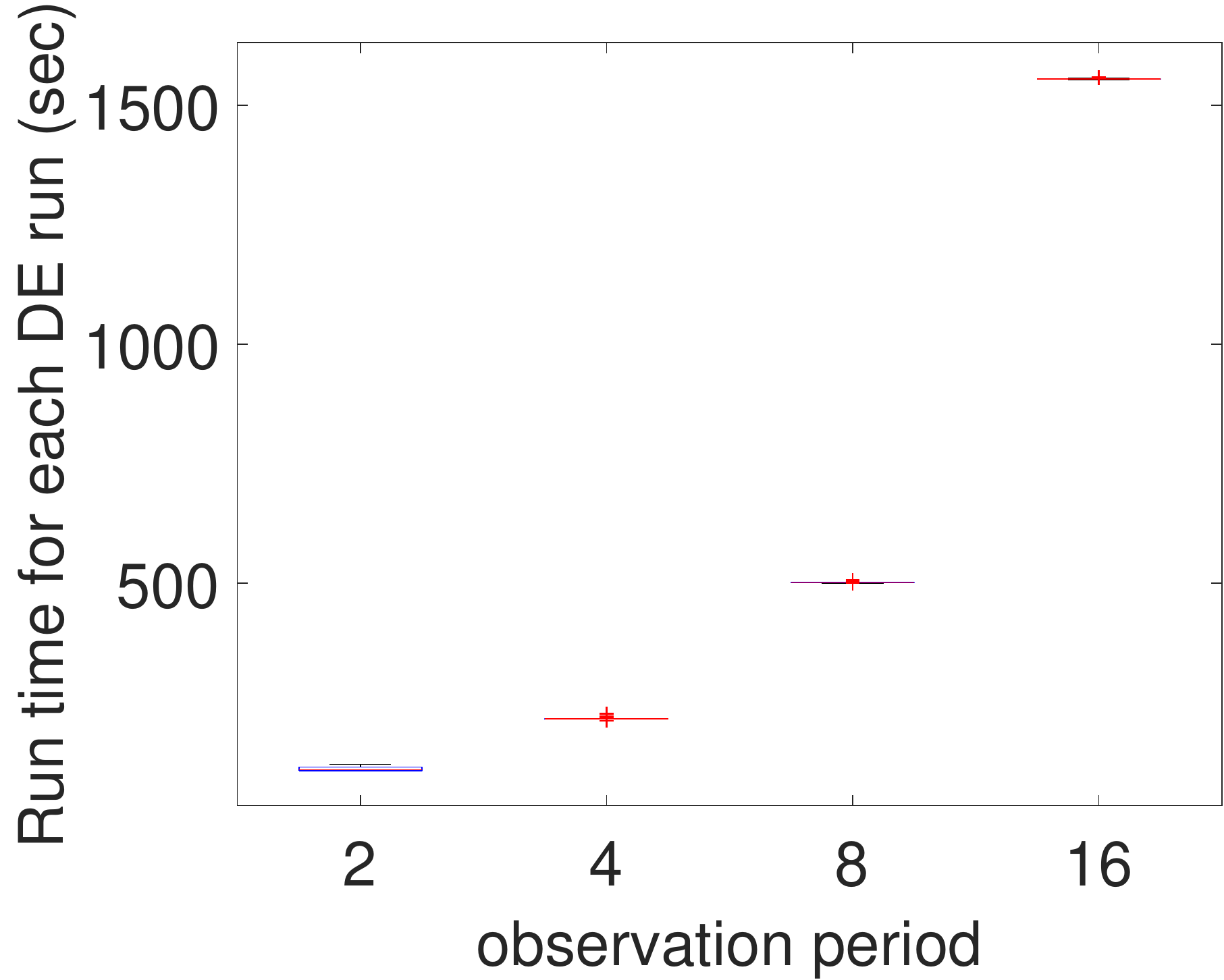}
    \caption{The running time of each DE run for Classic boids}
    \label{Fig:DERunTime}
\end{figure}

In order to perform the investigation, we need an insight into the computational cost of learning. Figure~\ref{Fig:DERunTime} shows the time needed for learning classic boids behavior based on different observation periods. Based on this, we decide on how long can the learning algorithm run in an on-line learning mode, i.e. how many generations can the DE-based learner evolve in order to ensure real-time performance of the sample-learn-predict cycle.

Taking into account that an increase in the observation period leads to an increase in the necessary running time for DE-based learning algorithm, we try to make the algorithm suitable as much as possible for all observation periods (look-back time) we used in the previous experiments, i.e. 2, 4, 8 and 16 time-steps. Thus, the adopt the following settings:
\begin{itemize}
    \item learning period is 600 time-steps, that is, 60 seconds for an update rate of 10ms;
    \item prediction period is set to 1200 time-steps in order to ensure a relevant comparison base between predicted and actual swarm movement;
    \item maximum number of generations in each on-line learning cycle is set to 50. Based on the results discussed earlier in Section~\ref{Sec:OfflineLearning}, this will allow the learning algorithm to finish learning when observation periods are 2 or 4 time-steps but learning for 8 and 16 time-steps observation periods will be slightly truncated.
\end{itemize}

With the settings established as above, in a swarm simulation of 10000 steps there are 15 complete sample-learn-predict cycles. We run simulations for all boid types, vision-based and network-based, for all four observation periods (2, 4, 8 and 16 time-steps), and we repeat simulations 10 times with different seeds for each case in order to ensure statistical consistency of results.

\begin{figure*}[h]
    \center
    \begin{subfigure}[b]{0.24\textwidth}
        \includegraphics[width=\textwidth,height=3.2cm]{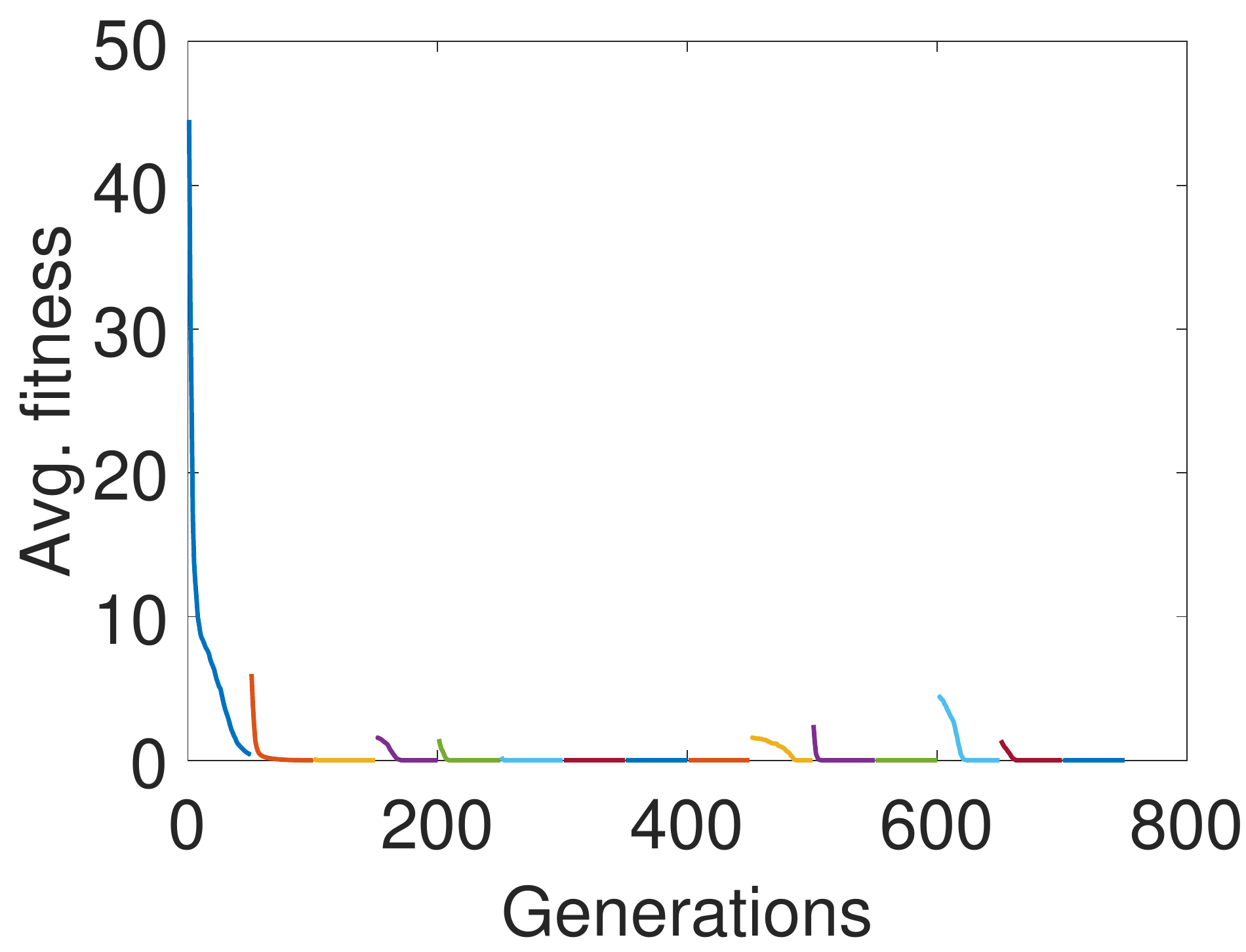}
        \caption{Classic Boids}
        \label{Fig:OnlineClassic2step}
    \end{subfigure} 
    \begin{subfigure}[b]{0.24\textwidth}
        \includegraphics[width=\textwidth,height=3.2cm]{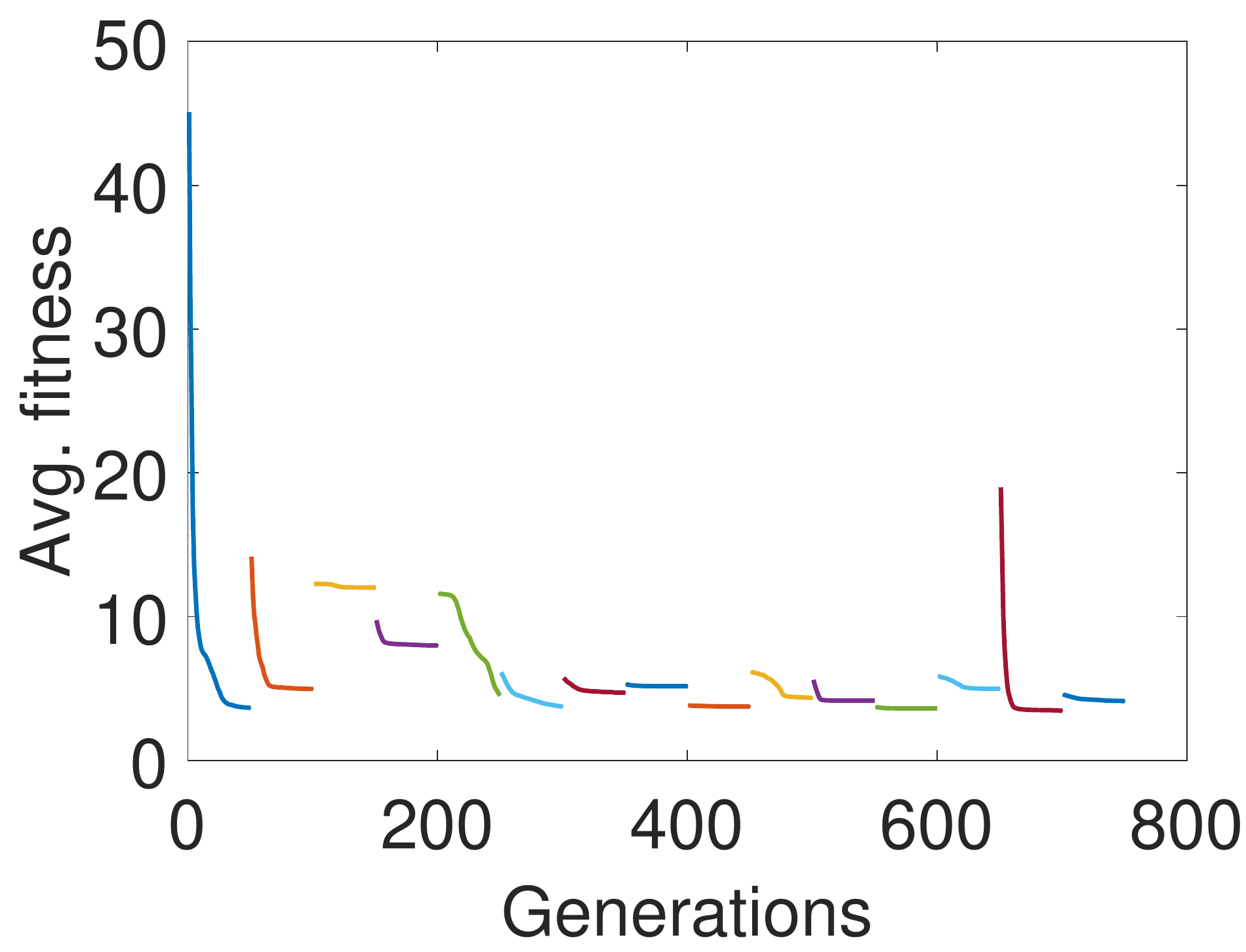}
        \caption{Erd\H{o}s\textendash R{\'e}nyi network}
        \label{Fig:OnlineERNet2step}
    \end{subfigure} 
    \begin{subfigure}[b]{0.24\textwidth}
        \includegraphics[width=\textwidth,height=3.2cm]{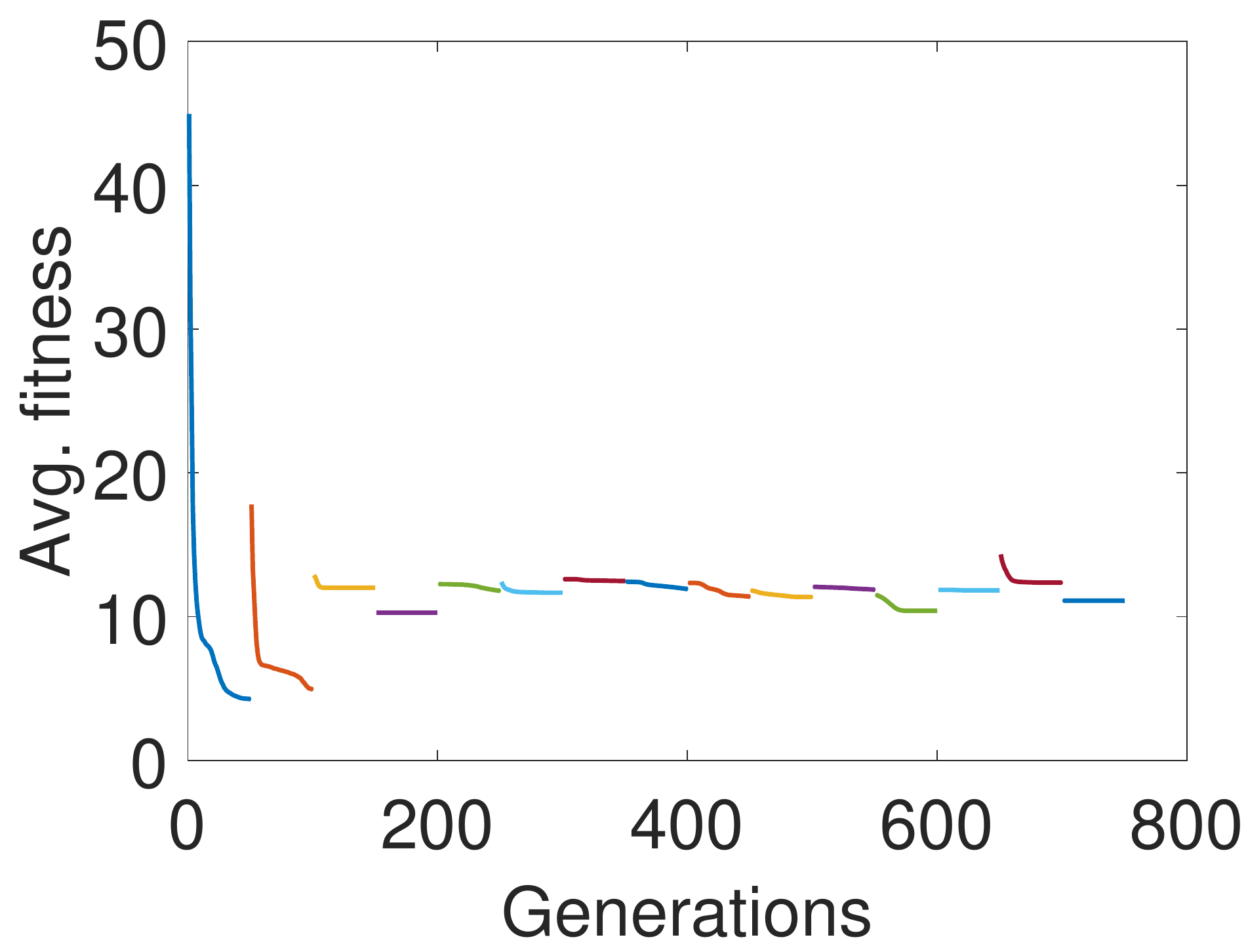}
        \caption{Scale-free network}
        \label{Fig:OnlineScaleFree2step}
    \end{subfigure} 
    \begin{subfigure}[b]{0.24\textwidth}
        \includegraphics[width=\textwidth,height=3.2cm]{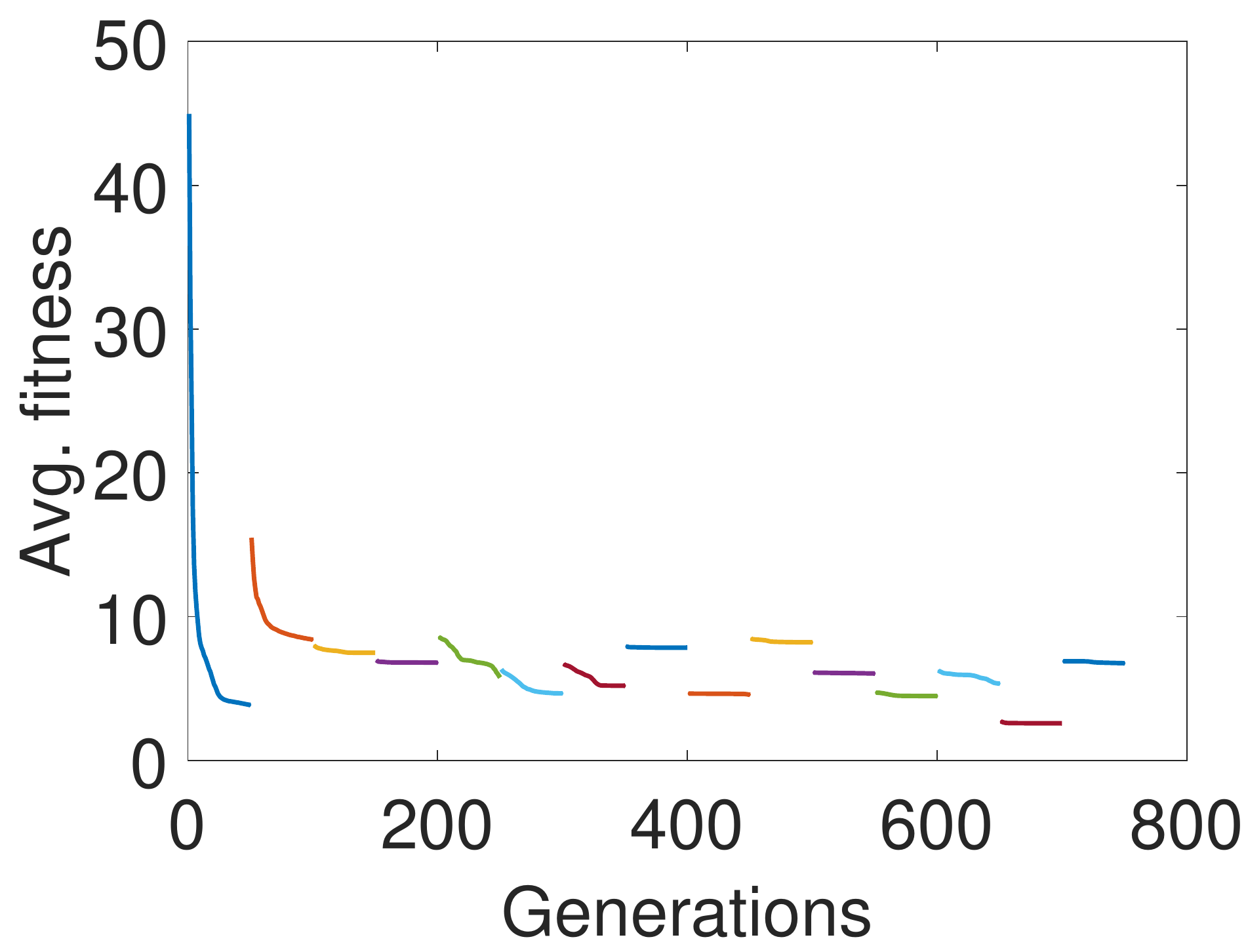}
        \caption{Small-world network}
        \label{Fig:OnlineSmallWorld2step}
    \end{subfigure}
    \caption{Average fitness evolution of 10 online  runs for each communication method when the observation period is 2 iterations.}
    \label{Fig:OnlineDEEvo2Step}
\end{figure*}

\begin{figure*}[h]
    \center
    \begin{subfigure}[b]{0.24\textwidth}
        \includegraphics[width=\textwidth,height=3.2cm]{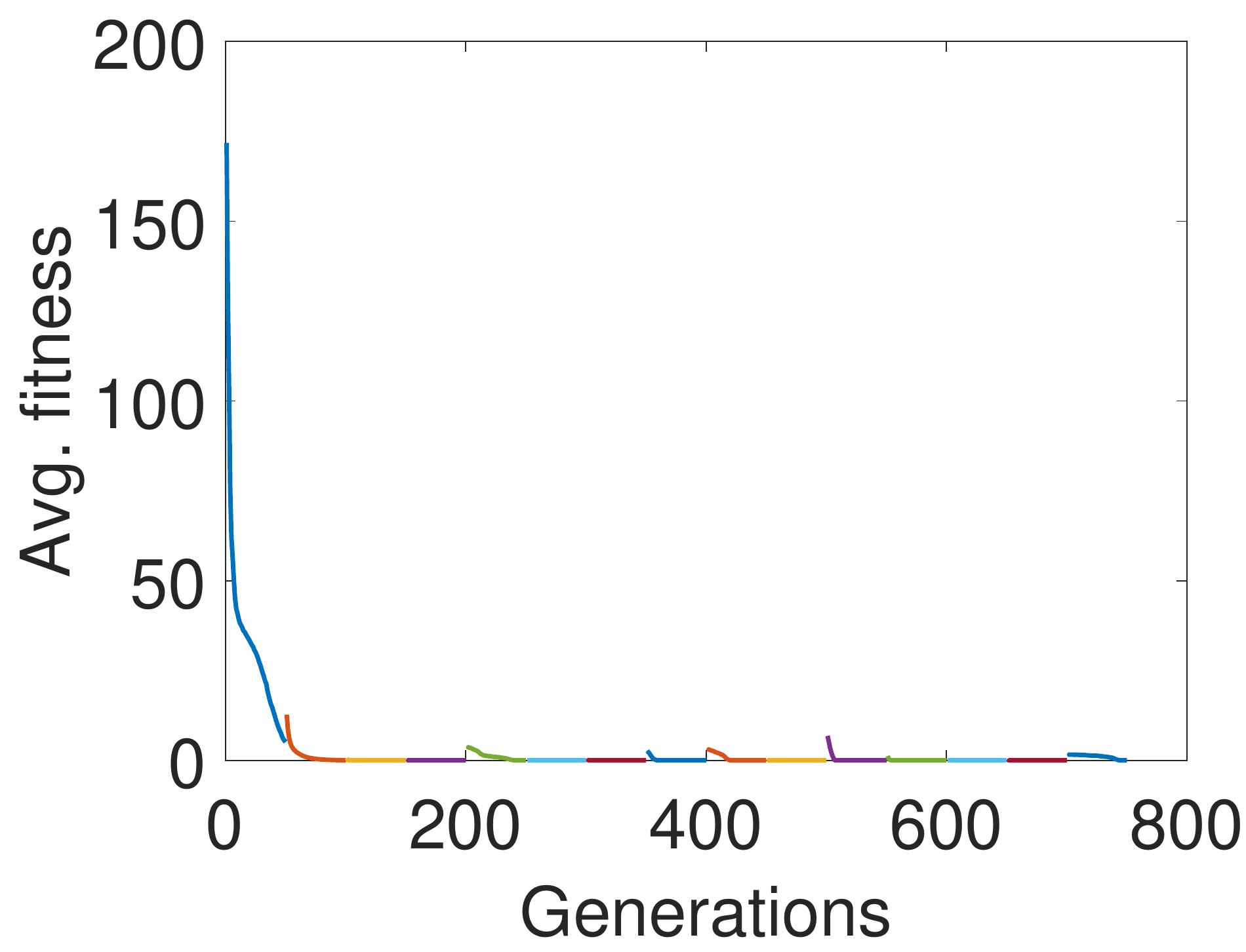}
        \caption{Classic Boids}
        \label{Fig:OnlineClassic4step}
    \end{subfigure} 
    \begin{subfigure}[b]{0.24\textwidth}
        \includegraphics[width=\textwidth,height=3.2cm]{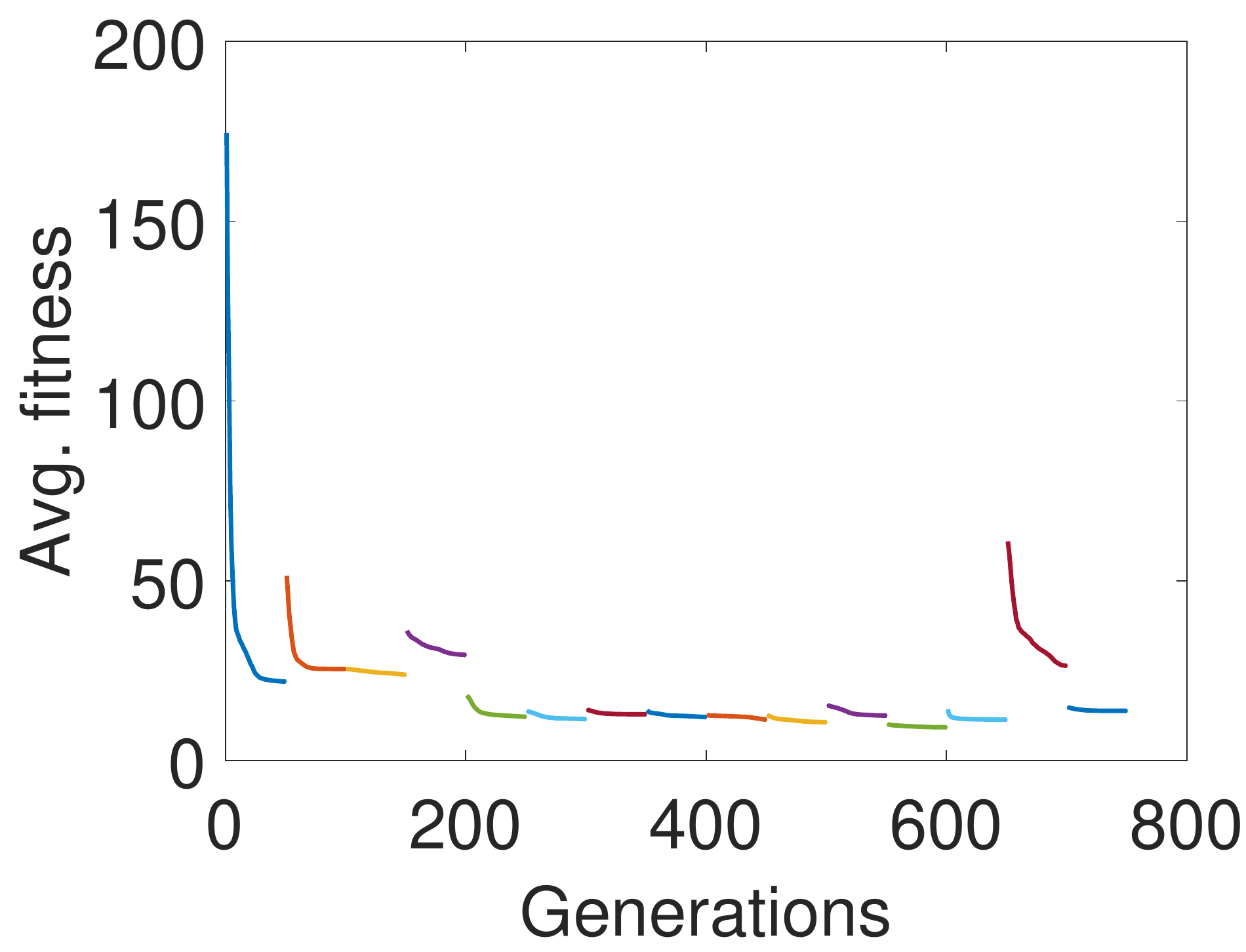}
        \caption{Erd\H{o}s\textendash R{\'e}nyi network}
        \label{Fig:OnlineERNet4step}
    \end{subfigure} 
    \begin{subfigure}[b]{0.24\textwidth}
        \includegraphics[width=\textwidth,height=3.2cm]{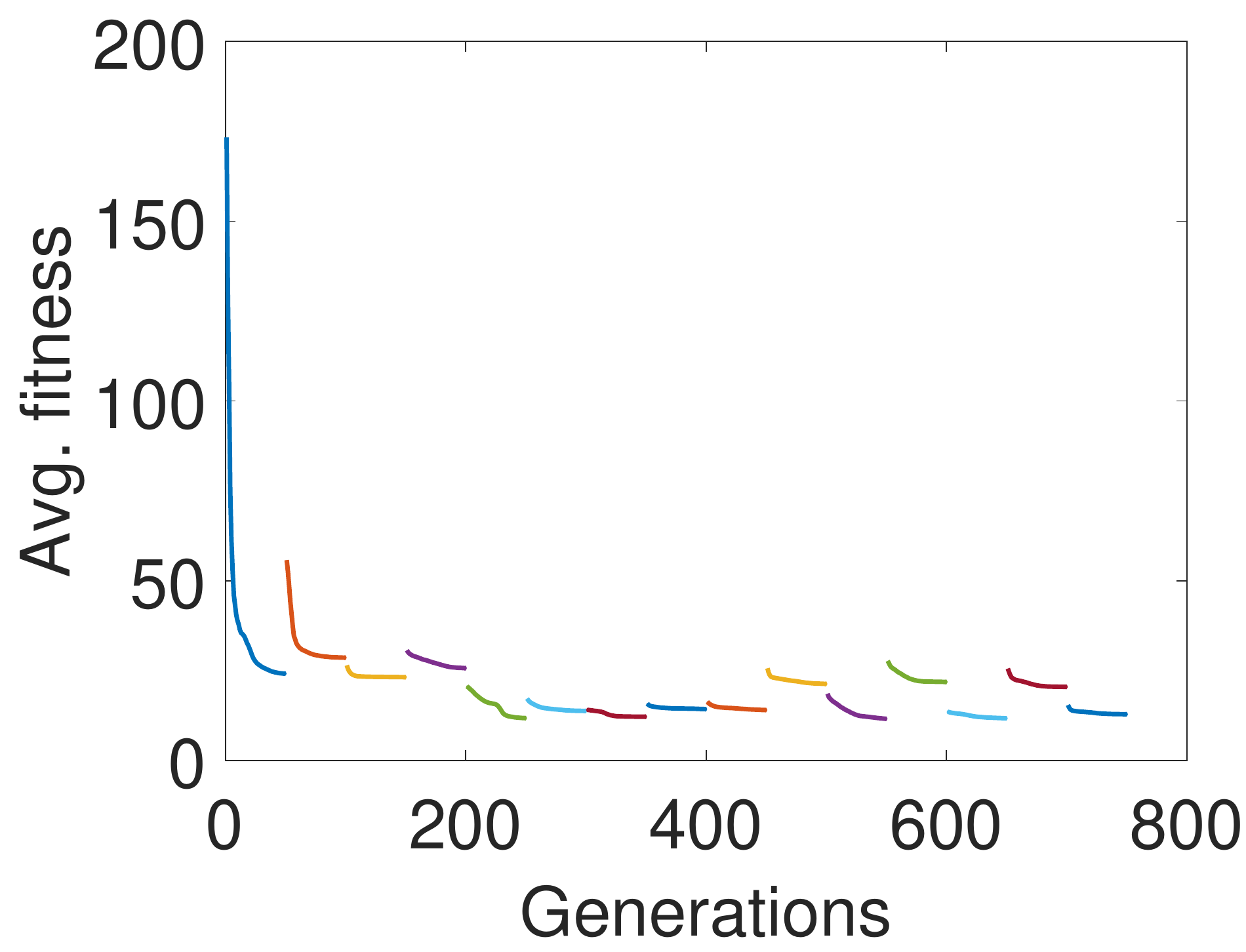}
        \caption{Scale-free network}
        \label{Fig:OnlineScaleFree4step}
    \end{subfigure} 
    \begin{subfigure}[b]{0.24\textwidth}
        \includegraphics[width=\textwidth,height=3.2cm]{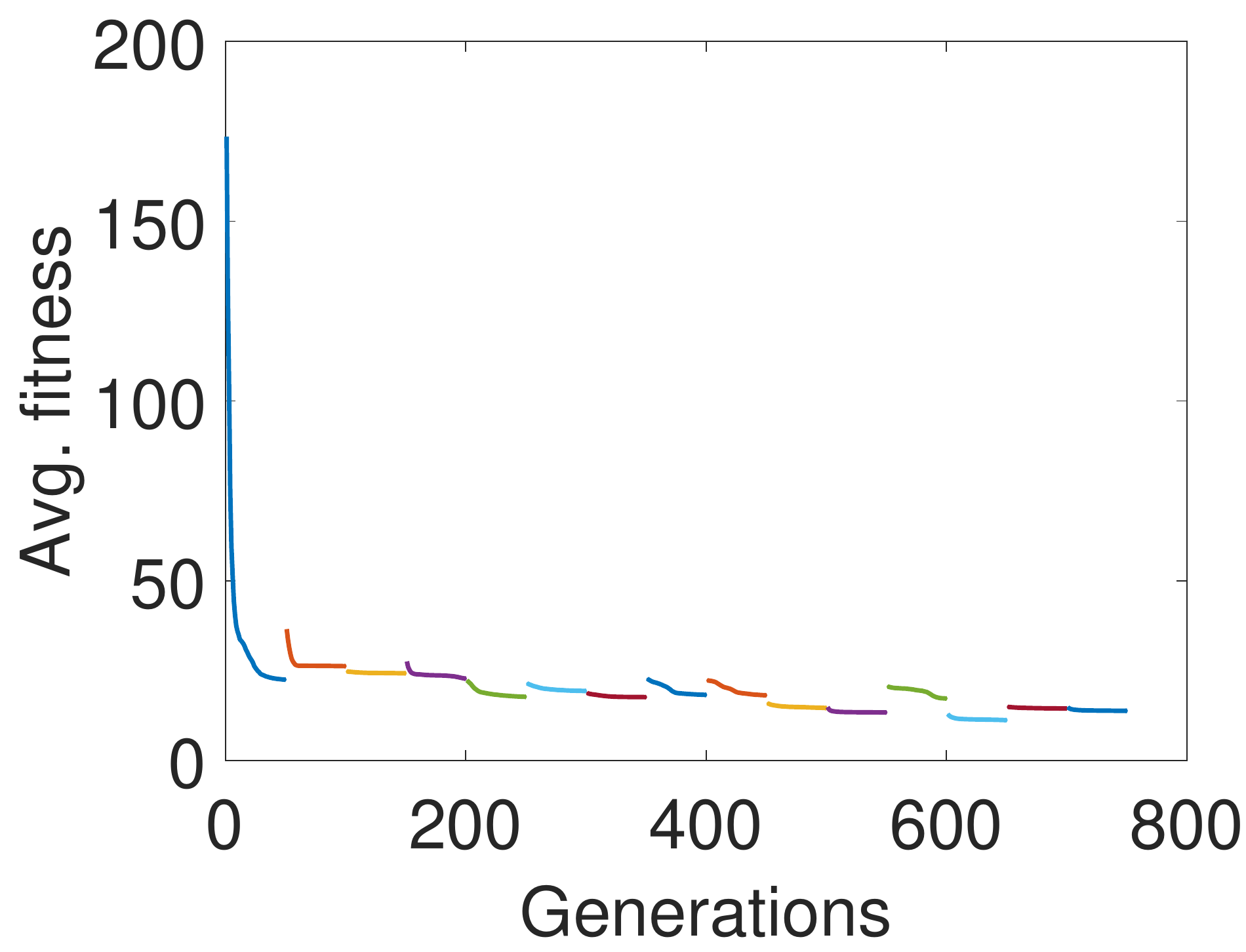}
        \caption{Small-world network}
        \label{Fig:OnlineSmallWorld4step}
    \end{subfigure}
    \caption{Average fitness evolution of 10 online  runs for each communication method when the observation period is 4 iterations.}
    \label{Fig:OnlineDEEvo4Step}
\end{figure*}

\begin{figure*}[h]
    \center
    \begin{subfigure}[b]{0.24\textwidth}
        \includegraphics[width=\textwidth,height=3.2cm]{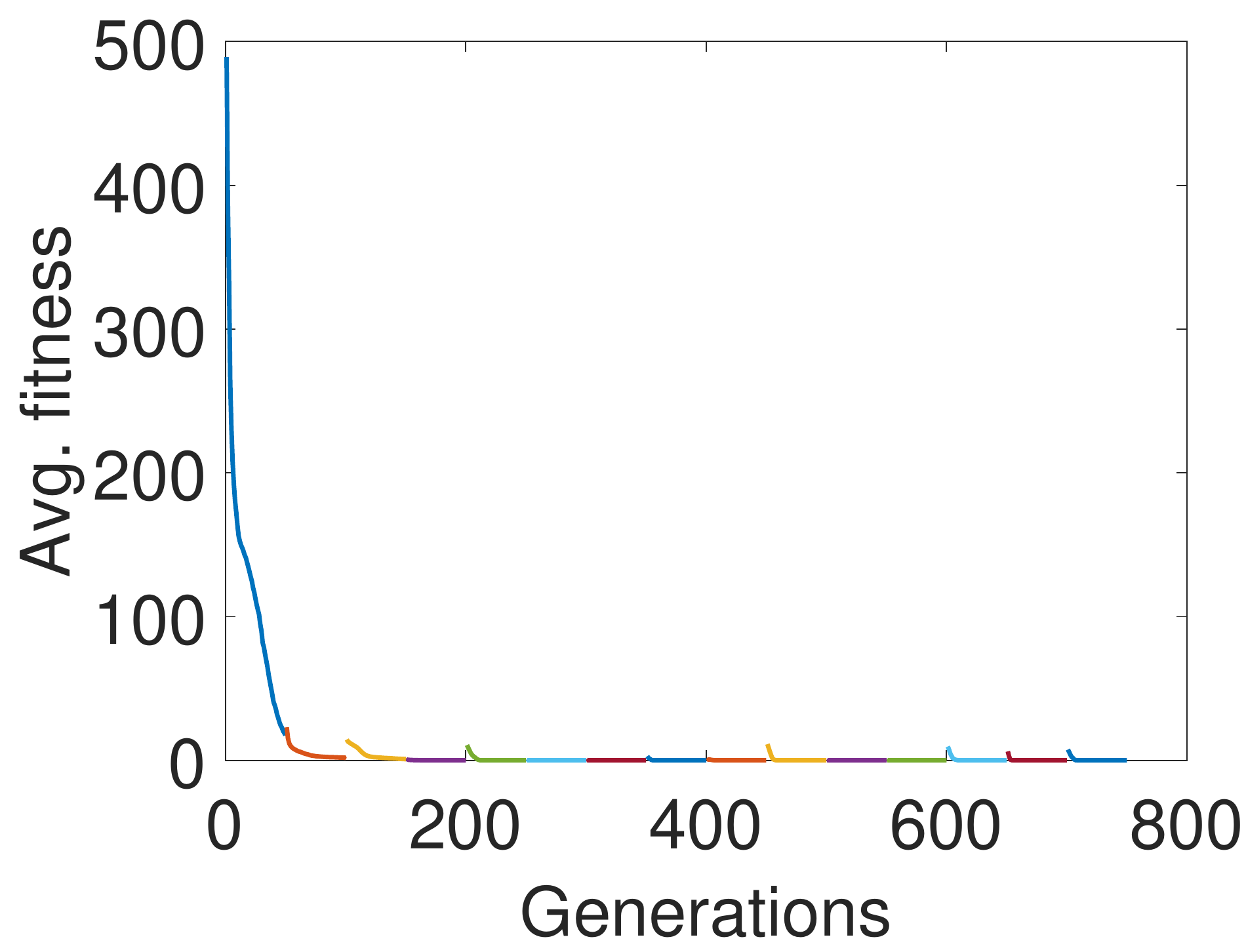}
        \caption{Classic Boids}
        \label{Fig:OnlineClassic8step}
    \end{subfigure} 
    \begin{subfigure}[b]{0.24\textwidth}
        \includegraphics[width=\textwidth,height=3.2cm]{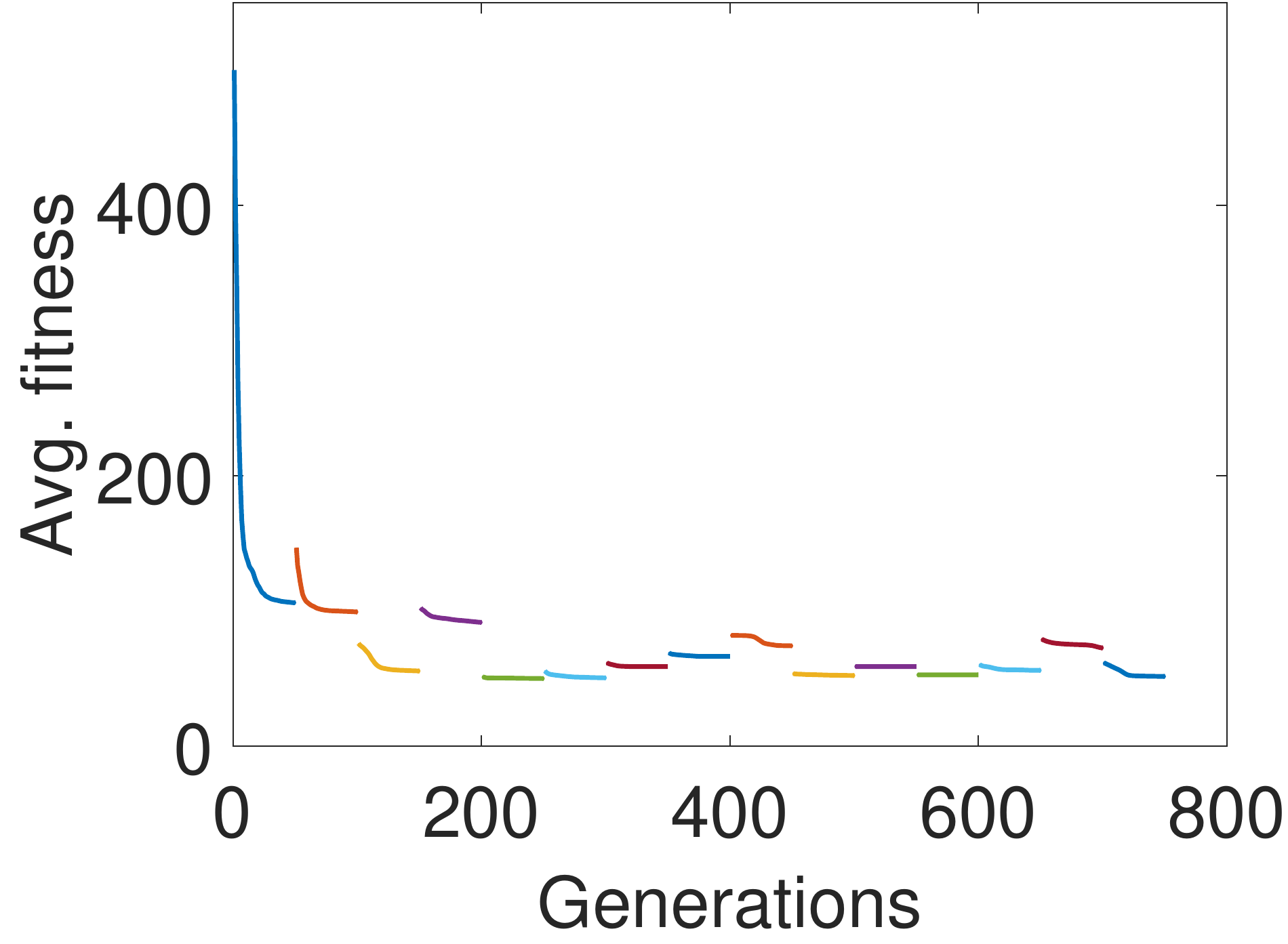}
        \caption{Erd\H{o}s\textendash R{\'e}nyi network}
        \label{Fig:OnlineERNet8step}
    \end{subfigure} 
    \begin{subfigure}[b]{0.24\textwidth}
        \includegraphics[width=\textwidth,height=3.2cm]{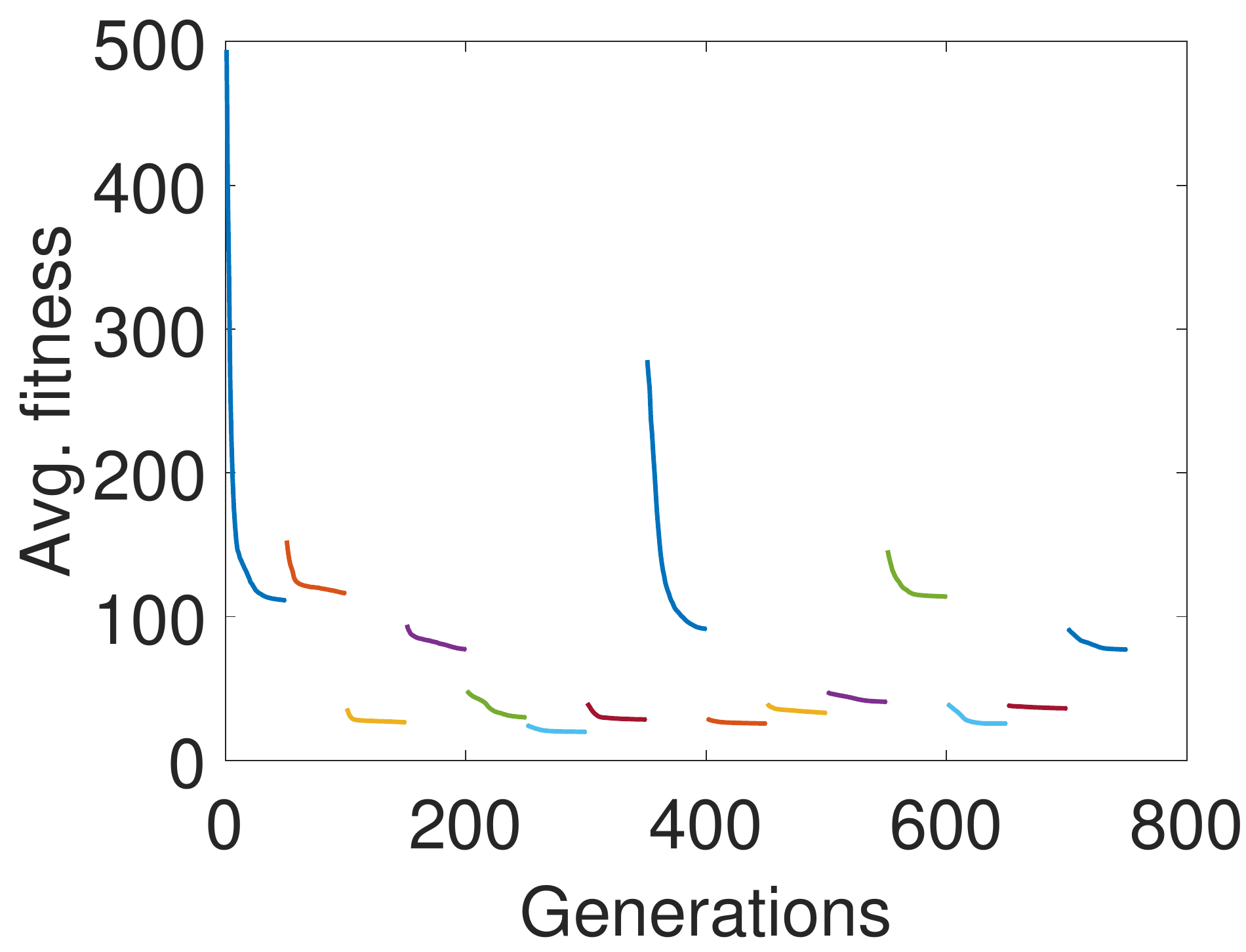}
        \caption{Scale-free network}
        \label{Fig:OnlineScaleFree8step}
    \end{subfigure} 
    \begin{subfigure}[b]{0.24\textwidth}
        \includegraphics[width=\textwidth,height=3.2cm]{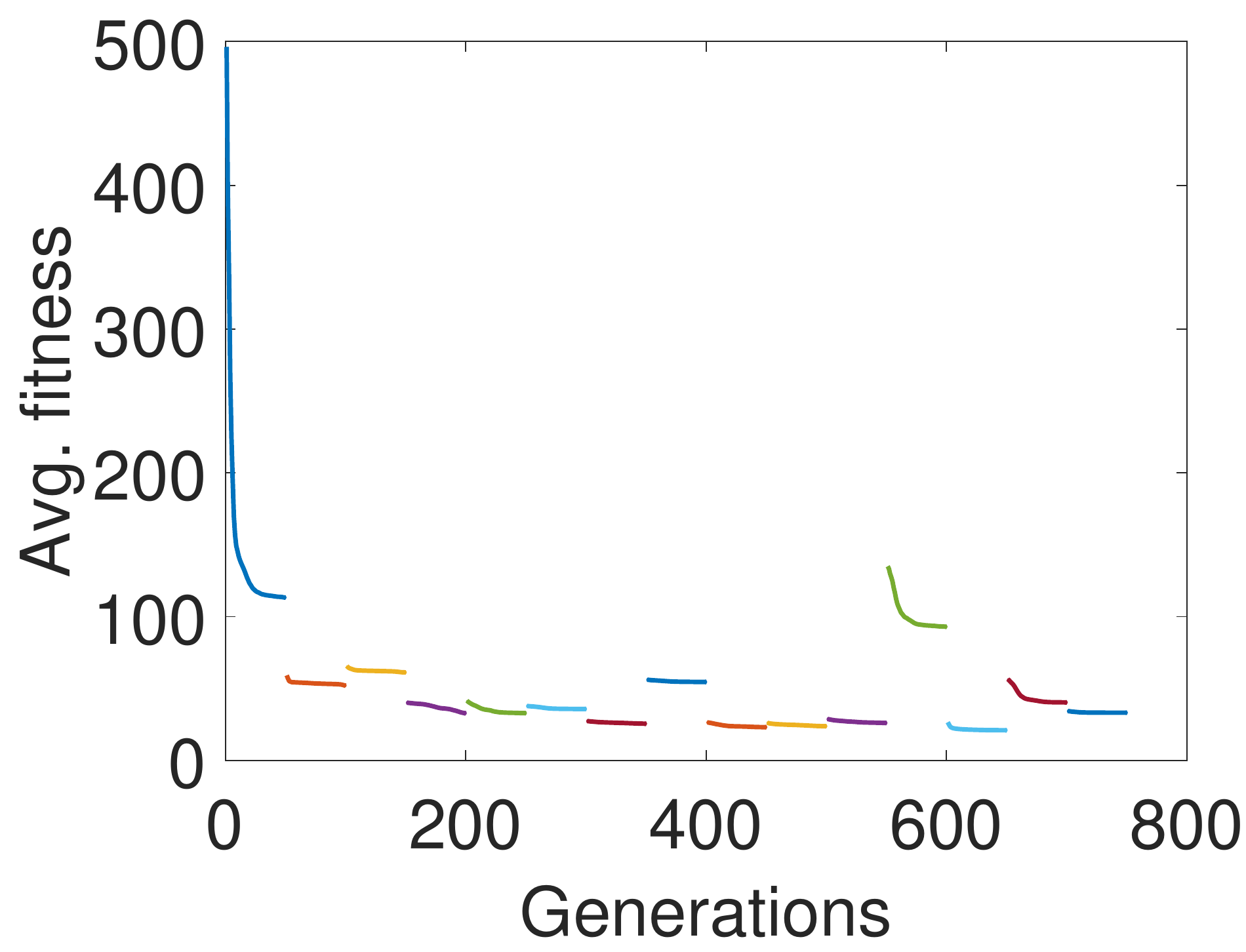}
        \caption{Small-world network}
        \label{Fig:OnlineSmallWorld8step}
    \end{subfigure}
    \caption{Average fitness evolution of 10 online runs for each communication method when the observation period is 8 iterations.}
    \label{Fig:OnlineDEEvo8Step}
\end{figure*}

\begin{figure*}[h]
    \center
    \begin{subfigure}[b]{0.24\textwidth}
        \includegraphics[width=\textwidth,height=3.2cm]{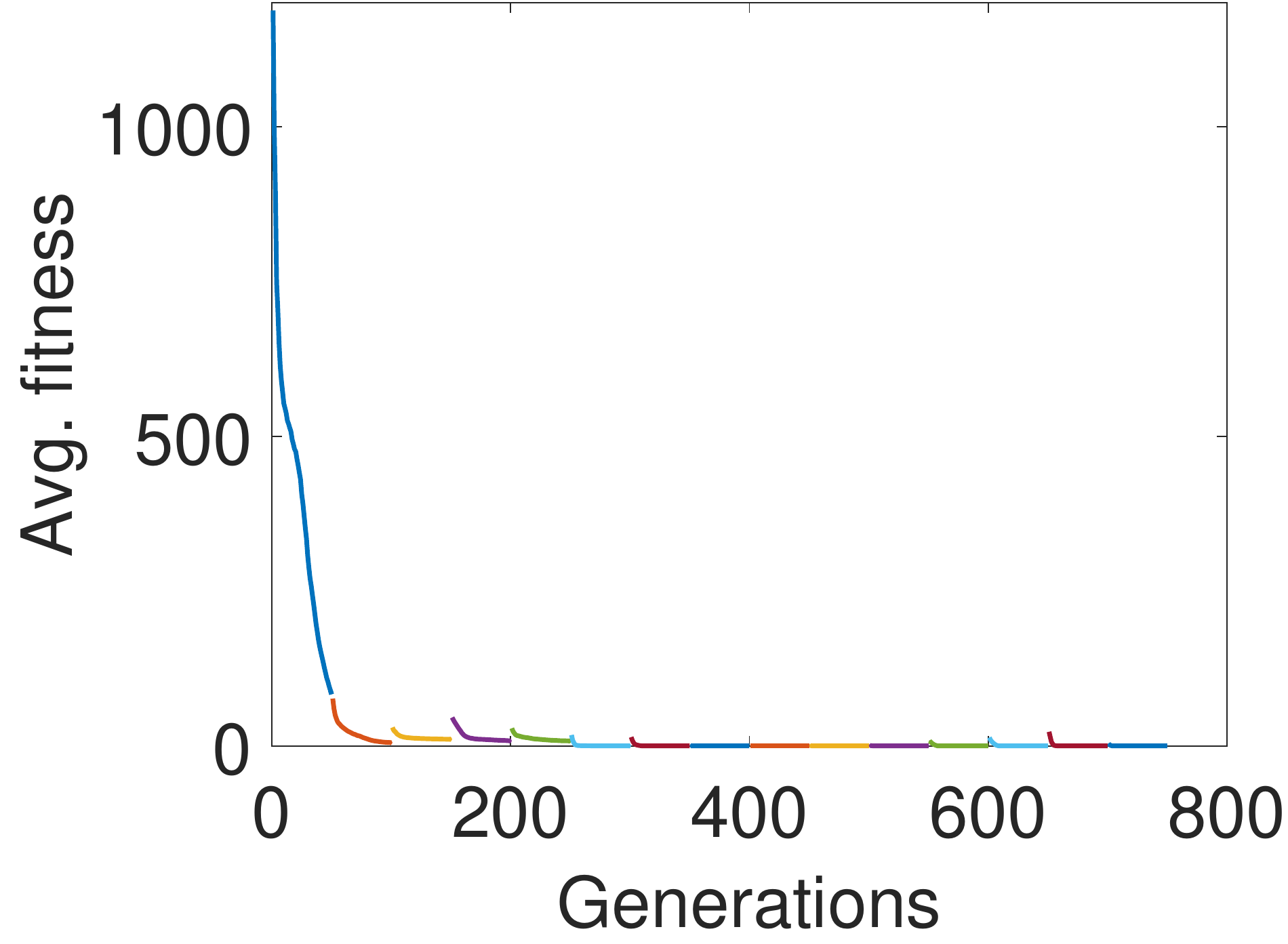}
        \caption{Classic Boids}
        \label{Fig:OnlineClassic16step}
    \end{subfigure} 
    \begin{subfigure}[b]{0.24\textwidth}
        \includegraphics[width=\textwidth,height=3.2cm]{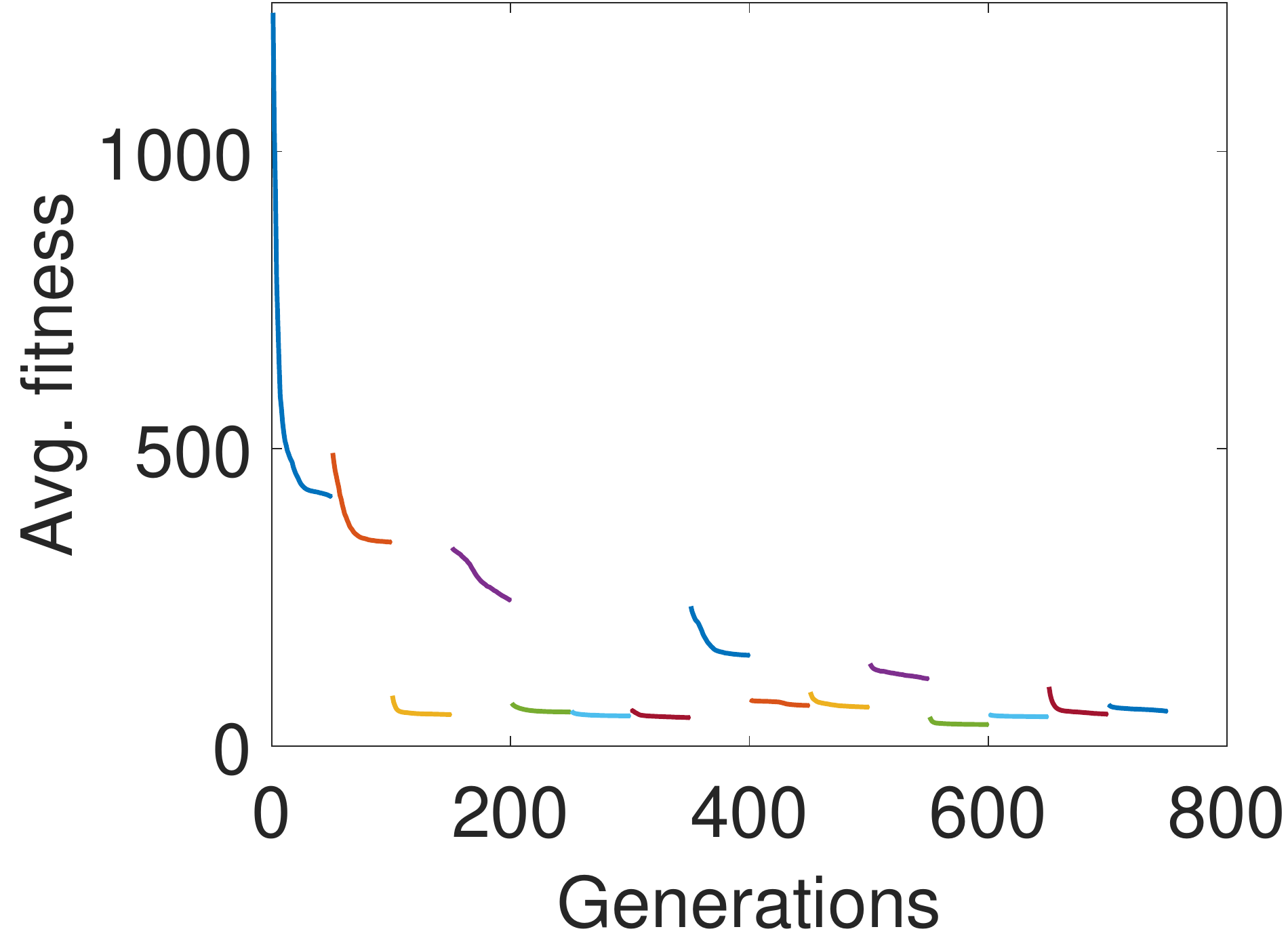}
        \caption{Erd\H{o}s\textendash R{\'e}nyi network}
        \label{Fig:OnlineERNet16step}
    \end{subfigure} 
    \begin{subfigure}[b]{0.24\textwidth}
        \includegraphics[width=\textwidth,height=3.2cm]{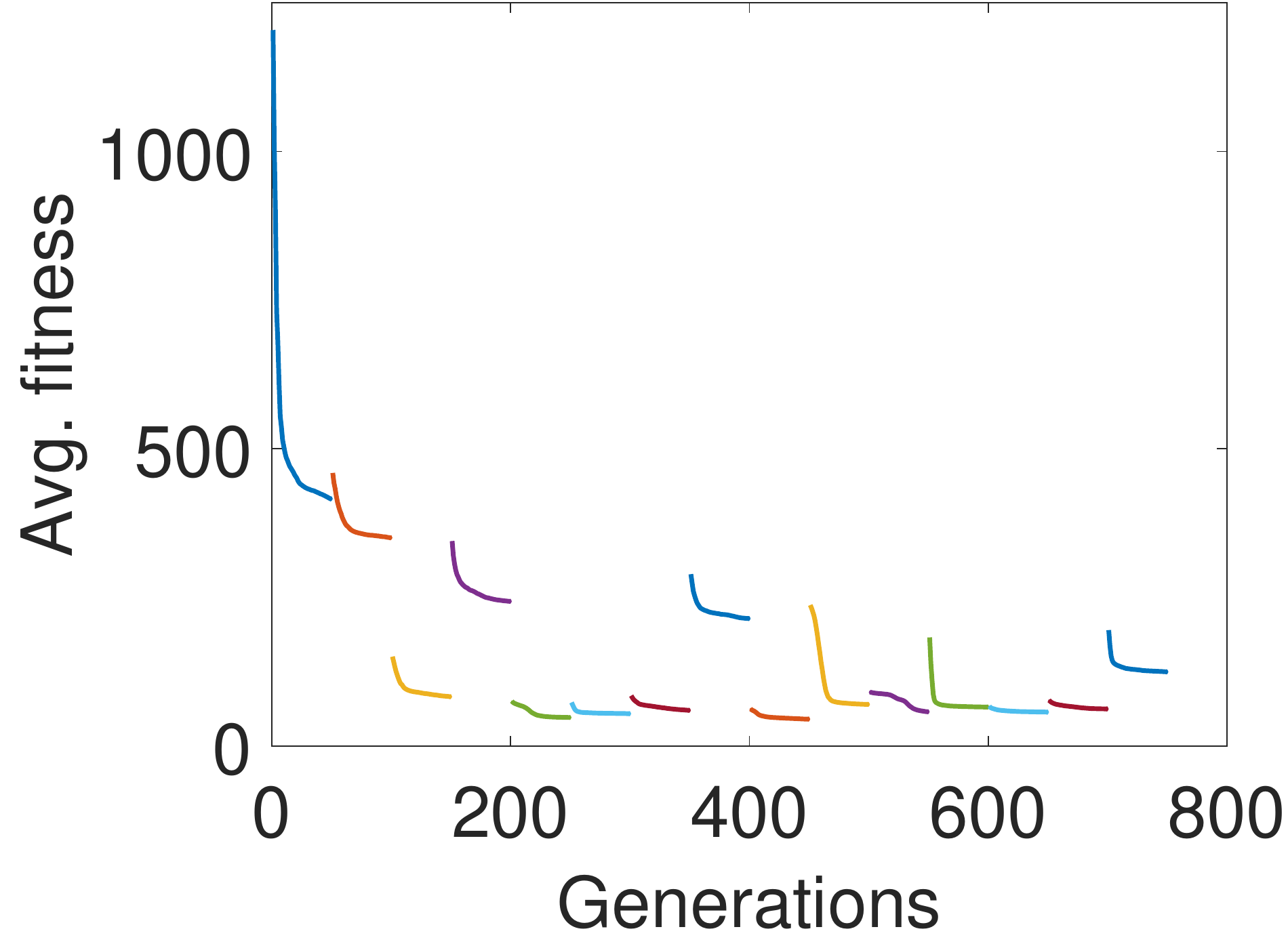}
        \caption{Scale-free network}
        \label{Fig:OnlineScaleFree16step}
    \end{subfigure} 
    \begin{subfigure}[b]{0.24\textwidth}
        \includegraphics[width=\textwidth,height=3.2cm]{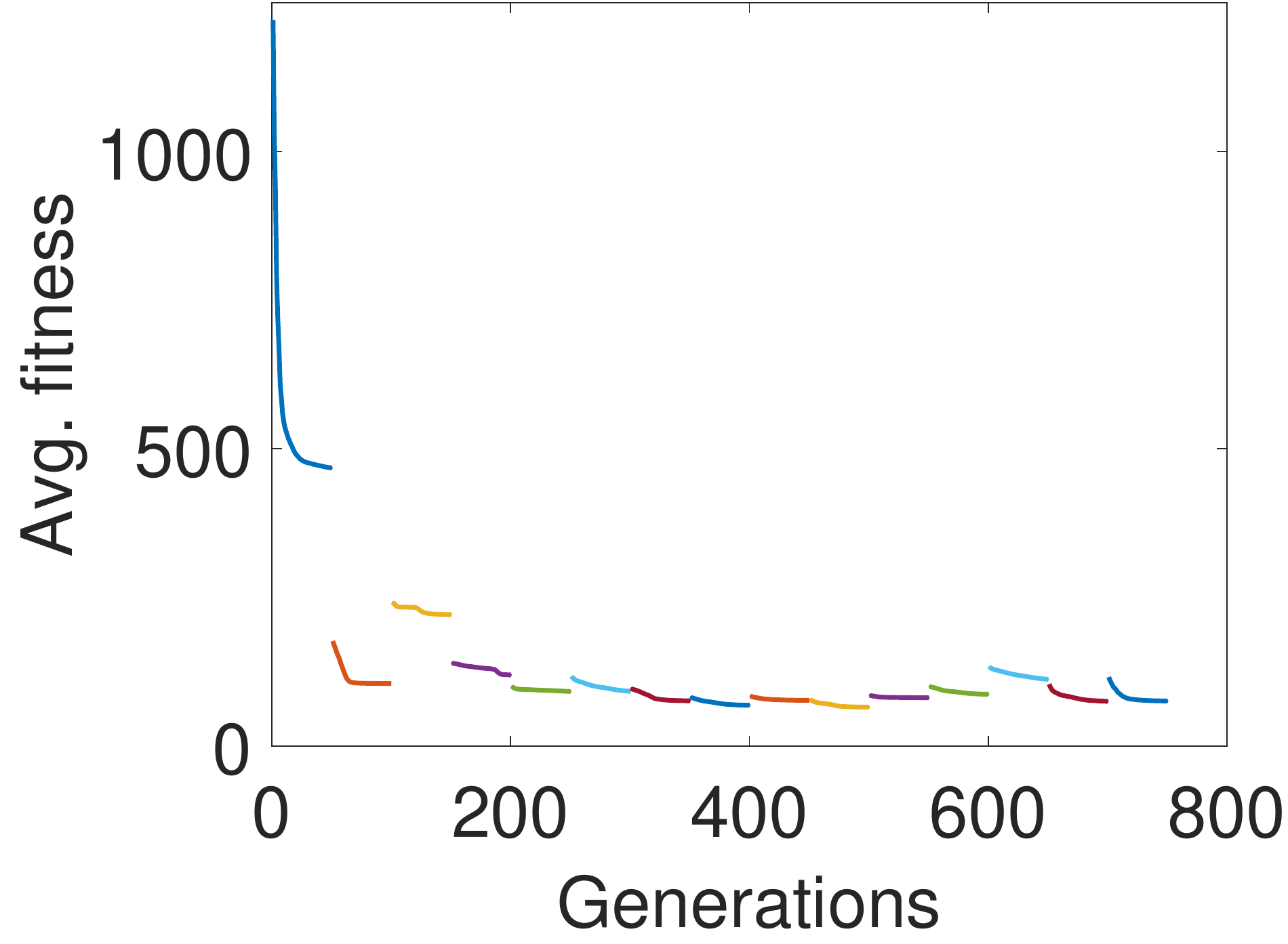}
        \caption{Small-world network}
        \label{Fig:OnlineSmallWorld16step}
    \end{subfigure}
    \caption{Average fitness evolution of 10 online runs for each communication method when the observation period is 16 iterations.}
    \label{Fig:OnlineDEEvo16Step}
\end{figure*}

Figures~\ref{Fig:OnlineDEEvo2Step},~\ref{Fig:OnlineDEEvo4Step},~\ref{Fig:OnlineDEEvo8Step}~and~\ref{Fig:OnlineDEEvo16Step} illustrate the average over 10 runs of the learning errors in each on-line learning-prediction cycle for all boid types and observation periods. It can be seen that the DE learning algorithm can always reduce the error to zero for classic boids in every learning cycle throughout the simulation. On the contrary, the algorithm is unable to reduce the error for any of the network-based boid types. In addition, for all network-based boid types we observe the same trend we observed in the off-line learning results, i.e. the learning performance actually decreases as the observation period increases. The reason for this is nevertheless the same like for the off-line learning case. Results suggest that the learner which targets vision-based parameters unaware of the existence of network-based neighborhoods will get more confused when observation periods increase, since an increased number of samples assumed to account for vision-based relations will approximate with less accuracy the actual network-based relations.

\begin{figure*}[h]
    \center
    \begin{subfigure}[b]{0.24\textwidth}
        \includegraphics[width=\textwidth,height=3.2cm]{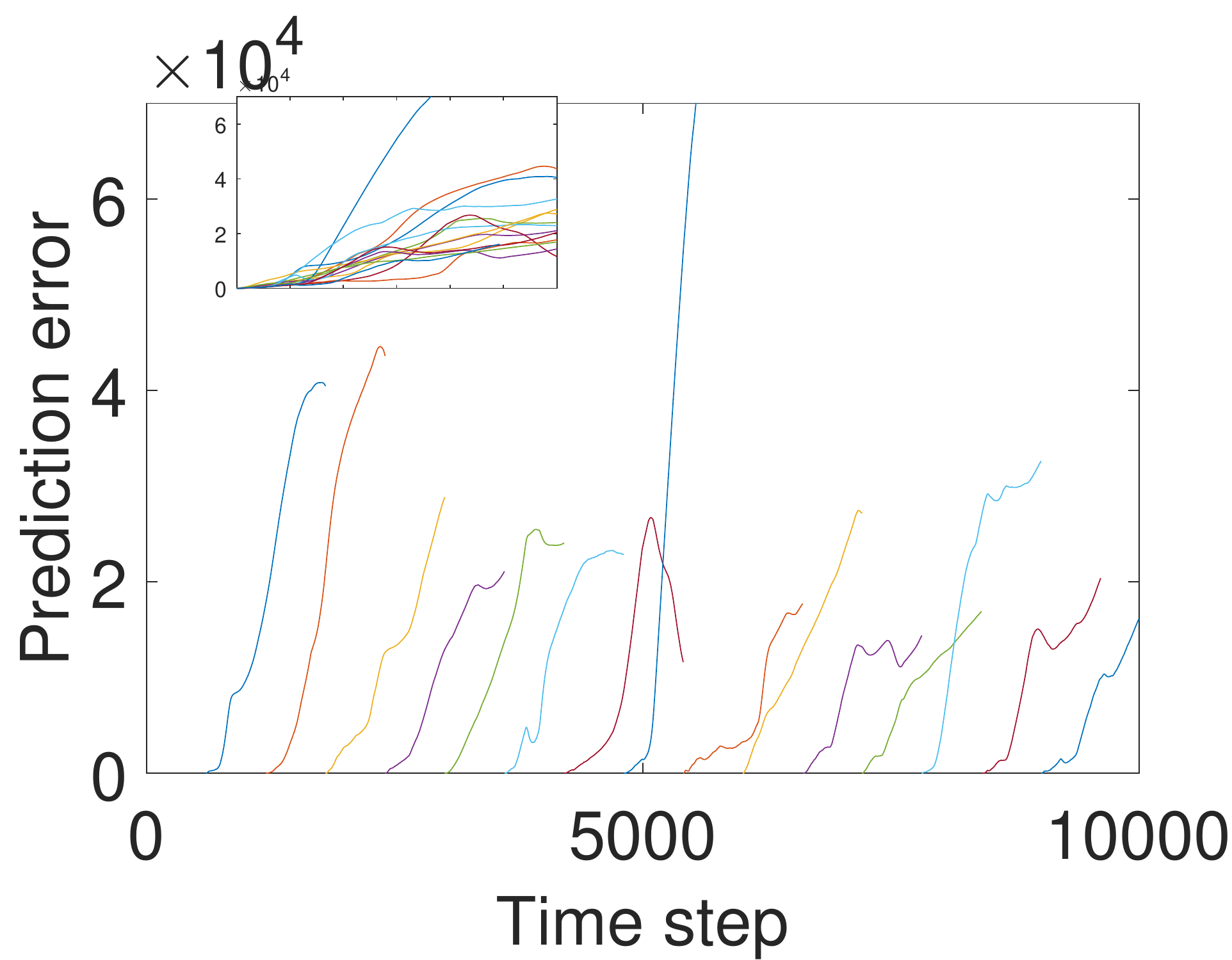}
        \caption{Classic boids}
        \label{Fig:PredictionClassic2step}
    \end{subfigure} 
    \begin{subfigure}[b]{0.24\textwidth}
        \includegraphics[width=\textwidth,height=3.2cm]{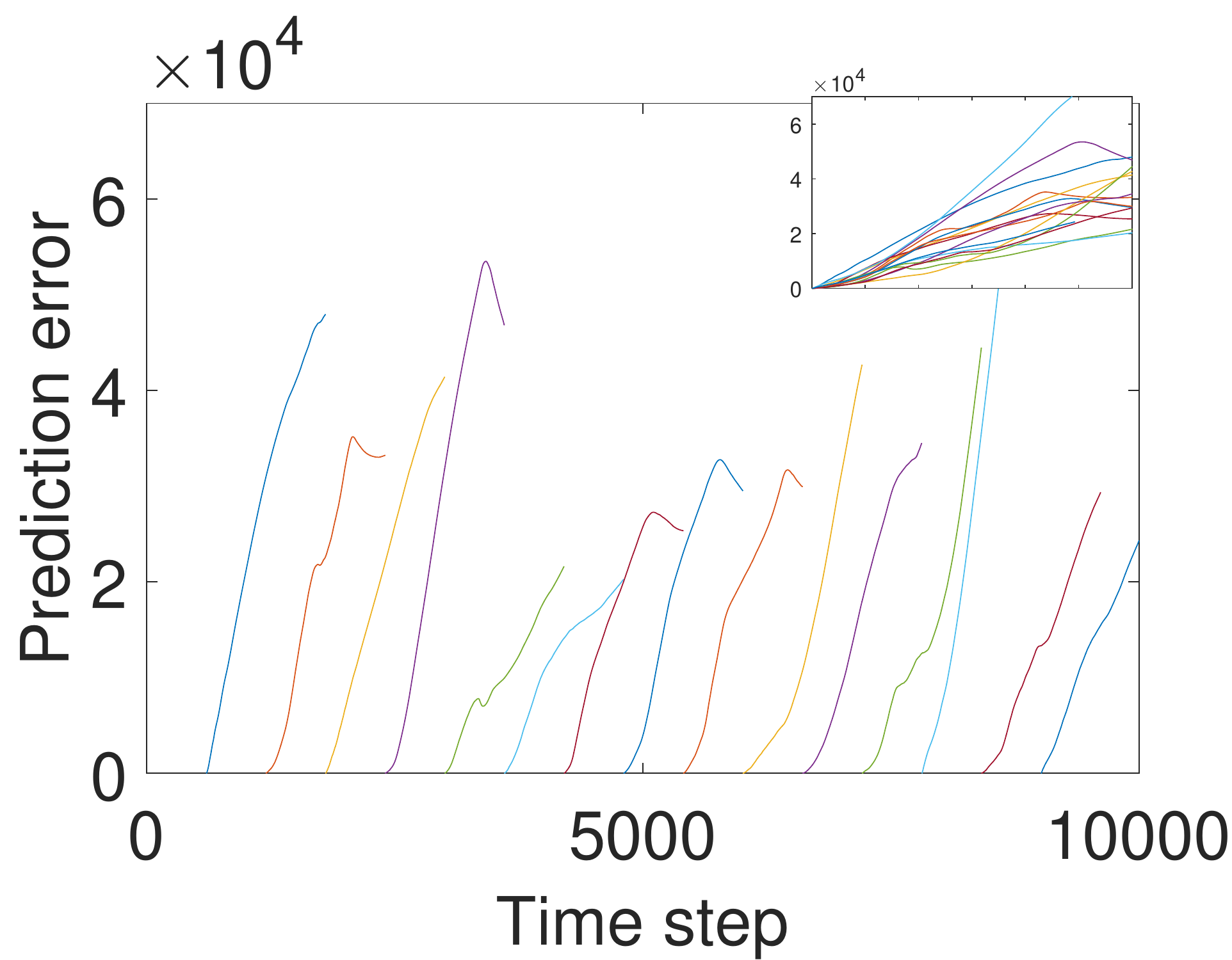}
        \caption{Erd\H{o}s\textendash R{\'e}nyi network}
        \label{Fig:PredictionERNet2step}
    \end{subfigure} 
    \begin{subfigure}[b]{0.24\textwidth}
        \includegraphics[width=\textwidth,height=3.2cm]{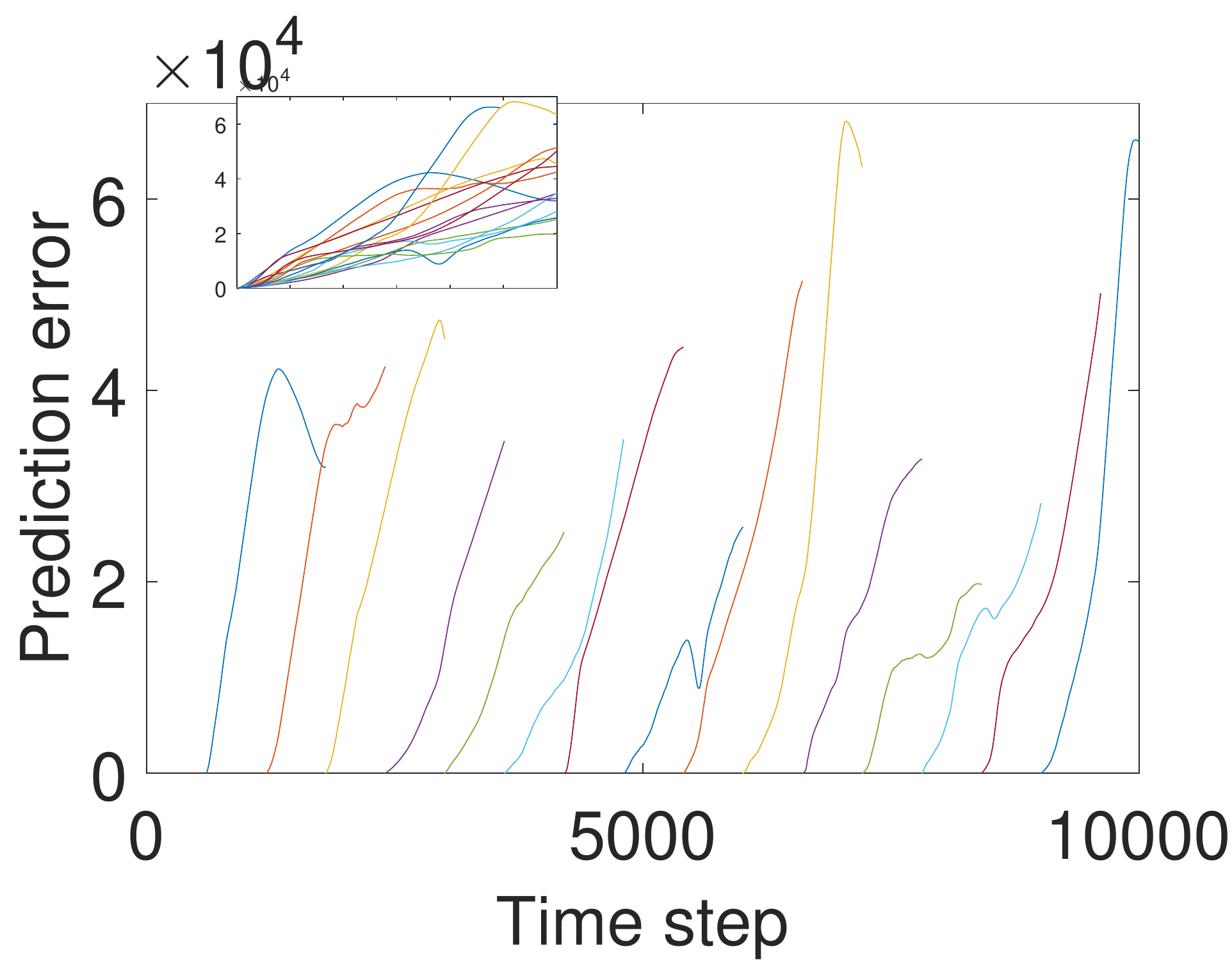}
        \caption{Scale-free network}
        \label{Fig:PredictionScaleFree2step}
    \end{subfigure} 
    \begin{subfigure}[b]{0.24\textwidth}
        \includegraphics[width=\textwidth,height=3.2cm]{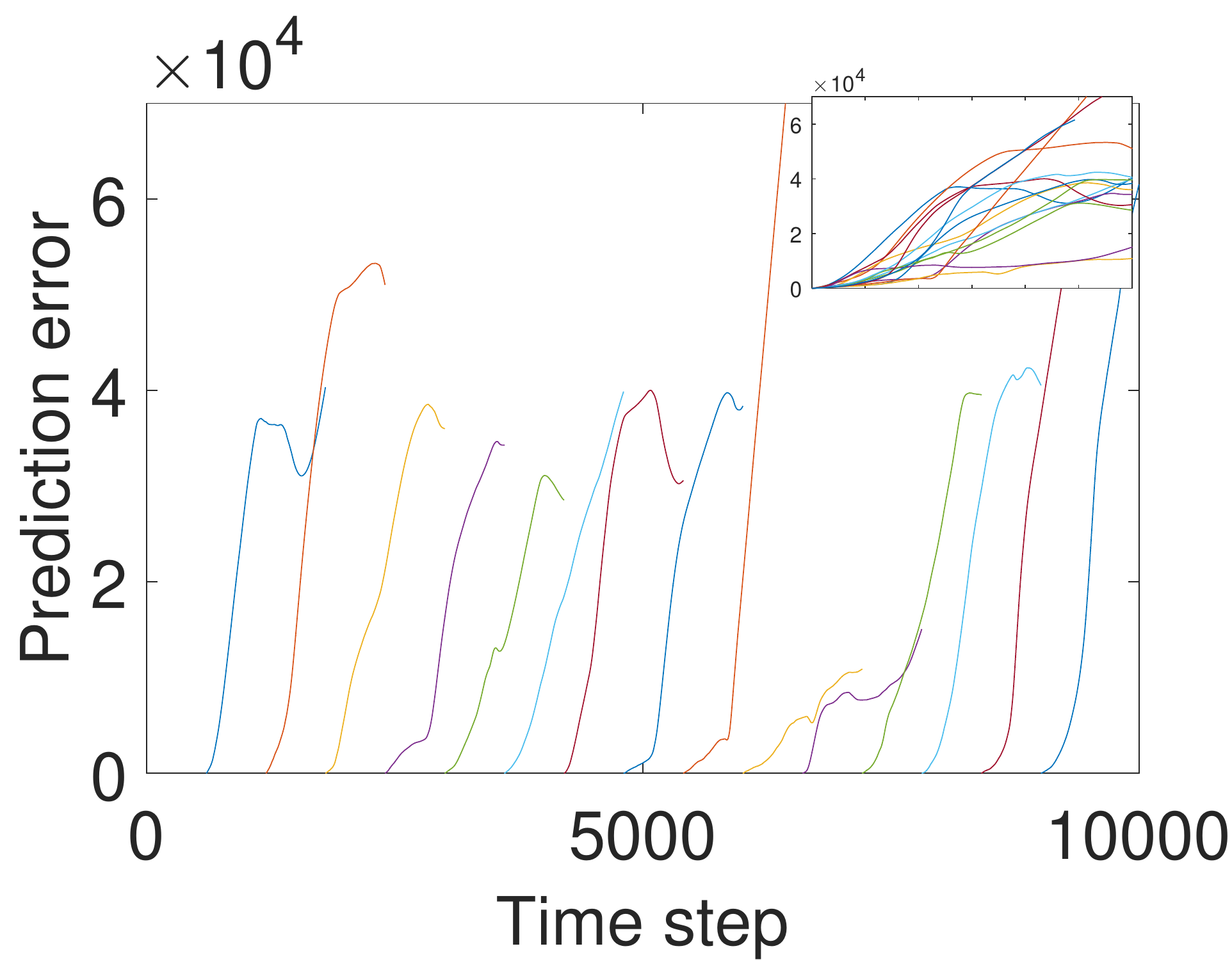}
        \caption{Small-world network}
        \label{Fig:PredictionSmallWorld2step}
    \end{subfigure}
    \caption{Average prediction errors (10 runs, top 5 individuals, observation period 2, online learning) for each communication method.}
    \label{Fig:Prediction2Step}
\end{figure*}

\begin{figure*}[h]
    \center
    \begin{subfigure}[b]{0.24\textwidth}
        \includegraphics[width=\textwidth,height=3.2cm]{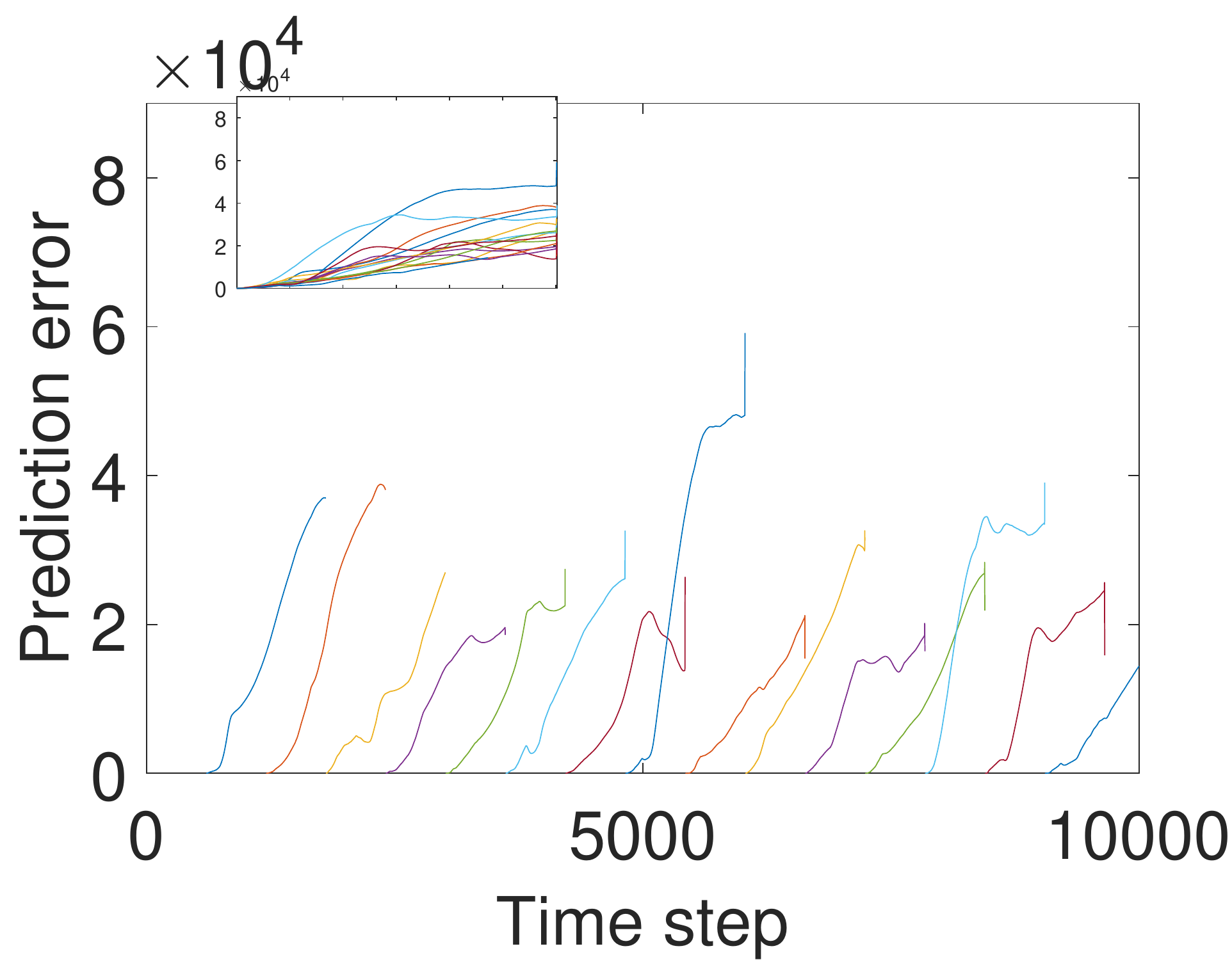}
        \caption{Classic boids}
        \label{Fig:PredictionClassic4step}
    \end{subfigure} 
    \begin{subfigure}[b]{0.24\textwidth}
        \includegraphics[width=\textwidth,height=3.2cm]{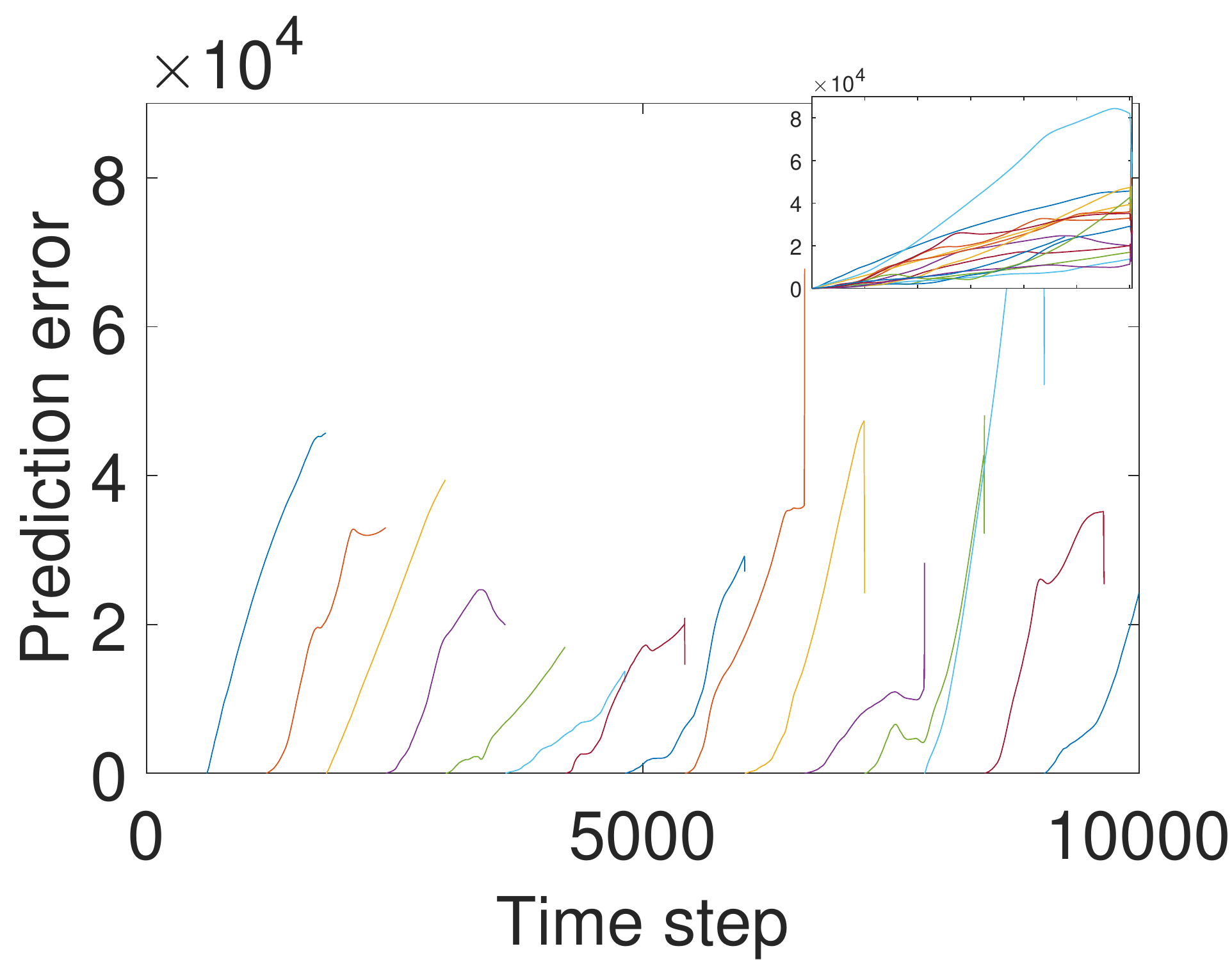}
        \caption{Erd\H{o}s\textendash R{\'e}nyi network}
        \label{Fig:PredictionERNet4step}
    \end{subfigure} 
    \begin{subfigure}[b]{0.24\textwidth}
        \includegraphics[width=\textwidth,height=3.2cm]{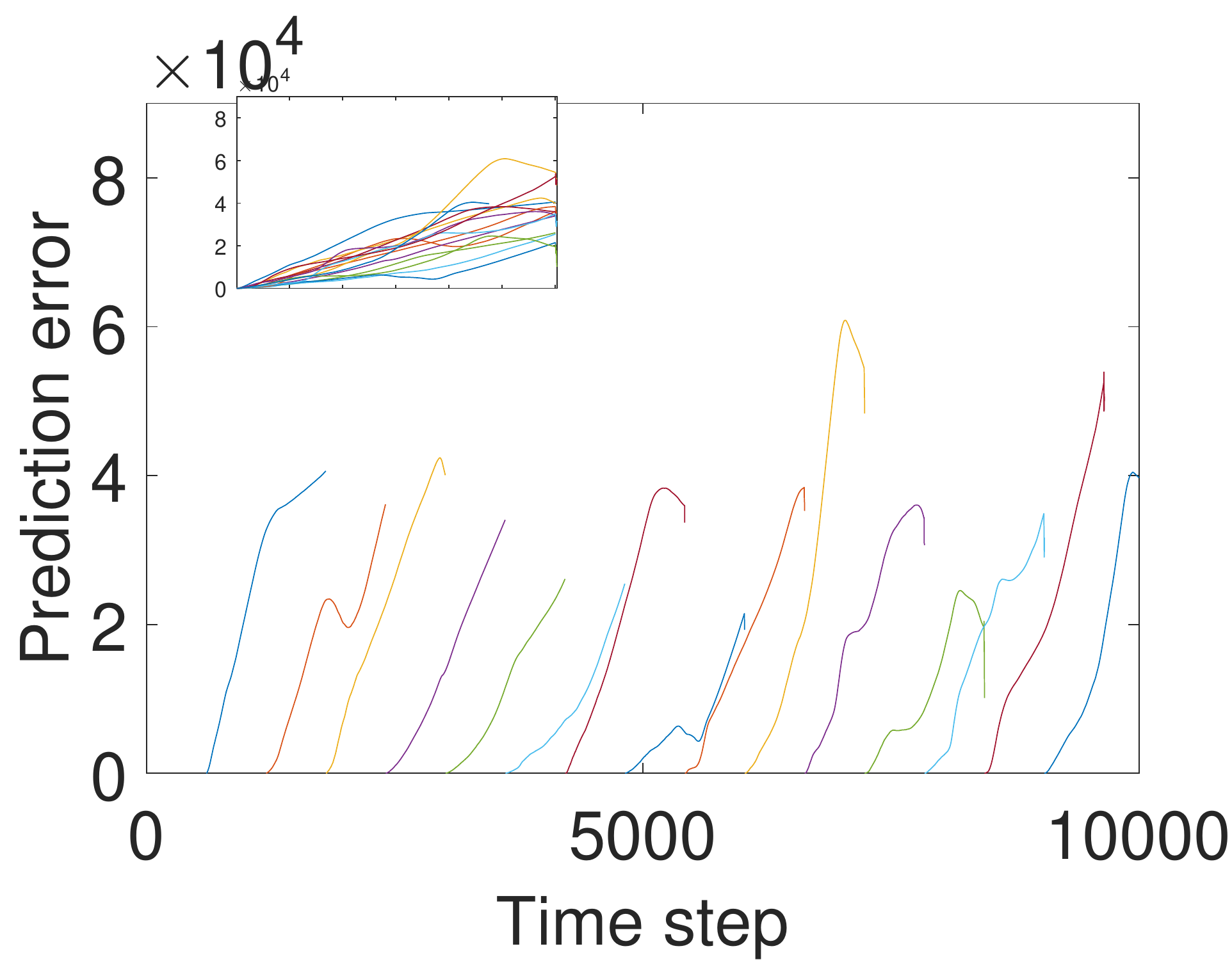}
        \caption{Scale-free network}
        \label{Fig:PredictionScaleFree4step}
    \end{subfigure} 
    \begin{subfigure}[b]{0.24\textwidth}
        \includegraphics[width=\textwidth,height=3.2cm]{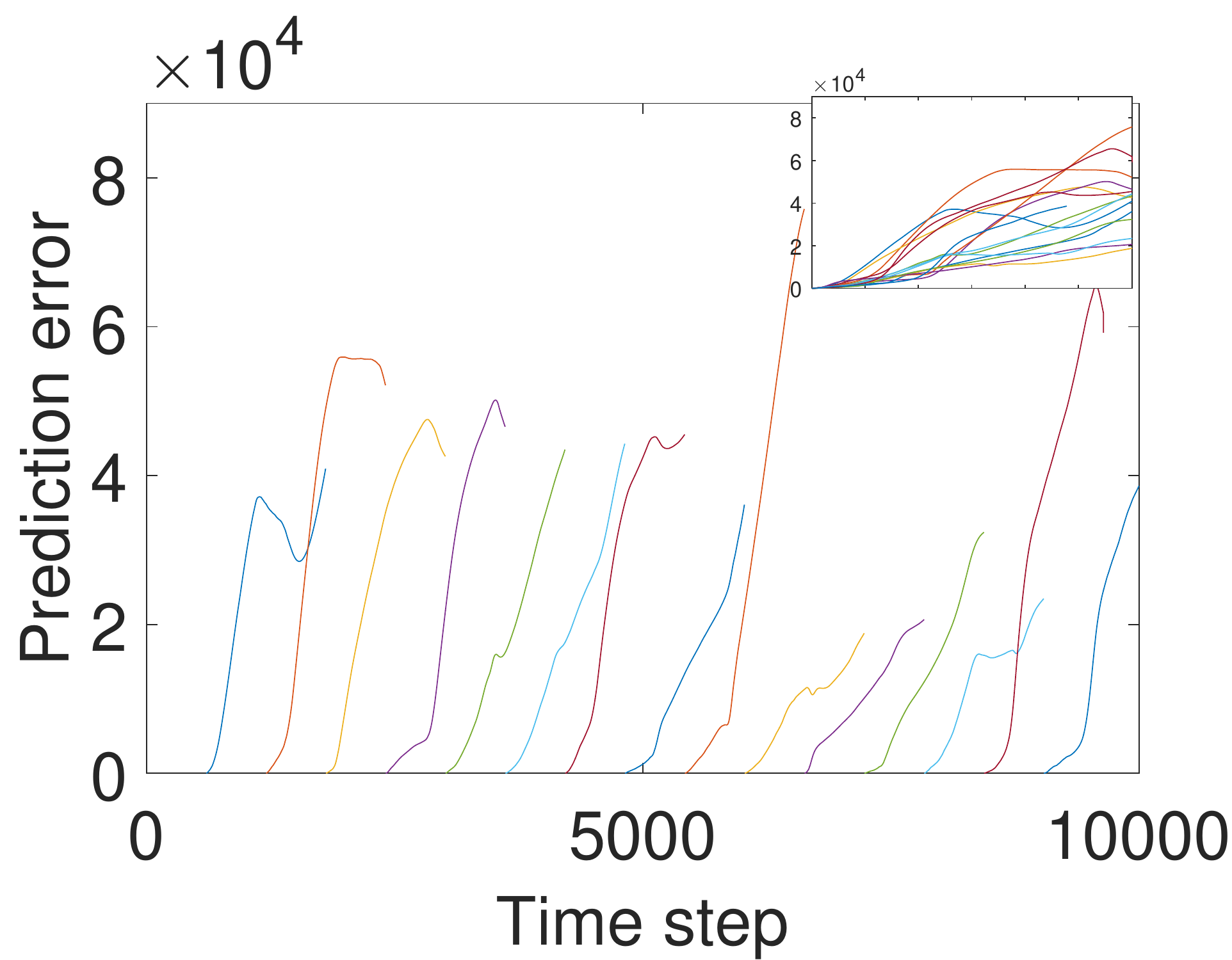}
        \caption{Small-world network}
        \label{Fig:PredictionSmallWorld4step}
    \end{subfigure}
    \caption{Average prediction errors (10 runs, top 5 individuals, observation period 4, online learning) for each communication method.}
    \label{Fig:Prediction4Step}
\end{figure*}

\begin{figure*}[h]
    \center
    \begin{subfigure}[b]{0.24\textwidth}
        \includegraphics[width=\textwidth,height=3.2cm]{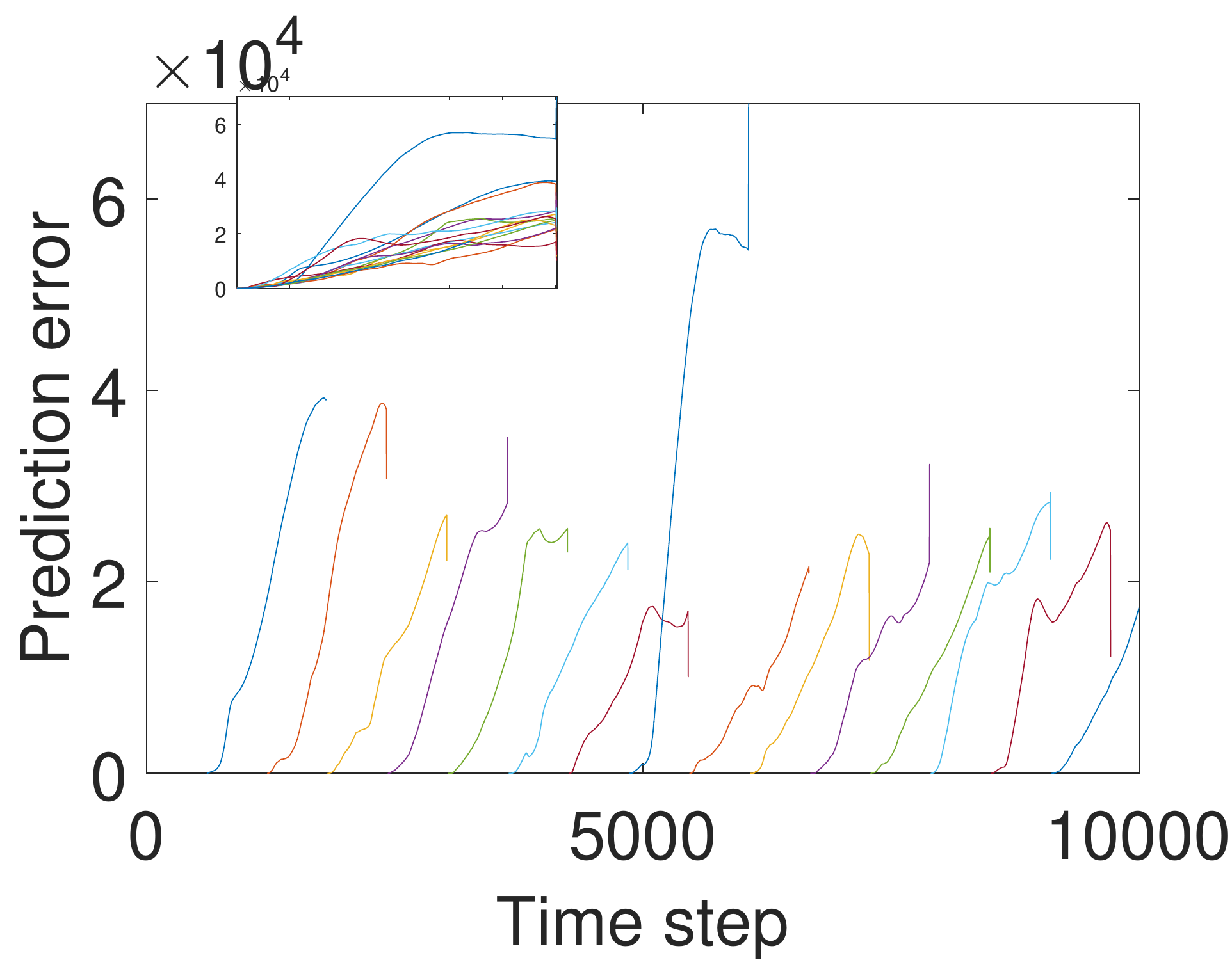}
        \caption{Classic boids}
        \label{Fig:PredictionClassic8step}
    \end{subfigure} 
    \begin{subfigure}[b]{0.24\textwidth}
        \includegraphics[width=\textwidth,height=3.2cm]{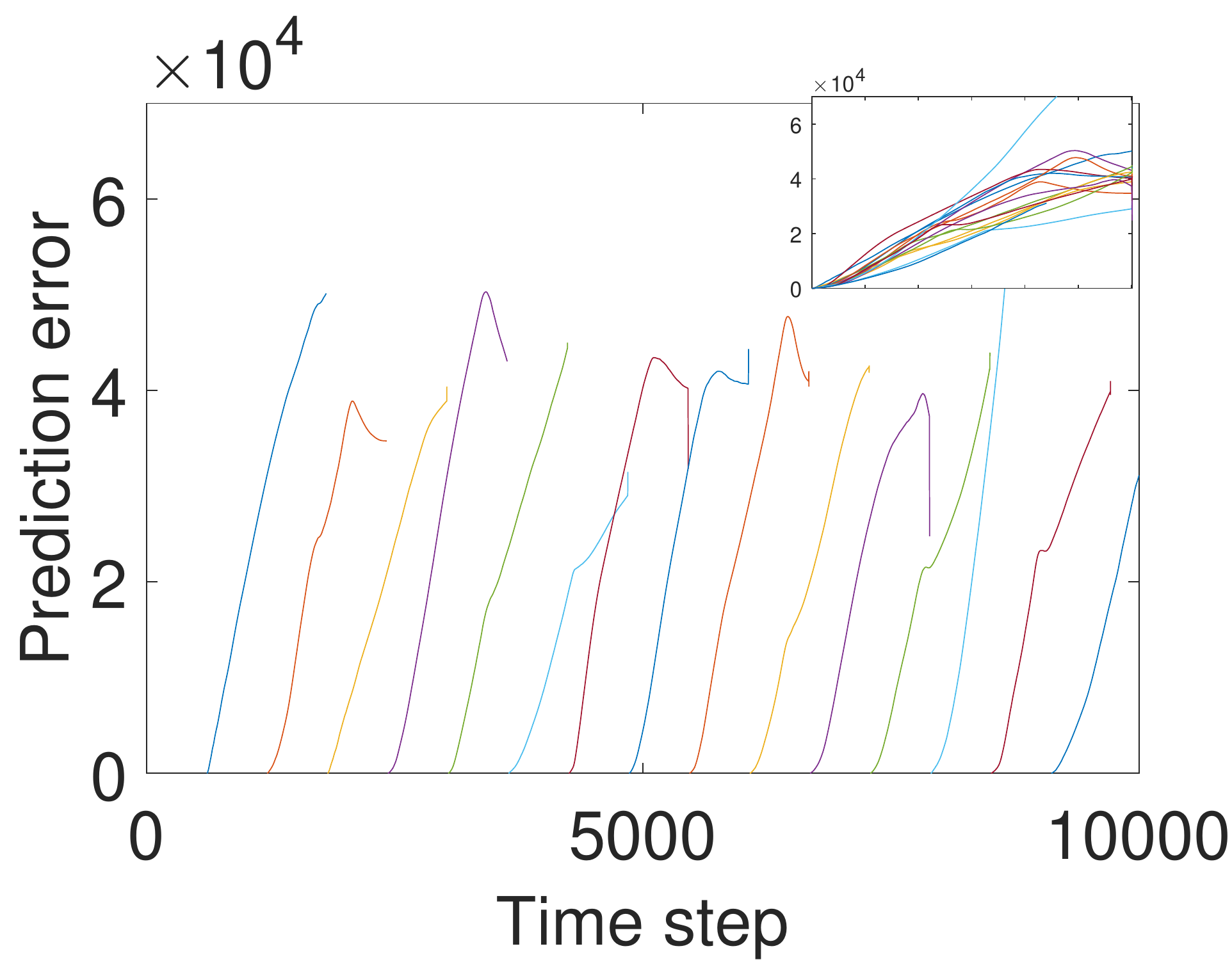}
        \caption{Erd\H{o}s\textendash R{\'e}nyi network}
        \label{Fig:PredictionERNet8step}
    \end{subfigure} 
    \begin{subfigure}[b]{0.24\textwidth}
        \includegraphics[width=\textwidth,height=3.2cm]{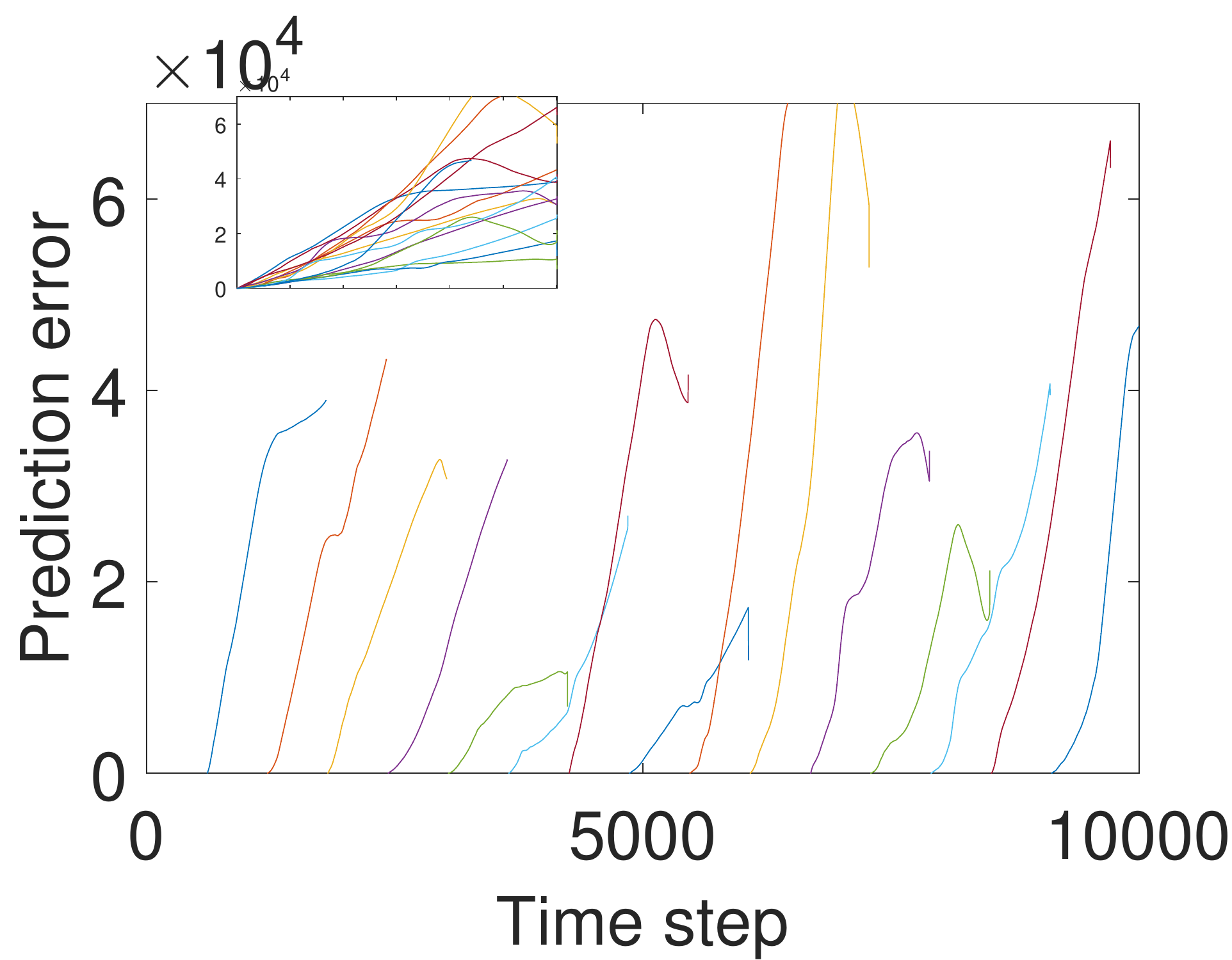}
        \caption{Scale-free network}
        \label{Fig:PredictionScaleFree8step}
    \end{subfigure} 
    \begin{subfigure}[b]{0.24\textwidth}
        \includegraphics[width=\textwidth,height=3.2cm]{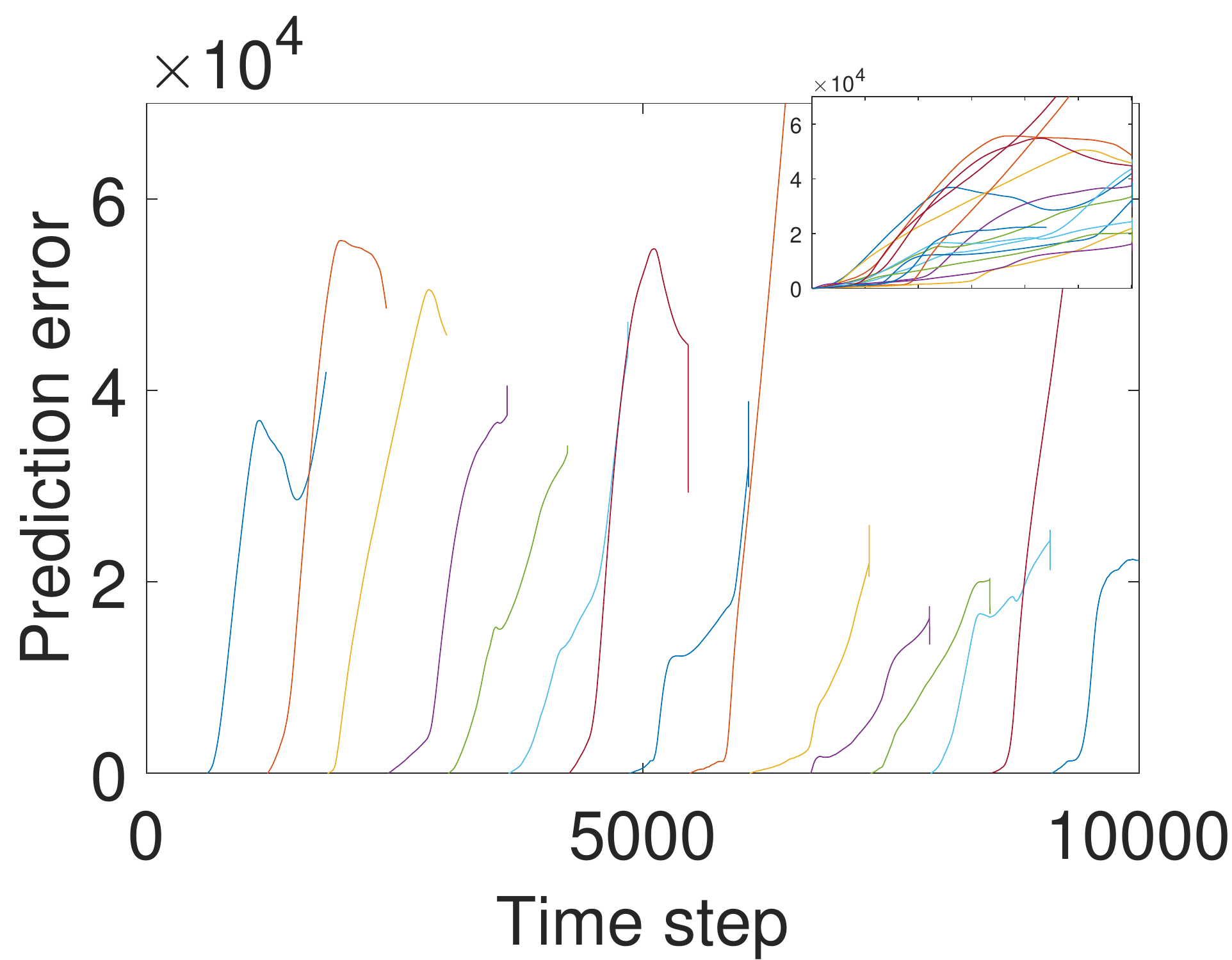}
        \caption{Small-world network}
        \label{Fig:PredictionSmallWorld8step}
    \end{subfigure}
    \caption{Average prediction errors (10 runs, top 5 individuals, observation period 8, online learning) for each communication method.}
    \label{Fig:Prediction8Step}
\end{figure*}

Further, in Figures~\ref{Fig:Prediction2Step},~\ref{Fig:Prediction4Step},~\ref{Fig:Prediction8Step}~and~\ref{Fig:Prediction8Step} we illustrate the prediction outcomes corresponding to the learning cycles throughout the swarm simulation, for all boid types and observation periods. The errors for each learning cycle represent the cumulated error over the whole population of 100 boids, and the figures display the average value of the cumulated errors over 10 runs of each case. The graphical presentation of the cumulated error versus time of simulation shows the magnitude of this error versus time, for each learning-prediction cycle throughout the swarm simulation. Based on these graphs it can be seen that the prediction error tends to be lower for classic boids, when compared to all network-based boids. However, this is not resulting unequivocally from this representation. For this reason, we also provide in the focused insights of insight of each figure an alternate representation of the cumulated errors, in which temporal information was suppressed in order to reveal errors\textquoteright  \  magnitude over their respective 1200 time-steps more clearly. The insight views show how from a magnitude point of view the prediction errors of classic boids tend to stabilize at lower values when compared to the network-based boids.

In summary, the investigation of both learning and prediction errors show that, similar to the off-line learning case, in the case of on-line learning the same trends are retained, i.e. the external observer can produce better learning and prediction results when observing the classic vision-based boids. This confirms once more that a swarm of network-based boids is more robust against external learners that try to infer the individual boids parameters underpinning the observable swarm behavior.

\subsection{Discussion}
First, we investigated how network-based neighbouring mechanisms in boids influences the resultant collective swarm behavior. For the proposed network-based boids, we used three well-known network topologies: scale-free, small-world and Erd\H{o}s\textendash R{\'e}nyi, and compared them with one model of classic vision-based boids recently reported in the literature~\cite{harvey2015application} as exhibiting stable and consistent behavior. Results presented in Section~\ref{Sec:EvalSwarmBehav} reproduced for classic vision-based boids the expected swarming quality, as reported in~\cite{harvey2015application}, and also showed that all network-based boids were able to produce swarming quality above that of the vision-based counterpart. 


Then, we investigated the robustness of both classic and network-based swarms, under the assumption that an external observer attempts to learn the underlying parameters of their individual rules. Thus, we intended to show that the ability to learn and predict is lower in the case of network-based boids. Results demonstrated this in two situations. First, we used off-line learning to show the maximal learning capability of the DE algorithm, in order to (1) validate the algorithm and (2) show how the external learner would perform in the ideal case when it has enough time to perform the learning off-line. Results showed that maximal learning capability was lower when applied against all network-based boids. Second, we investigated the realistic situation in which the external observer must learn and predict in real-time in a continuous and cyclic process of acquiring samples, learning and predicting. On-line learning enforced fixed periods for each step in the cycle, calculated with respect to DE computational cost and swarm simulation parameters, which truncated the learning and produced imperfect results. While ensuring consistency through an appropriate comparison base, results showed that the observer was able to learn and predict accurately the behavior of vision-based boids, but not that of the network-based ones. 


In summary, we believe that the proposed concept of network-based neighboring in swarms is a significant contribution, equally important to both Boids and their derived fields of applications. Since the nature of this study touched the fundamental aspects of Boids, and relied on graph theoretical aspects, we envisage that significant amount of further work can take place in the direction of the foundations of swarms, where existing studies on sensory-based swarms can be rethought in the dimension of network connectivity. However, if the foundation of swarm behavior is touched, subsequent fields like PSO or massively parallel crowds simulation can certainly benefit from the proposed changes. \rtext{We also mentioned earlier in the paper the field of swarm robotics, which inspired us somewhat in relation to the second objective of the paper.} However, we did not put much emphasis on this aspect in the current paper to maintain the scope of the paper within the theoretic abstract swarms. We do believe though that network-based neighboring in swarms is going to be useful for swarm robotics, where the embodiment of the particles and the information and communication technologies that enable them are the \textquoteleft testbed\textquoteright  \ where network relations between individuals can achieve maximum potential.

\begin{figure*}[h]
    \center
    \begin{subfigure}[b]{0.24\textwidth}
        \includegraphics[width=\textwidth,height=3.2cm]{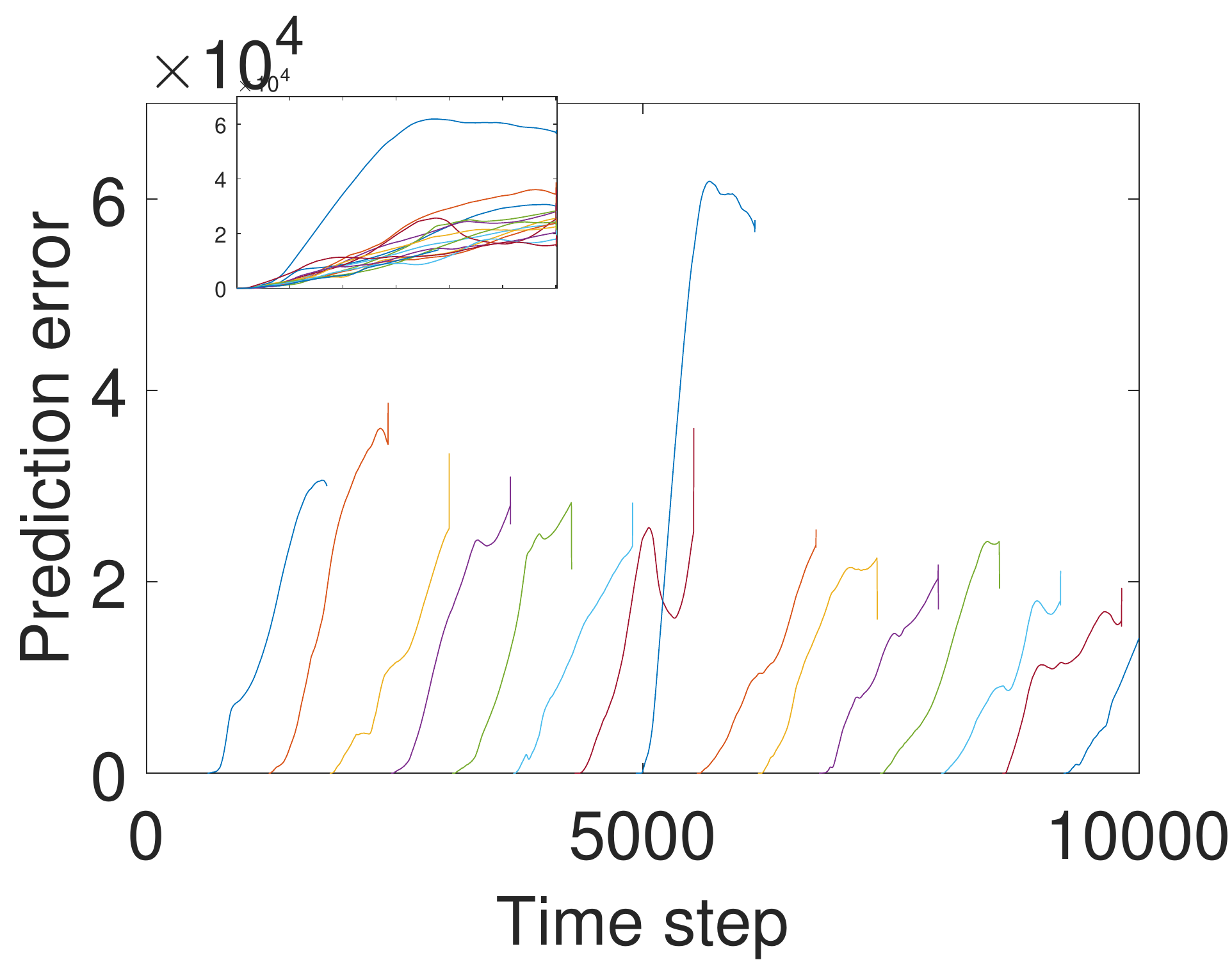}
        \caption{Classic boids}
        \label{Fig:PredictionClassic16step}
    \end{subfigure} 
    \begin{subfigure}[b]{0.24\textwidth}
        \includegraphics[width=\textwidth,height=3.2cm]{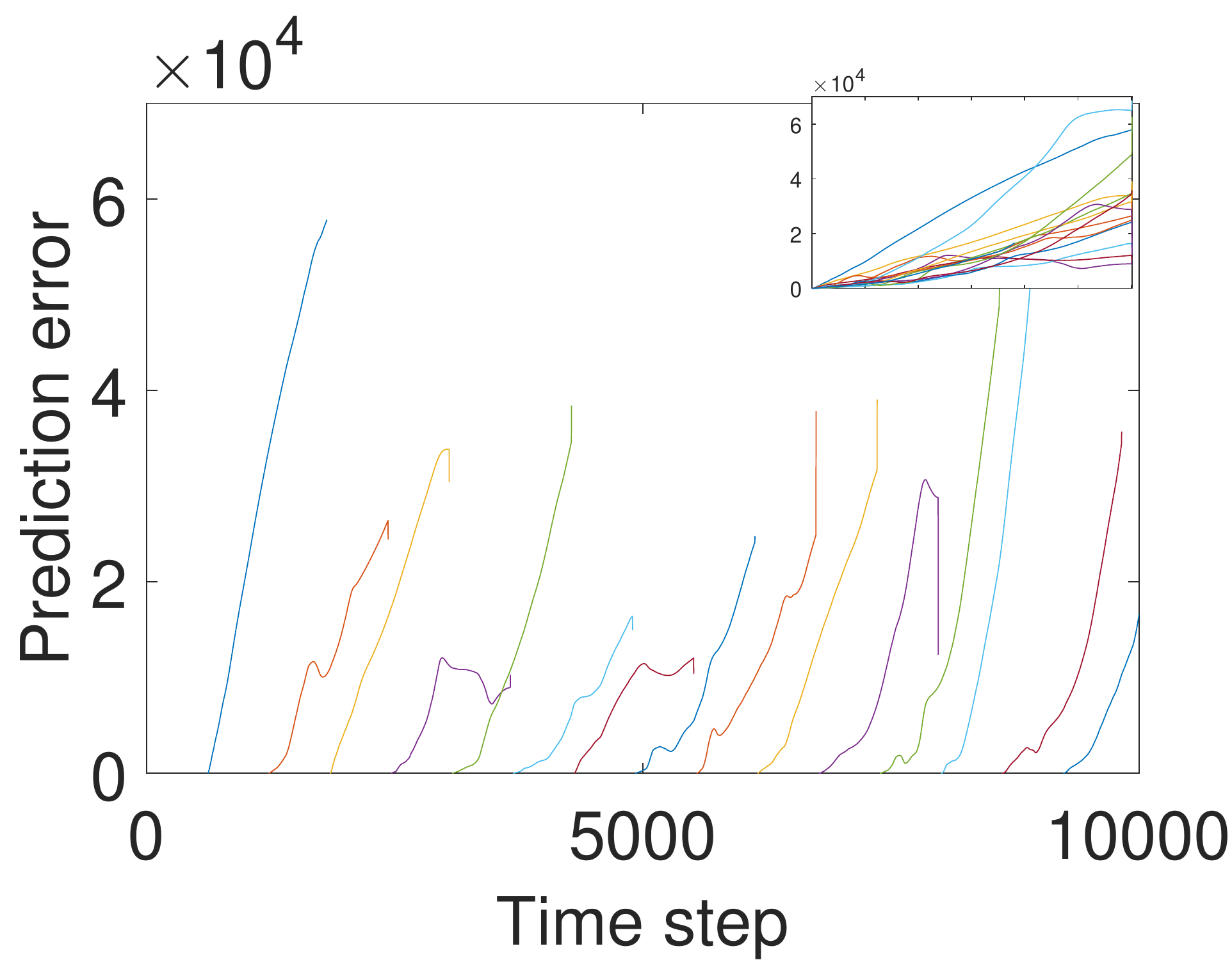}
        \caption{Erdos-Renyi network}
        \label{Fig:PredictionERNet16step}
    \end{subfigure} 
    \begin{subfigure}[b]{0.24\textwidth}
        \includegraphics[width=\textwidth,height=3.2cm]{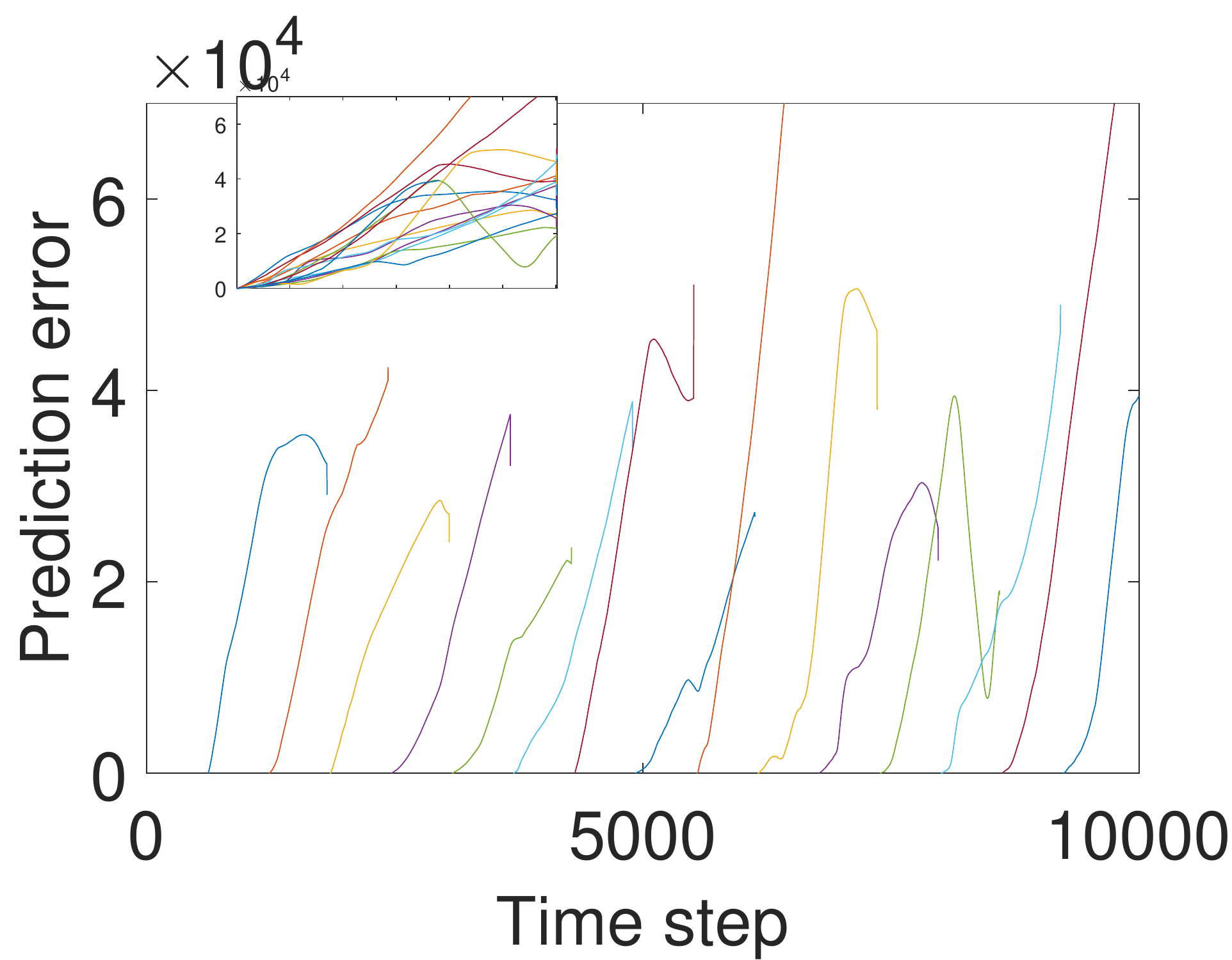}
        \caption{Scale-free network}
        \label{Fig:PredictionScaleFree16step}
    \end{subfigure} 
    \begin{subfigure}[b]{0.24\textwidth}
        \includegraphics[width=\textwidth,height=3.2cm]{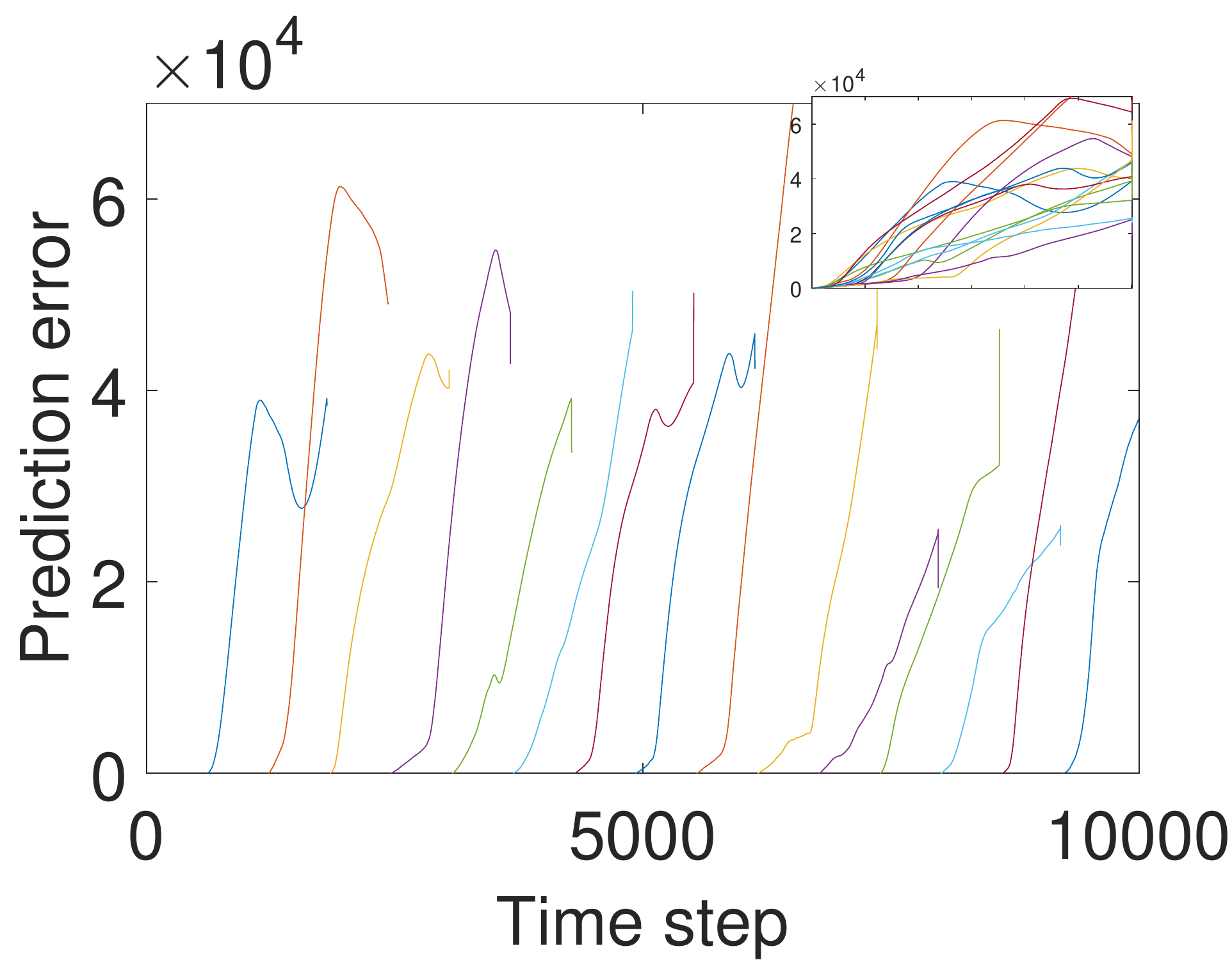}
        \caption{Small-world network}
        \label{Fig:PredictionSmallWorld16step}
    \end{subfigure}
    \caption{Average prediction errors (10 runs, top 5 individuals, observation period 16, online learning) for each communication method.}
    \label{Fig:Prediction16Step}
\end{figure*}

\section{Conclusions} \label{Section:Conclusions}
In this study, we proposed a network-based approach to Boids behavior in which we stepped away from classic definition of boids neighborhoods through sensory perception and Euclidian space locality, and considered graph theoretical network relations instead of sensory-based relations. We intended to demonstrate that if collective behavior is based on network relations between individuals rather than on sensory-based relations in Euclidean space, then the resultant collective behavior of the swarm can be improved in two directions: (1) the network-based behavior leads to faster swarming and higher quality of the formation, and (2) the resultant swarm is more robust against adversarial learning which intends to infer its underlying individual behavior rules based on observation of its whole behavior.

The results confirmed both hypotheses, offering substantial evidence to believe that this novel view on swarms can open doors for significant amount of future work in both directions. Thus, we conclude this paper by proposing to extend the work to other form of swarming, such as Schooling and Fishing.




\begin{IEEEbiography}[{\includegraphics[width=0.9in,height=1.2in]{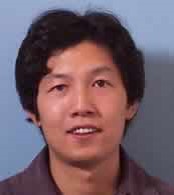}}]
{Jiangjun Tang} is a research associate in the School of Engineering and IT, UNSW-Canberra. He holds MS(IT) from Australian National University in 2004 and a PhD (CS) from UNSW-Canberra in 2012. His research interest includes Computational Red Teaming, behavior modelling, air traffic management, simulation and modelling, data mining, and cognitive science.
\end{IEEEbiography}

\begin{IEEEbiography}[{\includegraphics[width=0.9in,height=1.2in]{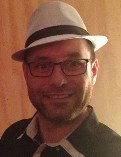}}]
{George Leu} is a Research Associate position in the School of Engineering and IT, UNSW-Canberra. He holds B.Eng.(Hon.) in communication systems engineering from the Military Technical Academy of Romania in 2003, a M.Eng. in ElecEng\&IT from the National Defense Academy of Japan in 2008, and a PhD in computer science from UNSW-Canberra in 2013. His research interest includes computational modelling of human cognition \& behaviour.
 \end{IEEEbiography}
 
\begin{IEEEbiography}[{{{{{\includegraphics[clip,width=0.9in,height=1.2in]{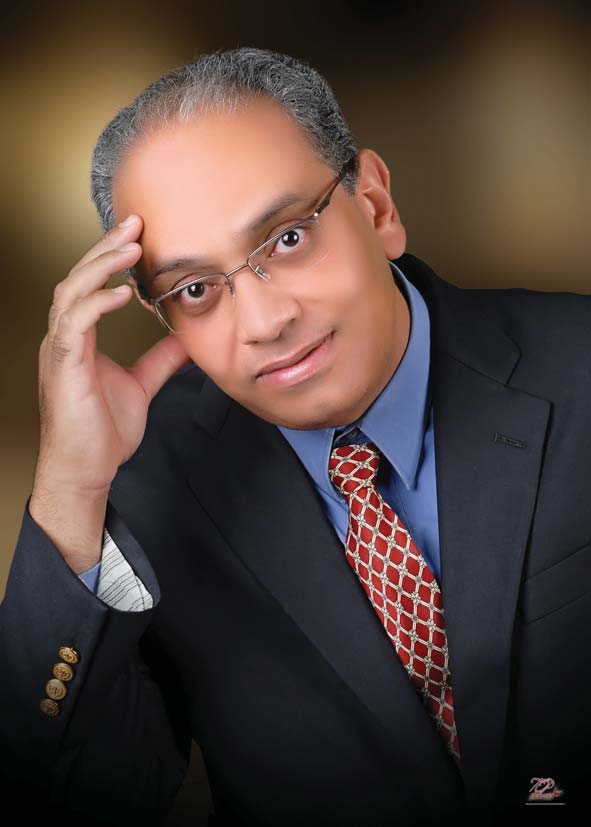}}}}}}]
{Hussein Abbass} is a Fellow of the UK Operational Research Society \& the Australian Computer Society. He is the Vice-President for Technical Activities, IEEE Computational Intelligence Society. He is an AE of the IEEE Trans. Evolutionary Computation, IEEE Trans. Cognitive \& Developmental Systems, IEEE Trans. Cybernetics, IEEE Trans. Computational Social Systems and four other journals. His work on Trusted Autonomous Systems integrates cognitive science, operations
research and Artificial Intelligence.
\end{IEEEbiography}

\end{document}